\documentclass[acmtog]{acmart}
\acmSubmissionID{955}

\usepackage{booktabs} 

\usepackage{subfigure}
\usepackage{mdframed}
\usepackage{subfiles}
\usepackage[toc,page]{appendix}
\usepackage{fontawesome5}
\usepackage{hyperref}

\usepackage[ruled]{algorithm2e} 

\SetAlFnt{\small}
\SetAlCapFnt{\small}
\SetAlCapNameFnt{\small}
\SetAlCapHSkip{0pt}

\acmJournal{TOG}

\begin{document}

\title{Aladdin: Zero-Shot Hallucination of Stylized 3D Assets from Abstract Scene Descriptions}

\author{Ian Huang}
\affiliation{%
 \institution{Stanford University}
 \city{Stanford}
 \state{CA}
 \country{USA}}
\email{ianhuang@cs.stanford.edu}
\author{Vrishab Krishna}
\affiliation{%
 \institution{Stanford University}
 \city{Stanford}
 \state{CA}
 \country{USA}}
\email{vrishab@stanford.edu}
\author{Omoruyi Atekha}
\affiliation{%
 \institution{Stanford University}
 \city{Stanford}
 \state{CA}
 \country{USA}}
\email{oatekha@stanford.edu}
\author{Leonidas Guibas}
\affiliation{%
 \institution{Stanford University}
 \city{Stanford}
 \state{CA}
 \country{USA}}
\email{guibas@cs.stanford.edu}

\begin{teaserfigure}
\centering
\includegraphics[width=\textwidth]{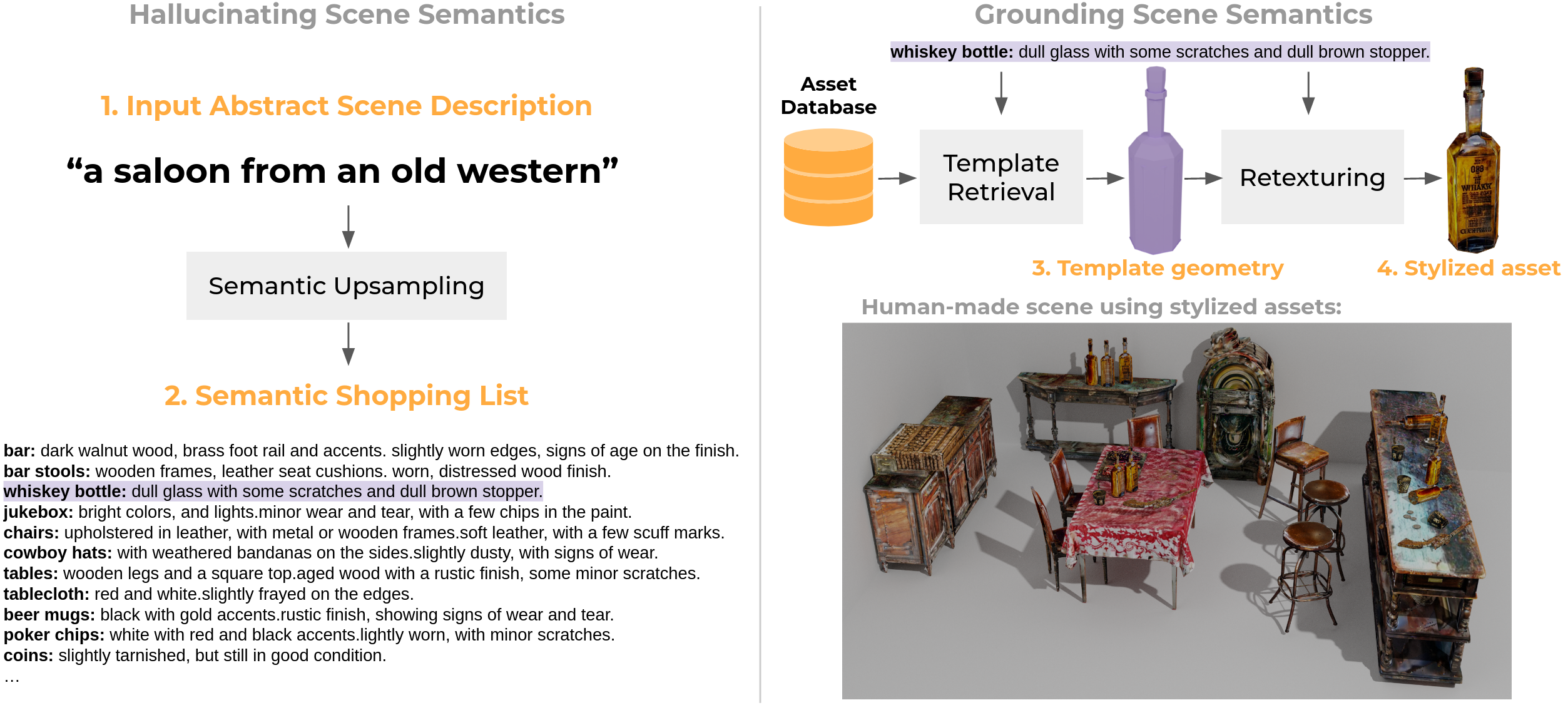}
\caption{Our system produces stylized assets to fit a scene description. Given an abstract scene description that does not provide details on what objects should be found within that scene, our system (1) infers a \emph{semantic shopping list}, a  human-readable and editable list of object categories and appearance attributes, and then uses this to (2) retrieve template shapes from a 3D asset database before (3) re-texturing them to fit the desired appearance attributes. The output of our system is a collection of textured meshes, which can be directly imported into 3D design software and used for other downstream tasks. Note the correspondences in the assets on the right with many of the desired object categories and appearance attributes generated by our system on the left!}
\label{fig:teaser}
\end{teaserfigure}

\begin{abstract}
What constitutes the ``vibe'' of a particular scene? What should one find in ``a busy, dirty city street'', ``an idyllic countryside'', or ``a crime scene in an abandoned living room''? The translation from abstract scene descriptions to stylized scene elements cannot be done with any generality by extant systems trained on rigid and limited indoor datasets. In this paper, we propose to leverage the knowledge captured by foundation models to accomplish this translation. We present a system that can serve as a tool to generate stylized assets for 3D scenes described by a short phrase, without the need to enumerate the objects to be found within the scene or give instructions on their appearance. 
Additionally, it is robust to open-world concepts in a way that traditional methods trained on limited data are not, affording more creative freedom to the 3D artist. Our system demonstrates this using a foundation model ``team'' composed of a large language model, a vision-language model and several image diffusion models, which communicate using an interpretable and user-editable intermediate representation, thus allowing for more versatile and controllable stylized asset generation for 3D artists. We introduce novel metrics for this task, and show through human evaluations that in 91\% of the cases, our system outputs are judged more faithful to the semantics of the input scene description than the baseline, thus highlighting the potential of this approach to radically accelerate the 3D content creation process for 3D artists.
\end{abstract}


%
%

%
%

\maketitle

\section{Introduction}

While language-to-shape generation has taken the world by storm, scene generation has been less accessible by language control partially because the language content needed to express details of a scene becomes prohibitively cumbersome for average human creators. Additionally, manually searching, selecting, and retexturing/restylizing assets from online 3D repositories is, in aggregate, a very time-consuming task for a single object, not to mention the task of doing so for 30-50 objects that may make up a 3D scene.

From a user's perspective, wouldn't it be convenient to say ``generate a scene of the financial district of New York'' and have the system infer \emph{what} should be in the scene and \emph{how} every item should look? In other words, we would like to build a system that hallucinates both semantic and visual detail from an \emph{abstract} high-level scene description -- something that human users can provide much more conveniently than fully enumerative language found in \cite{achlioptas2020referit3d, ilinykh2019tell, chang2015scenelex}.

While valuable, this is not a problem that can be solved using traditional machine learning approaches, primarily due to limitations in data -- the indoor scene datasets that dominate the domain of scene generation are largely limited in scene and object diversity \cite{roberts2021hypersim, song2017suncg, song2015sunrgbd, chang2017matterport3d, fu20213dfront}, far from open-vocabulary. The same can be said for language-scene multimodal datasets, where language labels are either limited or prohibitively enumerative to train for our primary task of interest \cite{achlioptas2020referit3d, chang2015scenelex}.

In this paper, we ask, how far can zero-shot inference using foundation models go, with their common sense understanding \cite{brown2020language, rombach2022high, radford2021learning}, in facilitating the 3D scene creation process for 3D artists? We introduce a system that allows 3D content creators to synthesize entire asset collections from abstract scene descriptions (e.g. “a busy city street”), by leveraging the immense amount of progress in Large Language Models and Vision Language Models.

To go from abstract description to stylized asset collections, we break the process into 3 stages. In the first stage, we ``semantically upsample'' the input abstract description into a plausible list of objects, attributes and appearances (which we call the ``semantic shopping list'') that may compose the described scene. For this, we use in-context prompting of LLM's \cite{brown2020language}, exploiting common sense knowledge of scene composition embedded within LLM's. The second stage requires a retrieval from an existing 3D asset database, given the attributes and appearances hallucinated in the semantic shopping list. We use visual and textual similarity given by large vision-language models like CLIP to retrieve top candidates. Finally, we use diffusion models to texture the surface of the objects given their hallucinated appearance attributes.

Our system uses natural language as an intermediary representation between these stages, for 3 reasons:
\textbf{(1) Interpretability and Editability}: This means that users can visualize, interpret and edit the intermediary outputs. This is important, since this work employs a ``team'' of foundation models for the first time, where the output of one may not necessarily -- in a zero-shot sense -- be optimal as an input into another to accomplish the user's artistic intent. 
\textbf{(2) Varying Abtraction Levels}: Given language's ability to represent information at a variety of abstraction levels, it as a medium that allows both the large language model (as well as user edits) to specify semantic constraints at a wide range of specificity. 
\textbf{(3) Moore's law, but for foundation models}: Given recent trends, we're anticipating that the foundation models used in this paper will have more powerful replacements soon. We expect that users of our system will be able to ``upgrade'' different modules with the latest models.

Our system integrates with existing 3D asset databases and treats its assets as templates for both the appearance and the geometry. The benefits of this are two-fold: (1) while large amounts of work that does scene generation is reliant and restricted on indoor scene datasets \cite{paschalidou2021atiss, chang2015scenelex, ma2018languagescenedatabases}, our method can generate outdoor scenes and radically out-of-distribution scenes as well, by leveraging diverse and larger-scale shape databases \cite{deitke2022objaverse, chang2015shapenet, selvaraju2021buildingnet} and (2) building ontop of a 3D asset store allows usage of such a system to be specialized, depending on the asset store provided, not to mention that it allows for nice priors, important for both geometric and textural manipulation \cite{michel2022text2mesh, hui2022neural, xu2022dream3d,  michel2022text2mesh}.


The main contribution of this paper is three-fold: 
(1) we present the task of \emph{stylized asset curation} given \emph{abstract scene descriptions}, which, to the knowledge of the authors, has not been considered in isolation.
(2) we present a system that tackles this task using the zero-shot capabilities of foundation models, and contribute a method that does this using semantic upsampling through in-context learning. 
(3) We introduce a new metric, CLIP-D/S, which can be used to measure both the diversity of the asset collection and the semantic alignment with respect to a target scene description.
In addition to quantitative and qualitative evaluations, our human evaluation experiments conducted using 72 evaluators showcases the efficacy of our system, and the value of semantic upsampling as a key powerhorse in the quality of the generated assets and assembled scenes. Code for our system and metrics can be found at \href{https://github.com/ianhuang0630/Aladdin}{https://github.com/ianhuang0630/Aladdin \faGithubSquare}.

\section{Related Works}

Works like \cite{michel2022text2mesh, xu2022dream3d, sanghi2022clipforge, fu2022shapecrafter, jun2023shapE, nichol2022pointe, lin2022magic3d, gao2022get3d, jain2022dreamfields, poole2022dreamfusion, changeit3d, huang2022ladis} focus on generating shapes from natural language. However, as many of them use non-mesh-based 3D representations like implicit representations \cite{xu2022dream3d, poole2022dreamfusion, jain2022dreamfields, lin2022magic3d}, extracting meshes from them gives rise to disruptive artifacts in both texture and geometry, limiting the usability of the asset in almost all 3D design applications. As such their outputs are not optimized for usage by human users, since composing and editing scenes using implicit representations of assets remains non-trivial. Additionally, such  systems are not optimized to read between the lines -- the desired output is oftentimes what is described verbatim, given its object-centric focus. However, for abstract scene descriptions, compositional understanding beyond what is typically captured by vision-language models is needed.

On the other hand, works on mesh generation and texturing using text prompts \cite{michel2022text2mesh, sanghi2022clipforge, xu2022dream3d} make use of vision-language models like CLIP \cite{radford2021learning} coupled with differentiable rendering to optimize the mesh to correspond to a certain text embedding. These methods manage to edit the mesh to become semantically similar to the text prompt, but since CLIP was not directly optimized to guide differentiable rendering, the resulting optimization often leads to improbable or unrealistic outputs, as can be seen by disruptive artifacts that often give a distorted and blocky feel to the outputs. Meanwhile, through newer generative models, the world knowledge obtained from large-scale image-text datasets is easily accessible. \cite{lin2022magic3d} takes a step in this direction, using image diffusion models to generate high resolution textures of a mesh. However, this requires a detailed description specific to the individual objects to be generated, which is not provided a priori in our problem setting.

Along this line, works like \cite{fridman2023scenescape}, \cite{hoellein2023text2room}, and \cite{zhang2023text2nerf} introduce pipelines that incorporate image diffusion models \cite{ho2020denoising, saharia2022imagen, rombach2022high} to create \emph{scenes}, but not in a way that allows assets that compose the scene to be easily and effectively extracted. Scenescape \cite{fridman2023scenescape} and Text2Room \cite{hoellein2023text2room} are two similar methods that make use of depth prediction models to craft a mesh using iterative predictions from a image diffusion model. SceneScape \cite{fridman2023scenescape}, making use of generated super-resolution videos, is biased towards producing scenes that are long and tunnel-like, thereby restricting the set of producible scenes. Similarly, Text2Room \cite{hoellein2023text2room} can only generate closed, star-convex meshes due to the depth projection approach. The fundamental drawback of these methods is that the end result is a single connected mesh with limited flexibility to extract and edit assets. Meanwhile, \cite{po2023compositional} recently introduced a model that uses locally conditioned diffusion to generate the scene compositionally by using different language instructions to generate different patches of the scene (e.g. ``a firepit'' in one part of the scene, ``a tent'' in another). However, not only do the generated assets suffer from the same aforementioned weaknesses in regards to mesh extraction, but the generative pipeline also requires \emph{fully enumerative} language input, in contrast with the focus of this work. 

Although considerate effort has been made towards 
collections of 3D scene datasets \cite{roberts2021hypersim, song2017suncg, song2015sunrgbd, chang2017matterport3d, fu20213dfront}
as well as 
training models to generate and position elements within indoor scenes \cite{chang2015scenelex, ma2018languagescenedatabases, paschalidou2021atiss, ritchie2019fast, wang2019planit, wang2021sceneformer}, these were not designed to handle open vocabularies of objects, which makes them limited for creative applications. Moreover, the latter works do not have the ability to re-texture scene elements to better match the input language description, which is a prime focus of our system.

\section {Our Method}

\subsection{Semantic Upsampling}

\begin{figure}
\centering
\includegraphics[width=0.5\textwidth]{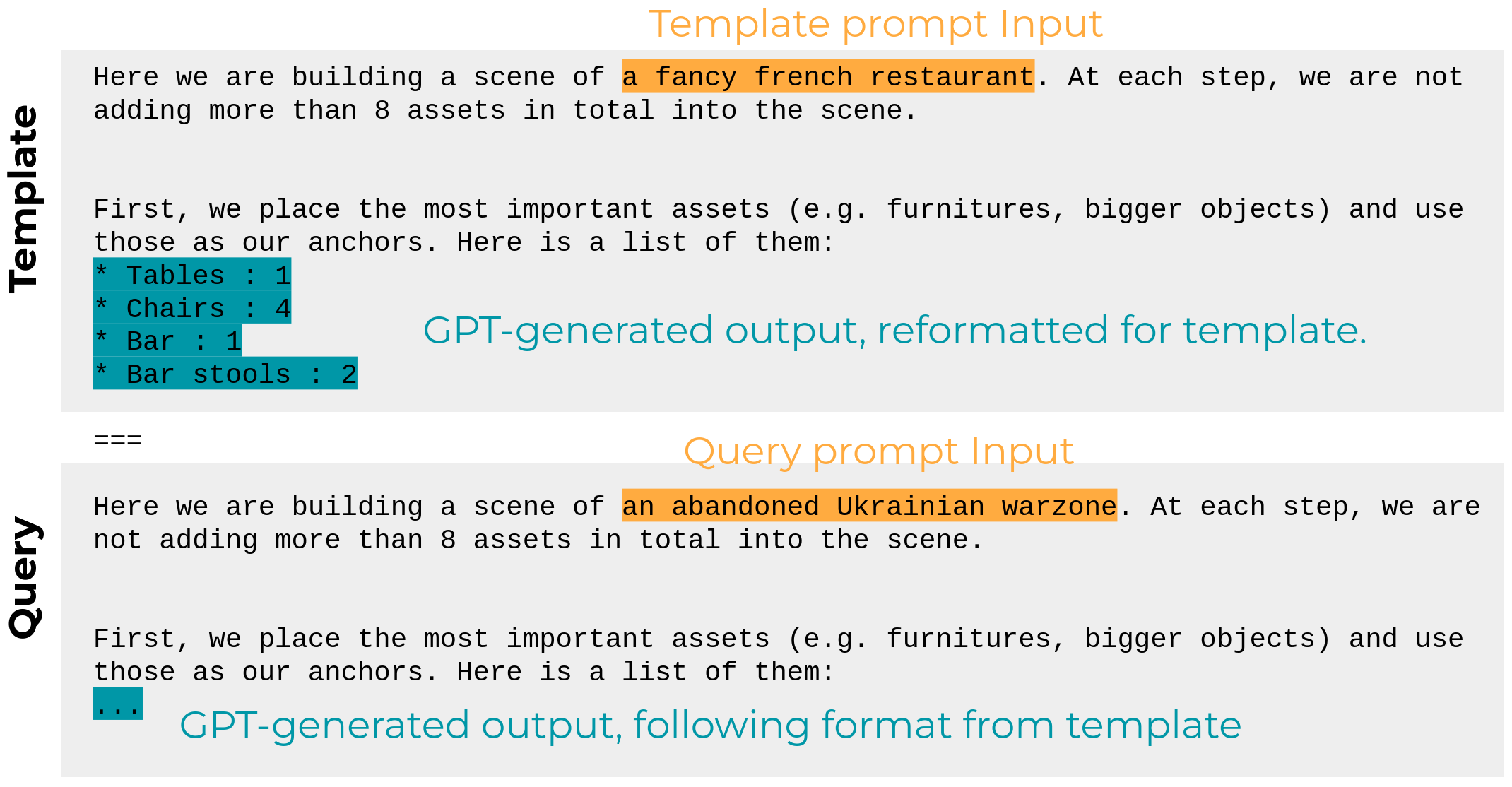}
\caption{The template and query segments of the GPT-3 input share the same structure up to the part where GPT-3 is prompted to do next-token prediction. We use this template for all generations of the anchor objects within the  scene.}
\label{fig:sem_template}
\end{figure}

Given an abstract scene description, our system ``upsamples'' the semantics of the scene description to the level of object categories, properties and appearance. To do this, we use few-shot prompting of GPT-3 \cite{brown2020language}, which has shown to be very useful in other settings \cite{wei2021incontext1, rubin2021incontext2, zhang2022incontext3, wang2022incontext4, zhou2022incontext5, wei2022incontext6, shin2022incontext7, zhao2021incontext8, chen2022incontext9, min2021incontext10, dong2022incontext11}.
 
\begin{figure}
\centering
\includegraphics[width=0.5\textwidth]{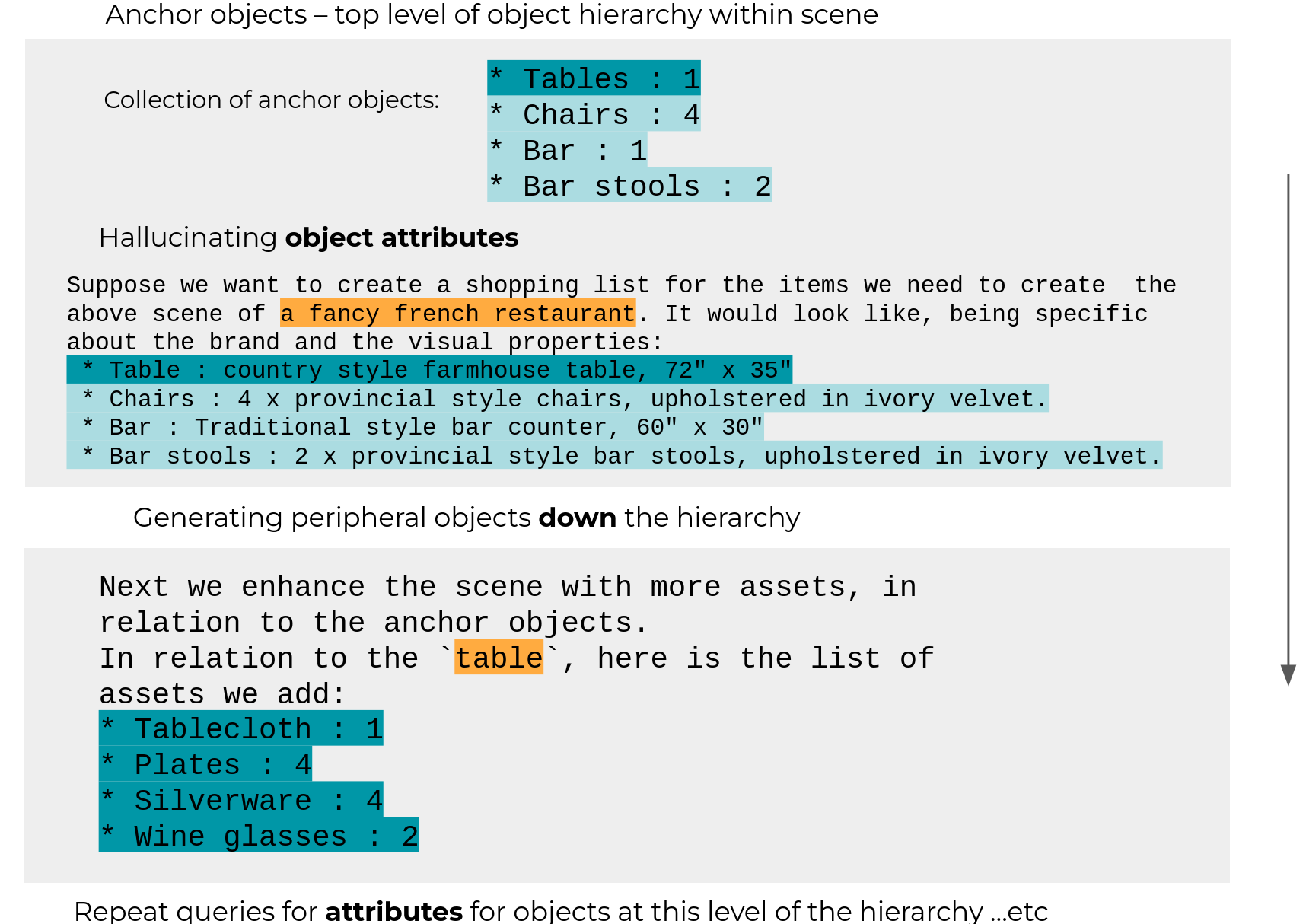}
\caption{We move down the scene hierarchy by asking GPT-3 to generate peripheral objects around each of the objects in the current level. Usually, this results in smaller and more peripheral objects that add to the realism of the scene. We use in-context learning again to generate their attributes.}
\label{fig:sem_hierarchy}
\vspace{-0.5cm}
\end{figure}

To do this, we create templates that cover a variety of different aspects of objects that may be found within the scene; object category, style, material properties, and condition (e.g. scratched, unused, rusted, ...). These templates can be found in the Appendix. Templates are used for two main reasons: (1) they effectively enforce a prior over the kind of attributes that one would like to use to describe objects within the scene and (2) they dictate a textual format that can be very easily parseable by our system (e.g. comma separated attributes, colon separation between object category and attributes.).

In practice, we found that querying for \textit{all} the objects within a scene at once can lead to degenerate results -- generating details for way too many objects at once may cause the objects chosen to ``drift'' semantically away from the prompt. As such, we adopt a more hierarchical approach, where we first use in-context learning to ask GPT-3 \cite{brown2020language} to generate a set of ``anchor'' objects (typically, this is a small set of 6-8 objects) and their attributes (Figure \ref{fig:sem_template}). For each of these anchor objects, we ask it to hallucinate objects (and their attributes) found ``around'' the anchor object (Figure \ref{fig:sem_hierarchy}), and  repeating this recursively down the hierarchy. This works fairly well to elucidate the hierarchy of objects, and can be useful for object placement (e.g. for a ``fancy french restaurant'', an anchor object generated is a table, and objects generated \textit{around} this anchor object are objects typically found \textit{on} the table). Additionally, doing this hierarchically means that for abstract descriptions that involve a large set of objects, we need only call this  procedure a few times before we arrive at the ``leaf'' objects within the implicit object hierarchy. A traversal through this hierarchy allows a  full list of objects and appearance attributes of objects likely to be found within the described scene. We will refer to this list as the \emph{semantic shopping list}. An example of this is shown on the left in Figure \ref{fig:teaser}.  

\subsection{Object retrieval \& retexturing}

Given the semantic shopping list from semantic upsampling, the system use CLIP \cite{radford2021learning} embeddings of both visual renderings and textual annotations of objects within asset databases to retrieve the template geometries for each object. 

This is, however, a nuanced objective; since all objects selected during retrieval will go through diffusion-based \emph{re}-texturing, it's tempting to disregard the original texturing altogether, and retrieve only using a query composed of the object category information from the semantic upsampling (ignoring object attributes, which will be "painted" on in a later stage). In practice, this leads to suboptimal retrieval results. Some object attributes (e.g. ``old'' in ``old car'') are less solely based in texture, affecting both the visual appearance and the geometry. Moreover, the pretrained model of CLIP was trained on natural images, which relies on color properties for accurate similarity evaluation (similar observations have also been reported in \cite{michel2022text2mesh}). As such, using a textureless rendering of the candidate asset can actually \textit{hurt} the retrieval performance.

To match the open-world vocabulary found in semantic shopping lists, it is essential to have a large and diverse asset database to choose from. For this paper, we've chosen to use a combination of Future3D~ \cite{fu20213d} and a 30K-subset of Objaverse~\cite{deitke2022objaverse}. Future3D specializes in objects commonly found in indoor environments, and is a useful dataset for the majority of ``base'' object found within indoor scenes, which we anticipate would make up of the majority of user scene queries. Objaverse is a lot more diverse in object category, and serves for the ``personality'' pieces of indoor scenes (e.g. the sword along the wall in Figure \ref{fig:kingshand1}), which allows the scene to be more faithful to the ``vibe'' communicated in the input description. Additionally, it contains object categories typically found outdoors, which allows our system to construct outdoor scenes in ways that previous scene generation pipelines cannot (see Figure \ref{fig:rustic_backyard}).

In our current implementation, we use the thumbnails of different assets to derive the CLIP image embeddings, since (1) these are readily available in most datasets and (2) human artists already use them to judge the appropriateness of a particular asset for their scene. Future works extending our pipeline can use more complex rendering techniques for different objects, and the question of how renderings should be done to encourage high accuracy 3D asset retrieval is an important direction for future work.

Enforcing stylistic consistency from the retrieval stage is hard. Empirically, we notice that using just the semantic shopping list alone often leads to retrieval of objects that are stylistically inconsistent in their template geometry, and thus do not aesthetically combine well once put in the same scene. This is because though semantic upsampling hallucinates visual details, it has no context of what would be important for stylistic consistency in the retrieved results \textit{downstream}. Therefore, we merge the abstract scene description into all retrieval and texturing queries, for all objects, as a fail-safe for when the semantic shopping list provides inadequate stylistic information. 

Given that many 3D assets have language annotations, we incorporate that information when determining the K-nearest neighbors through a simple linear weighting of the language- and image-based cosine similarities. Doing brings some more robustness to the retrieval process in the case when the asset thumbnail does not reflect the geometric content as well as its textual annotations do.


Once we have the template objects, we make use of pre-existing image generation pipelines to texture each retrieved object. Using an available depth-guided and language-guided image diffusion models \cite{rombach2022high}, we can generate images corresponding to views of an object and use differentiable rendering to optimize our mesh texture to match the generated image, while encouraging 3D consistency between different views through depth and language conditioning. We use the implementation of a recent paper \cite{richardson2023texture} to achieve this.

\vspace{-0.1cm}
\section{Experiments}

The main output of our system is a set of textured assets. To demonstrate the usability of our system outputs, we source ideas for input scene descriptions from 8 people who do not have any prior 3D design experience or experience interacting with our system. 20 such prompts were collected, ranging in \emph{plausibility} (from ``a romantic french restaurant'' to ``a church for strawberries''), \emph{emotional valence} (from ``a marvel-themed bedroom  for a five-year-old toddler'' to ``murder in an abandoned living room'') and \emph{complexity} (from ``a rustic backyard in the countryside'' to ``a busy street in downtown new york''). A full collection of the abstract scene descriptions can be found in the Appendix, as well as their corresponding visualizations.

We provide the prompts to our system, and -- for the purposes of this paper -- run our system in a fully automated way, sidestepping the possible option of user edits of the semantic shopping list between different stages. To do this, we use the same query string (generated from the semantic upsampling stage) for the retrieval and texturing stages, and automatically select the top-1 sample in CLIP-Similarity in the retrieval outputs for texturing. 

Note that to demonstrate the robustness of this system and the benefit of basing it on foundation models, we \emph{do not cherrypick between different runs for the same input prompt}. In other words, all visualizations of scenes are done based on assets generated in a single pass.

\subsection{Composing assets into scenes}
To construct the final scene, the authors of this paper import the generated assets into Blender \cite{blender} and create 3D scenes according to the following rules: (1) they are allowed to translate, rotate and scale any 3D asset in the generated collection along any axis, (2) they are allowed to add ground and wall planes to the scene, (3) they are allowed to omit subsets of the asset collection from visualization, (4) they are allowed to duplicate assets as many times as they wish, and (5) they are \emph{not} allowed to change the material properties of the textured mesh, except emissive properties for assets that should emit light (rare). On average, the importing, arrangement and rendering of a single scene took 20 minutes.

\subsection{Evaluating stylized asset collections}

Within the literature, CLIP-Similarity (or CLIP-S) has been recently used  to measure adherence of generative output to the semantics of the input text \cite{xu2022dream3d, fu2022shapecrafter}. Given a vision encoder $v$ and a language encoder $g$, rendered views of the object $x_i$ and associated language description $l$, CLIP-S is defined as:
\begin{equation}
    S (x, l) = \max_i v(x_i)^T g(l)
\end{equation}

However, using CLIP-S directly on our task has serious drawbacks. First, it's an observed phenomenon that CLIP's language model oftentimes behaves like a Bag-Of-Words model \cite{michel2022text2mesh,  yuksekgonul2022bagofwords}, where important relations between entities or concepts are often not reflected in its similarity evaluations. This motivates why it's inappropriate to use such a metric to evaluate the adherence of a stylized asset to \emph{the set of objects that likely composes} a scene of a particular semantic -- the relationships that are key to the idea of \emph{scene membership} (i.e. that an asset belongs to a scene) can be overpowered by the description of the scene itself. Empirically, we've found that such a metric tends to slightly favor the outputs of the system when semantic upsampling is \emph{not used} and assets are retrieved and textured according to the abstract scene description, though this comes as little surprise. The distinction is demonstrated in Figure \ref{fig:saloon_baseline}, which shows that CLIP (and by extension, CLIP-S) cannot favor \emph{assets that compose a scene} over assets that may resemble the abstract prompt but do not compose that scene. We would like a metric that exhibits this behavior.

To solve this problem, we introduce the idea of CLIP-Diversity (CLIP-D), a score that is high when the assets that is generated are semantically varied. This metric counteracts the favoring of systems that generate assets that are very narrowly aligned with the scene description. Assuming the same visual encoder $v$, and that $x^j_i$ is the renderings of asset $j$ from angle $i$, as well as a function $m$ that averages on the surface of the unit sphere over a set of points on the unit hypersphere of the CLIP embedding space, we define CLIP-Diversity as the \emph{negative} mean pairwise cossine similarity between assets within the collection:
\begin{multline}
 D(\{x^j\}_{j=1...N}) = \\
- \frac{2}{N(N-1)} \sum_{i < j} m\biggr(\{v(x^j_k)\}_{k=1...K}\biggr)^T m\biggr(\{v(x^i_k)\}_{k=1...K}\biggr)   
\end{multline}

However, diversity alone does not provide adherence to a language instruction and can be satisfied without consideration for the language prompt. As such, we construct CLIP-D/S, a metric that additively combines CLIP-D and CLIP-S over an asset collection $\{x^j\}$ and a language instruction $l$ from a collection $L$ of augmented utterances based on the scene description (see Section \ref{sec:system_outputs}), given equal weighting:

\begin{equation}
DS(\{x^j\}_{j=1...N}, l) = D(\{x^j\}_{j=1...N}) + \frac{1}{N|L|}\sum_{j} \sum_{l \in L} S(x^j, l)
\end{equation}

CLIP-D/S is therefore a combination of diversity and similarity, and can be used heuristically to measure the fidelity and usefulness of the asset collections that our system generates.

\subsection{System outputs}\label{sec:system_outputs}

\begin{figure*}
\centering
\includegraphics[width=0.9\textwidth]{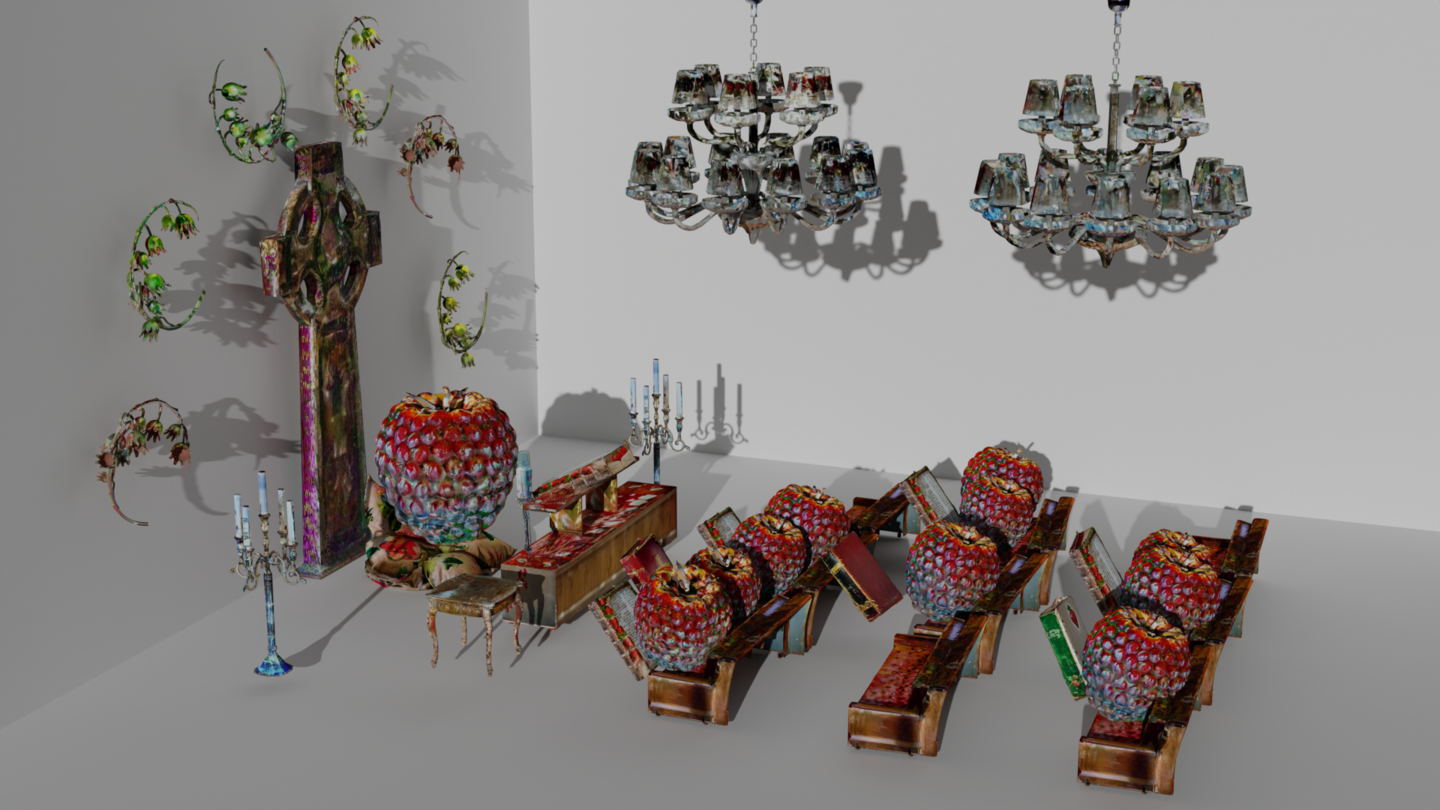}
\caption{A scene of ``a Church for Strawberries'' is one of the more out-of-distribution queries given to our system, but through a combination of human creativity and assets generated by our system, a rather funny scene emerges.}
\label{fig:strawberry_church}
\end{figure*}
 
\begin{figure*}
\centering
\includegraphics[width=0.9 \textwidth]{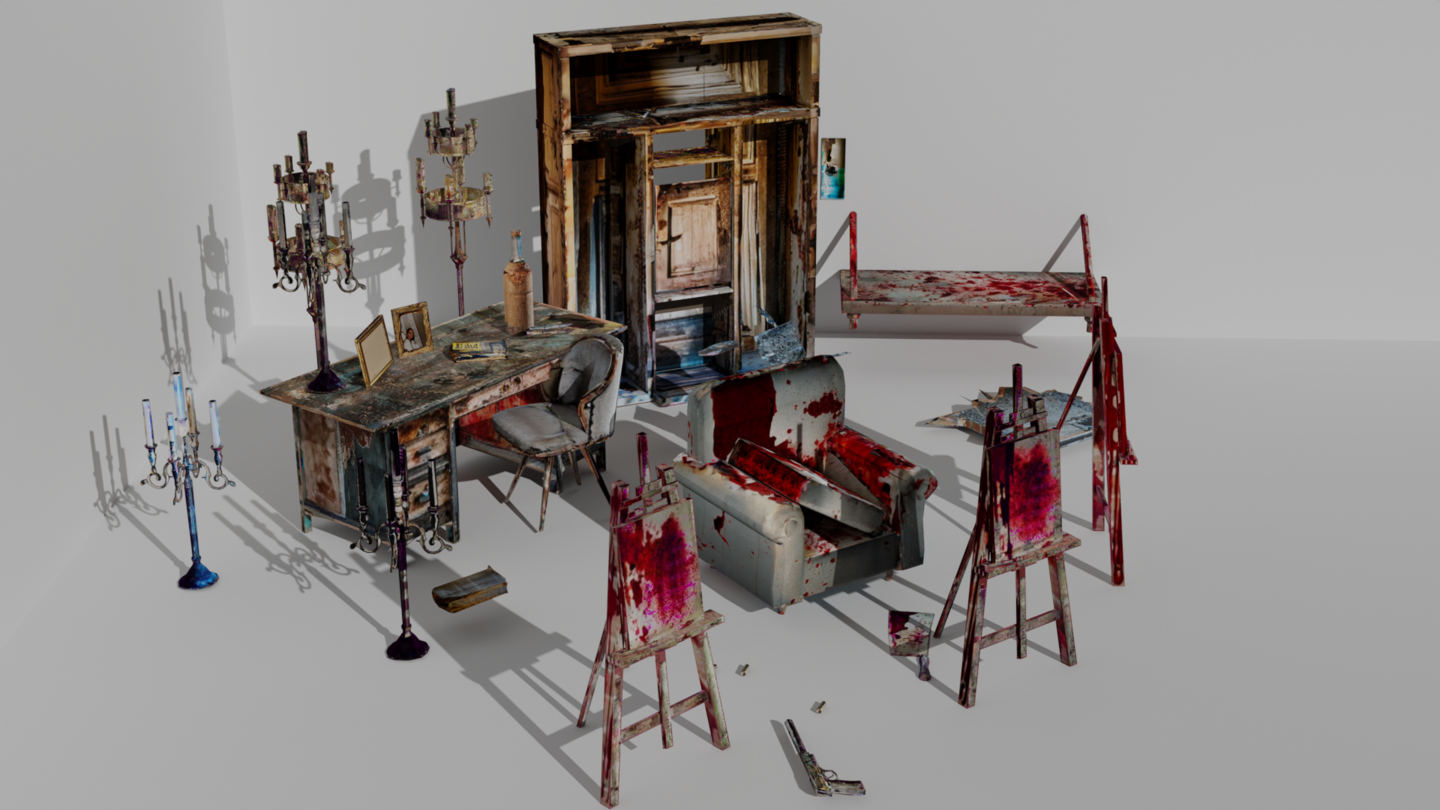}
\caption{A scene of ``a murder in an abandoned living room''. The hallucinations of semantic upsampling tells a gruesome and disturbing story, with the cleaver placed near the bloodied couch, the gun with empty shells around it, and crimson smears on the canvases. We acknowledge the graphic nature of this scene, include this example to show that our system is capable of producing scenes at the extremities of emotional valences, unlike more traditional scene generation systems.}
\label{fig:murder_abandoned_livingroom}
\end{figure*}

\begin{figure*}
\centering
\includegraphics[width=0.9 \textwidth]{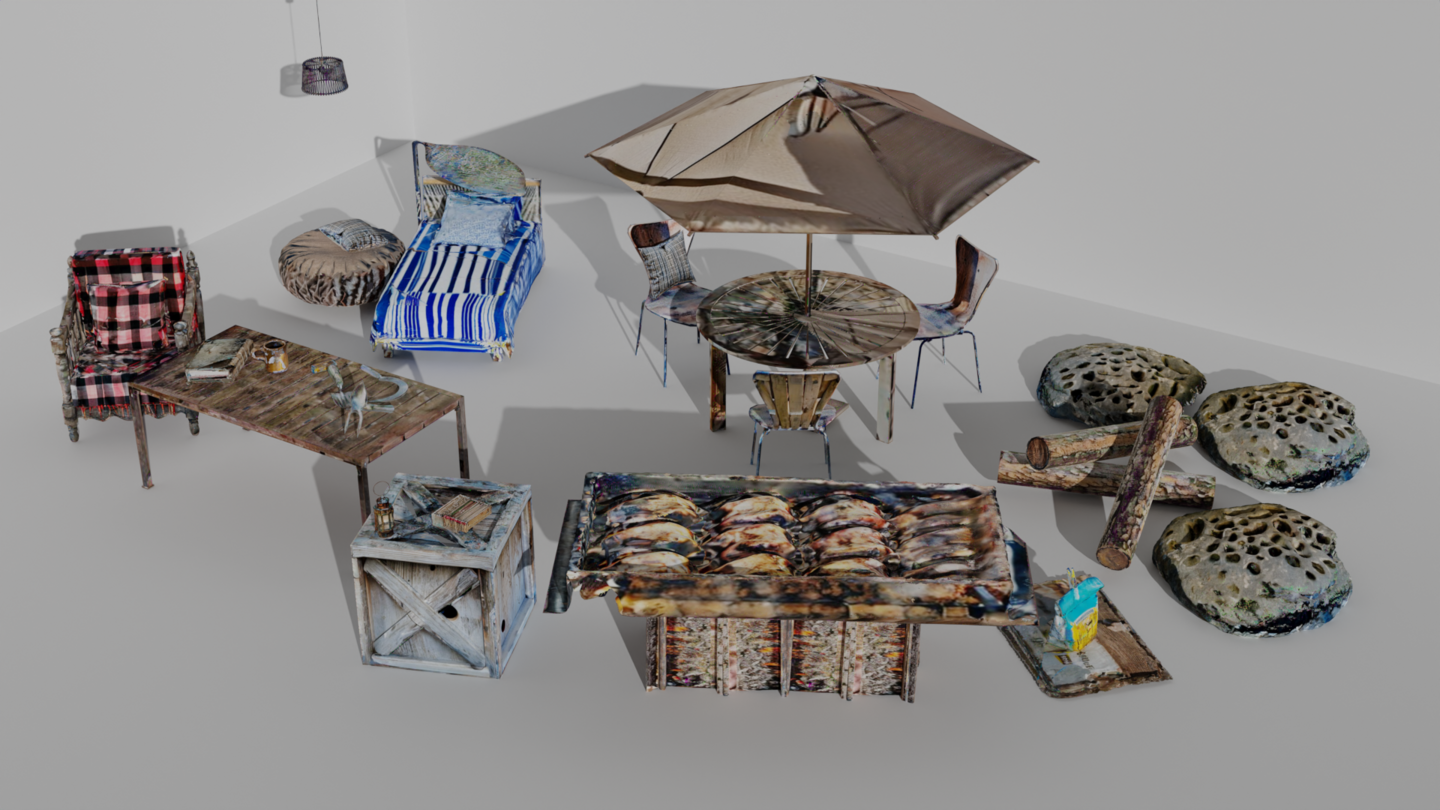}
\caption{A scene of ``a rustic backyard in the countryside''. This example demonstrates the potential of our system to create outdoor-esque scenes by using the common-sense reasoning of foundation models. Notice the elements that suggests the outdoor environment -- the gardening equipment on the table, the barbeque grill and bag of coal, the logs and rocks, match sticks and kerosene lamp, and the umbrella table.}
\label{fig:rustic_backyard}
\end{figure*}

\begin{figure*}
\centering
\includegraphics[width=0.9 \textwidth]{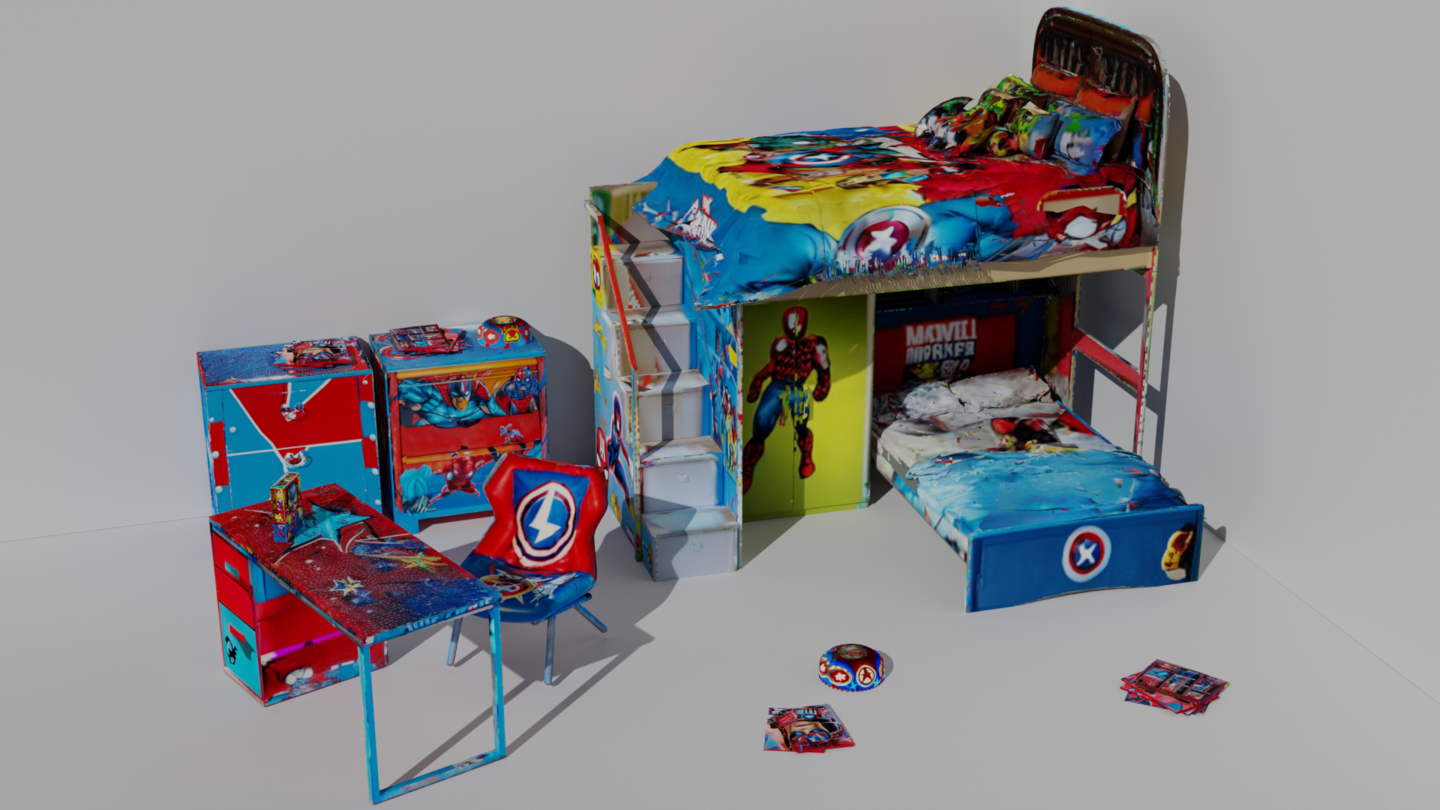}
\caption{A scene of ``a marvel-themed bedroom of  a five-year old toddler''. As foundation models are trained mostly by data on the internet, we observe that it's able to understand references to pop culture fairly well, resulting in this very prominently marvel-themed bedroom.}
\label{fig:marvel_bedroom}
\end{figure*}

\begin{figure}
     \centering
     \includegraphics[width=0.45\textwidth]{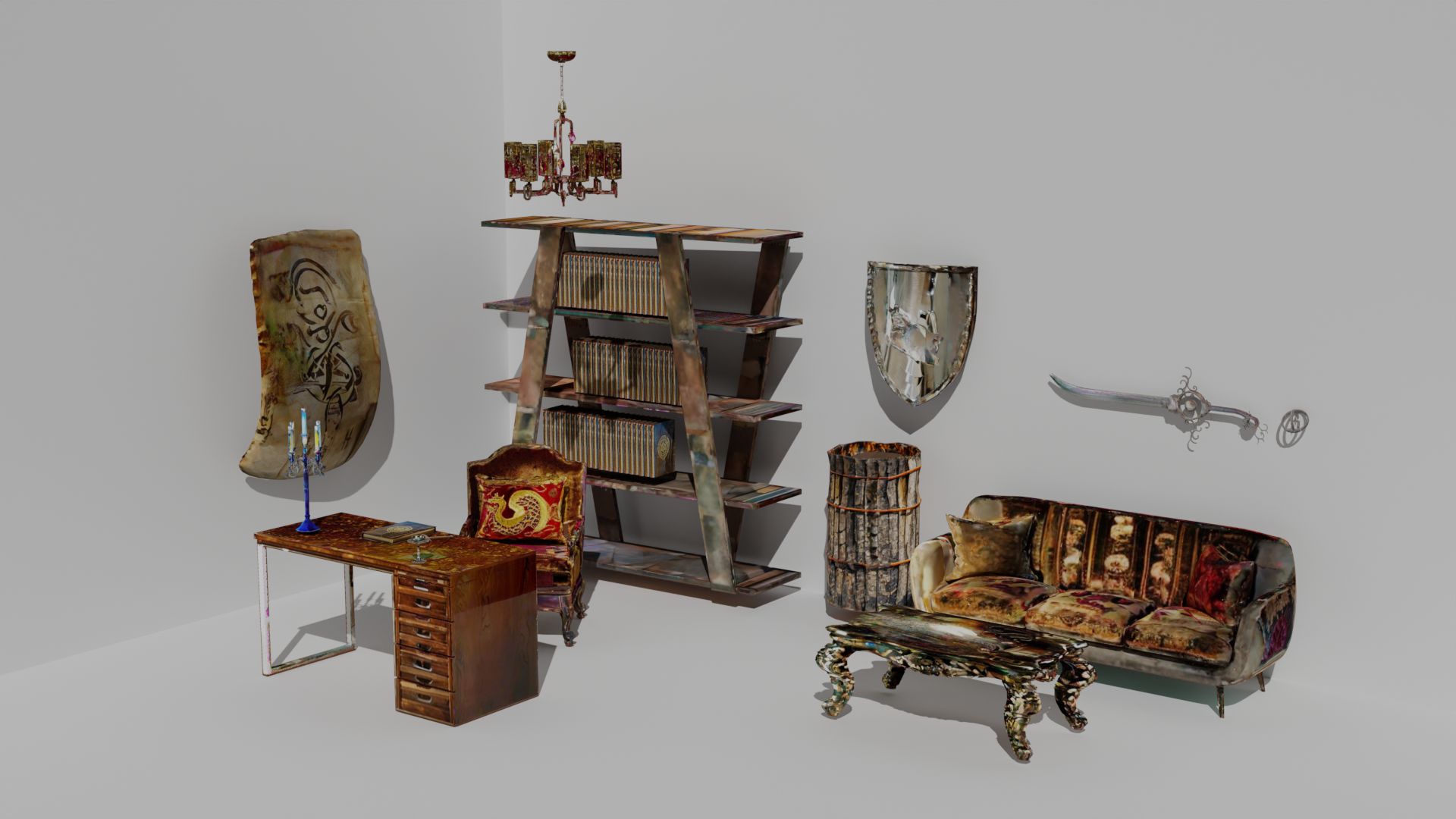}
     \caption{Asset arrangement from the \textbf{first} run of our system for the input ``office of the King's Hand in Game of Thrones''.}
     \label{fig:kingshand1}
\end{figure}

\begin{figure}
    \centering
    \includegraphics[width=0.45\textwidth]{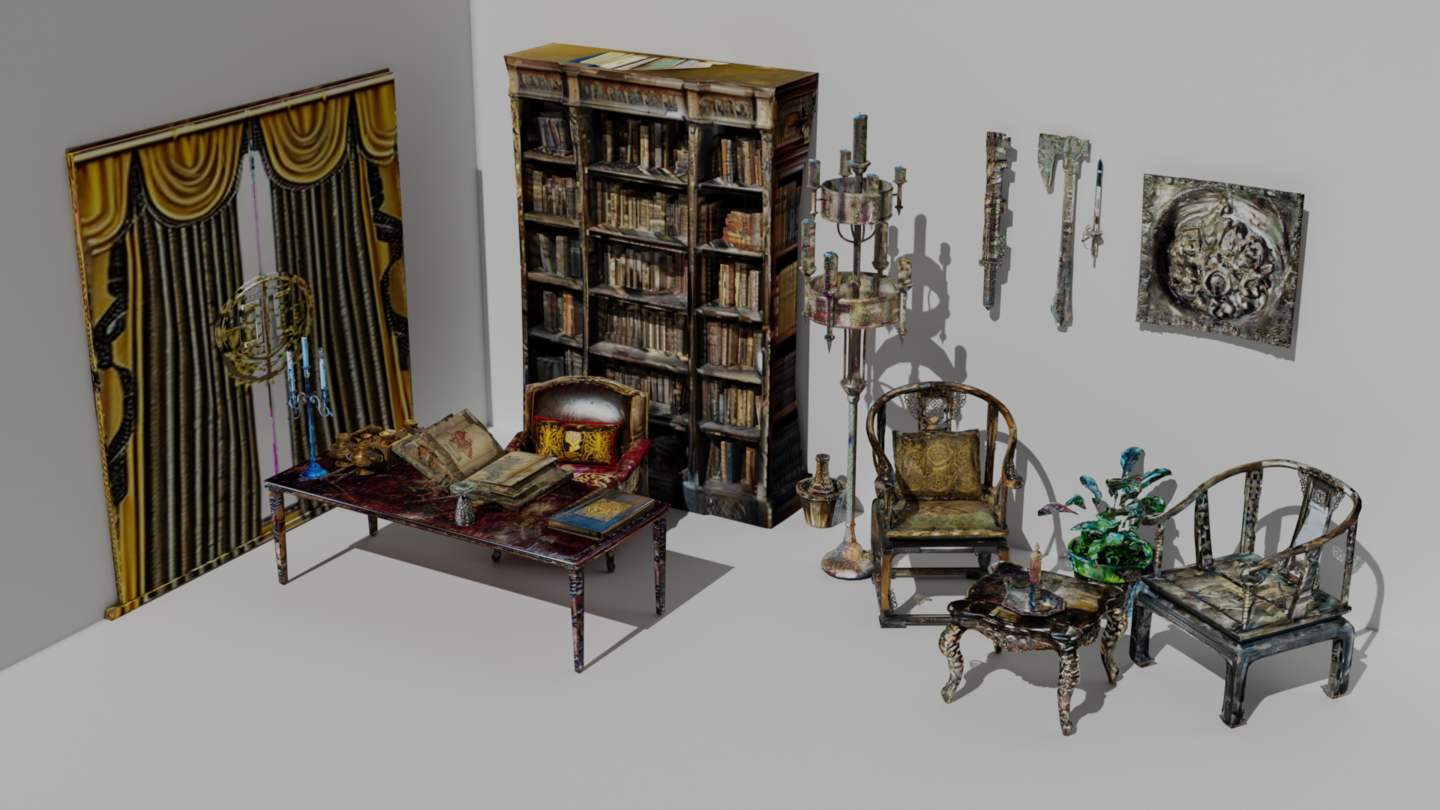}
     \caption{Asset arrangement from the \textbf{second} run of our system for the input ``office of the King's Hand in Game of Thrones''.}
     \label{fig:kingshand2}
\end{figure}

Figures \ref{fig:teaser},  \ref{fig:strawberry_church}, \ref{fig:murder_abandoned_livingroom}, \ref{fig:rustic_backyard}, \ref{fig:marvel_bedroom}, \ref{fig:kingshand1}, and \ref{fig:kingshand2} show some outputs from our system, arranged into 3D scenes. 
A longer list of examples, along with their corresponding semantic shopping lists, can be found in the Appendix. Predominantly, the benefit of this system is its ability to add ``character'' to a scene through inferring a wider, more diverse set of object categories.


A practical property of this system is that due to the inherent randomness present in next-token generation of GPT-3, running the system twice will create differing semantic shopping lists. This is a useful property for 3D artists, since this allows them to semantically densify their scenes by rerunning the semantic upsampling process. As shown in Figures \ref{fig:kingshand1} and \ref{fig:kingshand2}, multiple runs can come up with different but valid assets, where the union or intersection of them could create even richer and accurate scenes.

As a measure of stylistic adherence, we alternatively ask, given an asset we've generated for each of the scenes, how well can one predict which scene they were generated for? We use the CLIP-S metric as a zero-shot classifier to classify each of our 572 stylized assets across the 20 scenes that they were generated for (full list can be seen in Table \ref{tab:clipds_per_session} and in Appendix), and find a top-1 classification accuracy of \textbf{32.69\%}, substantially higher than the accuracy of guessing randomly (5\%). We consider the predicted scene to be the scene that maximizes the average CLIP-S score across $L$, which is a set of language augmentations on the abstract scene description: (1) ``an element in a scene of \verb|[SCENE DESCRIPTION]|'', (2) ``an object from a scene of \verb|[SCENE DESCRIPTION]|'', (3) ``a picture of an object form \verb|[SCENE DESCRIPTION]|'', (4) ``a rendering of an asset from a 3D scene of \verb|[SCENE DESCRIPTION]|'' and (5) ``\verb|[SCENE DESCRIPTION]|''. 

\subsection{The importance of semantic upsampling} \label{sec:semantic_autocompletion_ablation}

The main contribution of our work is the use of in-context learning to generate semantically meaningful details as they pertain to assets. How important is this step? We compare against a baseline method that is exactly the same as the method proposed, \emph{except} that it retrieve and retextures according to the input abstract scene description, instead of the semantic shopping list given by semantic upsampling. For this, we retrieve and retexture the top-$K$ assets that have the highest CLIP-similarity with the abstract scene description , where $K$ is the number of assets generated by our method.

\begin{figure}
\centering
\includegraphics[width=0.45\textwidth]{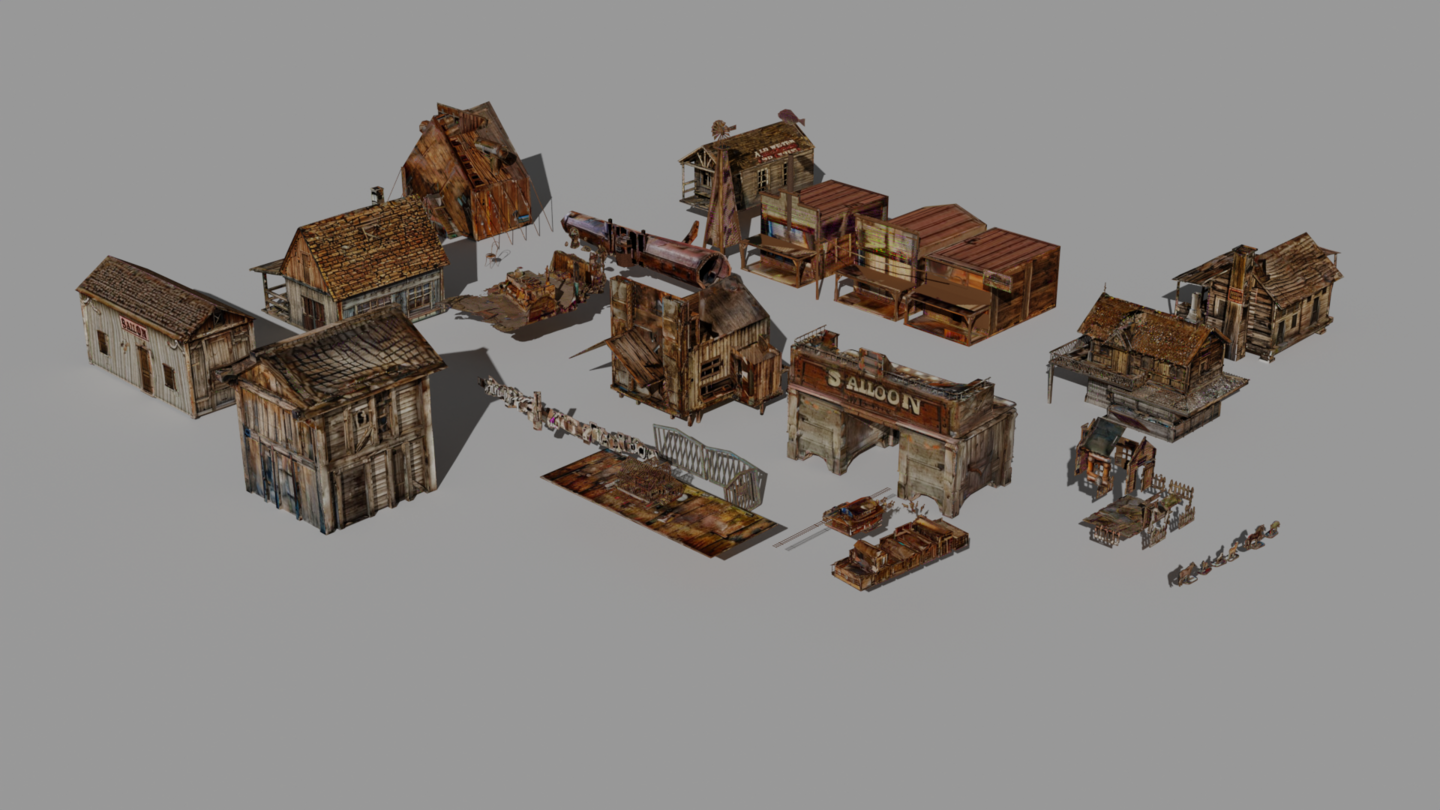}
\caption{Retrieval and retexturing for a scene of ``a saloon from an old western'' \emph{without} the use of semantic upsampling.  The outputs can be very narrowly aligned with the ``western saloon''  concept, but are not elements that can compose the scene described, unlike the assets in Figure \ref{fig:teaser}.}
\label{fig:saloon_baseline}
\end{figure}

Table \ref{tab:clipds_per_session} shows the impact of removing semantic upsampling on the diversity (CLIP-D) and the CLIP-D/S score for the generated assets for each of the 20 scenes. This corroborates the observation that in the best case, as shown in Figure \ref{fig:saloon_baseline}, assets that align very well with the scene itself might get retrieved, resulting a narrow selection that cannot be used to compose the scene. Or, as is often the case, an erroneous template shape is retrieved, and confuses the downstream retexturing to produce poorly textured 3D assets. Please see the Appendix for examples of this phenomenon.

\begin{table}[t]

\begin{tabular}{ccc|cc}
\toprule
Scene Reference & D (b)& D (o) $\uparrow$ & D/S (b) & D/S (o) $\uparrow$ \\
\midrule
rustic backyard & -0.84 & \textbf{-0.80} & -0.61 & \textbf{-0.60} \\
futuristic teahouse & -0.89 & \textbf{-0.81} & -0.67 & \textbf{-0.62} \\
confucius bedroom & -0.86 & \textbf{-0.81} & -0.61 & \textbf{-0.58} \\
alien teagarden & -0.84 & \textbf{-0.80} & -0.63 & \textbf{-0.60} \\
retro arcade & -0.84 & \textbf{-0.79} & -0.59 & \textbf{-0.56} \\
anne frank room & -0.87 & \textbf{-0.80} & -0.64 & \textbf{-0.57}\\
hades cave & -0.85 & \textbf{-0.76} & -0.63 & \textbf{-0.56} \\
shrek home & -0.86 & \textbf{-0.78} & -0.59 & \textbf{-0.53} \\
smurf house & -0.88 & \textbf{-0.77} & -0.65 & \textbf{-0.56} \\
mad scientist restaurant & -0.81 & \textbf{-0.78} & -0.61 & \textbf{-0.60} \\
western saloon & -0.82 & \textbf{-0.79} & -0.60 & \textbf{-0.59} \\
occult cult & -0.82 & \textbf{-0.79} & -0.61 & \textbf{-0.59} \\
marvel bedroom & -0.91 & \textbf{-0.87} & -0.63 & \textbf{-0.59} \\
murder room & -0.85 & \textbf{-0.77} & -0.62 & \textbf{-0.57} \\
strawberry church & -0.84 & \textbf{-0.79} & -0.58 & \textbf{-0.57} \\
poseidon living room & -0.83 & \textbf{-0.77} & -0.61 & \textbf{-0.55} \\
north korean classroom & -0.85 & \textbf{-0.77} & -0.62 & \textbf{-0.57} \\
antichrist vatican & -0.82 & \textbf{-0.77} & -0.60 & \textbf{-0.57} \\
romantic restaurant & -0.81 & \textbf{-0.78} & -0.59 & -0.59 \\
busy new york street & -0.79 & -0.79 & \textbf{-0.62} & -0.63 \\
\bottomrule
\end{tabular}
\caption{A comparison of CLIP-D (abbreviated D) and CLIP-D/S (abbreviated D/S) for all 20 scenes created using assets generated by our method (abbreviated \textbf{o}) and those generated by a baseline method (abbreviated \textbf{b}), which does not use semantic upsampling. Our method generally produces both higher performance in both CLIP-D (diversity) and CLIP-D/S.}
\label{tab:clipds_per_session}
\vspace{-1cm}
\end{table}

\subsection{The importance of retexturing}

\begin{figure}
    \centering
    \includegraphics[width=0.45\textwidth]{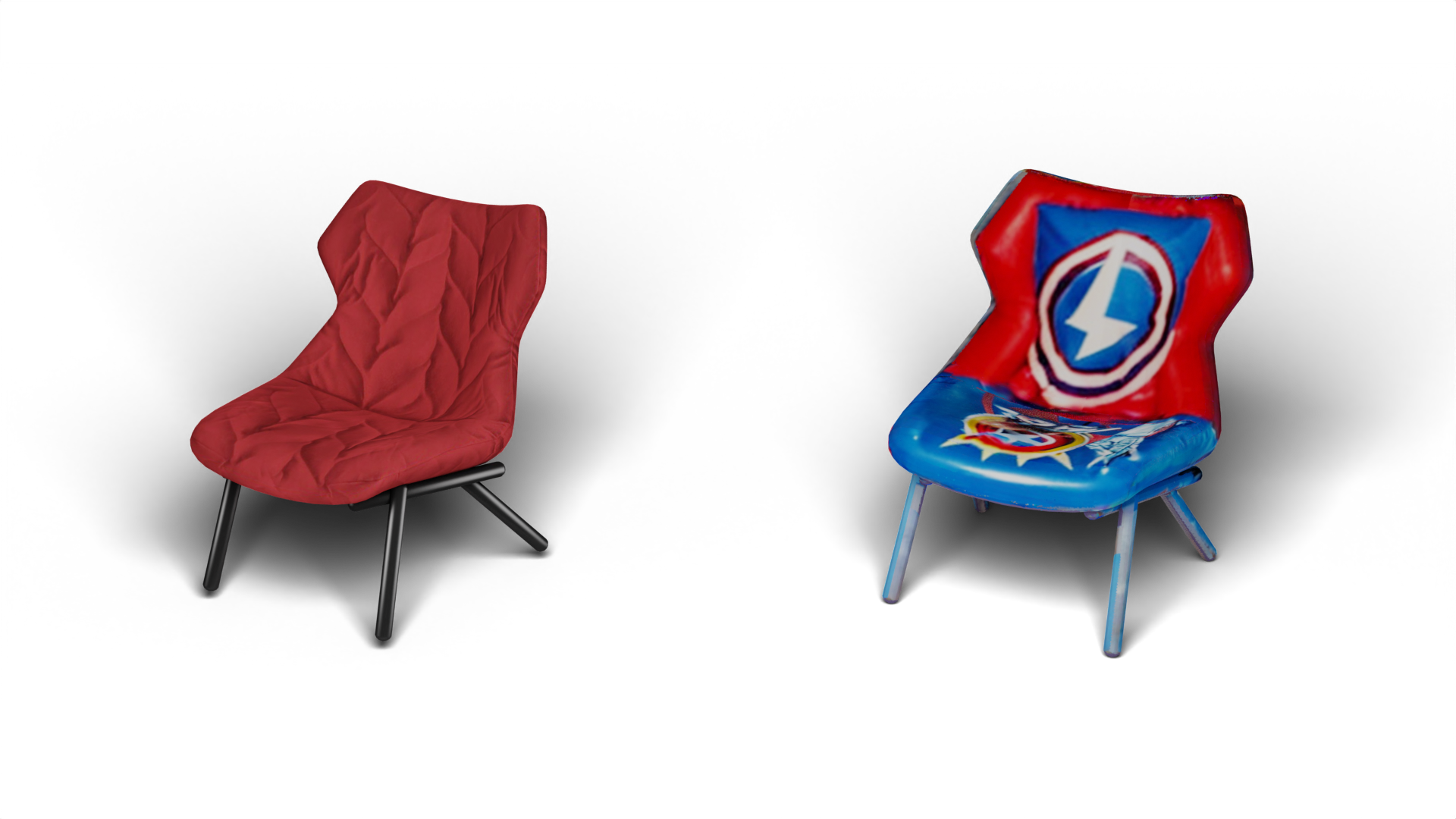}
     \caption{The original Future3D asset retrieved and the same asset retextured by our system for the scene described by ``a marvel-themed bedroom of a five-year old toddler''. The effective texturing prompt created by semantic upsampling is ``chair, in a scene of a marvel-themed bedroom for a five-year-old toddler, red and blue colors. no signs of wear and tear, firm supporting cushions.'' Note  how the semantic adherence to the scene description increases after retexturing!  }
     \label{fig:before_after_texturing}
\end{figure}

Given the ever-growing 3D asset collections, what is the benefit of replacing pre-fabricated textures using the last step of our system? Table \ref{tab:texturing_ablations} reflects what happens to the CLIP-S score (w.r.t. the abstract scene description) when we use the original texture, compared to that of our re-textured objects. This shows that in general, retexturing using the output from semantic upsampling allows an increase in visual similarity with respect to the abstract description of the \emph{whole} scene. An example of this can be seen in Figure \ref{fig:before_after_texturing}.
\begin{table}[t]
\begin{tabular}{ccc|c}
\toprule
Scene Reference & Orig. $\uparrow$ & Retextured $\uparrow$ & \% Improved \\
\midrule
rustic backyard & 0.17 & \textbf{0.20} & 80.00 \\
futuristic teahouse & \textbf{0.21} & 0.19 & 28.00 \\
confucius bedroom & 0.23 & 0.23 & 63.33 \\
alien teagarden & 0.18 & \textbf{0.20} & 70.59 \\
retro arcade & 0.21 & \textbf{0.23} & 66.67 \\
anne frank room & 0.18 & \textbf{0.23} & 85.19 \\
hades cave & 0.20 & \textbf{0.20} & 74.29 \\
shrek home & 0.18 & \textbf{0.25} & 93.55 \\
smurf house & 0.18 & \textbf{0.21} & 85.00 \\
mad scientist restaurant & \textbf{0.19} & 0.18 & 44.74 \\
western saloon & 0.16 & \textbf{0.19} & 81.82 \\
occult cult & 0.20 & 0.20 & 48.28 \\
marvel bedroom & 0.23 & \textbf{0.27} & 84.38 \\
murder room & 0.20 & \textbf{0.21} & 60.00 \\
strawberry church & 0.20 & \textbf{0.22} & 78.26 \\
poseidon living room & 0.20 & \textbf{0.21} & 56.00 \\
north korean classroom & 0.17 & \textbf{0.21} & 85.00 \\
antichrist vatican & 0.18 & \textbf{0.20} & 73.91 \\
romantic restaurant & 0.17 & \textbf{0.20} & 76.60 \\
busy new york street & 0.15 & \textbf{0.16} & 68.75 \\
\bottomrule
\end{tabular}
\caption{The mean CLIP-S scores of generated asset collections w.r.t. their abstract scene description, with (Retextured) and without (Orig.) the retexturing using the semantic shopping lists. ``\% Improve'' indicates the percentage of assets in the collection whose CLIP-S scores increased after retexturing. This shows that the retexturing is a valuable step of the pipeline to return assets that are more aligned with the scene semantics.}
\label{tab:texturing_ablations}
\vspace{-1cm}
\end{table}

\subsection{User study}

The metrics used to evaluate our method thus far are heuristical. To gauge the true value of our system, we conduct a user study composed of 72 human evaluators, across 11 randomly selected scenes in the full list of 20, for a total of 792 annotations. Each human evaluators are first shown two options: (1) a scene rendering composed of assets generated by our system and (2) a scene rendering composed of assets generated by the baseline system (see Section \ref{sec:semantic_autocompletion_ablation}). To decouple semantic alignment of the composed scene from the quality and diversity of the assets themselves, each evaluator is then asked two questions: (1) which arrangement of 3D assets is more accurate/faithful to the scene description? (2) If you were a 3D artist, which group of assets would you use to create a scene that matches the scene description? (Considering diversity, quality ...etc). Evaluators can only select one of the two options for each question. Please see the Appendix for the images shown to the human evaluators.

\begin{table}[t]
\begin{tabular}{ccc|cc}
\toprule
Scene reference & Q1(base) & Q1(our) & Q2(base) & Q2(our) \\
\midrule
poseidon living room  & 25 \% & \textbf{75\%}  & 23.6\% & \textbf{76.4\%} \\
romantic restaurant &  9.7\% & \textbf{90.3\%} & 19.4\% & \textbf{80.6\%} \\
retro arcade & 6.9 \% & \textbf{93.1\%} & 5.6\% & \textbf{94.4\%} \\
anne frank room & 31.9\% & \textbf{68.1\%} & 22.2\% & \textbf{77.8\%} \\
smurf house & 18.1\% & \textbf{81.9\%} & 25\% & \textbf{75\%} \\
murder room & 4.2\% & \textbf{95.8\%} & 9.7\% & \textbf{90.3\%} \\
shrek home & 9.7\% & \textbf{90.3\%} & 12.5\% & \textbf{87.5\% }\\
confucius bedroom & \textbf{63.9\%} & 36.1\%  & \textbf{59.7\%} & 40.3\% \\
marvel bedroom & 16.7\% & \textbf{83.3\%} & 25\% & \textbf{75\%} \\
futuristic teahouse & 48.6\% & \textbf{51.4\%} & 50\% & 50\% \\
western saloon & 12.5\% & \textbf{87.5\%} & 19.4\% & \textbf{80.6\%} \\
\bottomrule
\end{tabular}
\caption{ The percentage of human evaluators who selected each option for the two questions. \textbf{Q1} indicates the first question: ``Which arrangement of 3D assets is more accurate/faithful to the scene description?'' \textbf{Q2} indicates the  second question:  ``If you were a 3D artist, which group of assets would you use to create a scene that matches the scene description? (Considering diversity, quality ...etc)''. Our system (\textbf{our}) is consistently favored for both questions over the baseline (\textbf{base}), except for one scene. This will be expanded upon further in the Appendix.}
\label{tab:user_study}
\vspace{-0.8cm}
\end{table}

A summary of user selections for these two questions for each of the 11 scenes is shown in Table \ref{tab:user_study}. Note that in 10 out of the 11 scenes, the study showed that using the assets generated by our system allows the creator of the scene to better match the semantics of the abstract scene description, compared to assets generated by a version of our system without semantic upsampling. Additionally in 9 out of the 11 scenes, the assets were also considered a better selection for 3D artists for creating similar scenes. This demonstrates the efficacy of semantic upsampling in producing both more diverse and relevant assets, and their ability to constitute more semantically aligned 3D scenes. 

\section{Discussion \& Conclusion}

In this paper, we present a system that leverages the common sense understanding of LLMs, Vision-Language models and Diffusion models to tackle the problem of 3D assets stylization given abstract scene descriptions. Our system uses the key insight that to generate higher quality elements that compose a scene, we can mine the common sense understanding of GPT-3 to semantically upsample the scene semantics based on the abstract scene description using in-context learning. The result is an intermediary representation that is human-readable, editable, and conducive towards higher quality retrieval and texturing of 3D assets. 

As a framework for 3D asset generation, our system offers an easy method to transfer the world knowledge of foundational models extracted from modalities like image and text to identifying, texturing and composing meshes that can be used to construct a scene. As the reasoning, generation and texturing potential of these underlying foundation models improve, so would our system outputs.

We showcase our system in action using diverse language inputs, and show the importance of various aspects of our framework through both quantitative metrics and user studies. In addition, we demonstrate the power and the robustness to our framework afforded by leveraging foundation models for this task in a zero-shot manner.

Although our work makes an important step towards scene synthesis, there are still many open questions to be addressed in future research.
For instance, generating valid scene-layouts in an open-vocabulary and generalizable way remains a challenge. Furthermore, future efforts in inferring more 3D consistent texture maps as well as material properties from generative image models are also valuable. Finally, it would also be useful to develop methods that can adequately generate appropriate backgrounds for asset collections. 

\bibliographystyle{ACM-Reference-Format}
\bibliography{ref}


\begin{thebibliography}{53}


\ifx \showCODEN    \undefined \def \showCODEN     #1{\unskip}     \fi
\ifx \showDOI      \undefined \def \showDOI       #1{#1}\fi
\ifx \showISBNx    \undefined \def \showISBNx     #1{\unskip}     \fi
\ifx \showISBNxiii \undefined \def \showISBNxiii  #1{\unskip}     \fi
\ifx \showISSN     \undefined \def \showISSN      #1{\unskip}     \fi
\ifx \showLCCN     \undefined \def \showLCCN      #1{\unskip}     \fi
\ifx \shownote     \undefined \def \shownote      #1{#1}          \fi
\ifx \showarticletitle \undefined \def \showarticletitle #1{#1}   \fi
\ifx \showURL      \undefined \def \showURL       {\relax}        \fi
\providecommand\bibfield[2]{#2}
\providecommand\bibinfo[2]{#2}
\providecommand\natexlab[1]{#1}
\providecommand\showeprint[2][]{arXiv:#2}

\bibitem[Achlioptas et~al\mbox{.}(2020)]%
        {achlioptas2020referit3d}
\bibfield{author}{\bibinfo{person}{Panos Achlioptas}, \bibinfo{person}{Ahmed
  Abdelreheem}, \bibinfo{person}{Fei Xia}, \bibinfo{person}{Mohamed Elhoseiny},
  {and} \bibinfo{person}{Leonidas Guibas}.} \bibinfo{year}{2020}\natexlab{}.
\newblock \showarticletitle{Referit3d: Neural listeners for fine-grained 3d
  object identification in real-world scenes}. In
  \bibinfo{booktitle}{\emph{Computer Vision--ECCV 2020: 16th European
  Conference, Glasgow, UK, August 23--28, 2020, Proceedings, Part I 16}}.
  Springer, \bibinfo{pages}{422--440}.
\newblock


\bibitem[Achlioptas et~al\mbox{.}(2022)]%
        {changeit3d}
\bibfield{author}{\bibinfo{person}{Panos Achlioptas}, \bibinfo{person}{Ian
  Huang}, \bibinfo{person}{Minhyuk Sung}, \bibinfo{person}{Sergey Tulyakov},
  {and} \bibinfo{person}{Leonidas Guibas}.} \bibinfo{year}{2022}\natexlab{}.
\newblock \showarticletitle{{ChangeIt3D}: Language-Assisted 3D Shape Edits and
  Deformations}.
\newblock \bibinfo{journal}{\emph{\url{https://changeit3d.github.io/}}}
  (\bibinfo{year}{2022}).
\newblock


\bibitem[Brown et~al\mbox{.}(2020)]%
        {brown2020language}
\bibfield{author}{\bibinfo{person}{Tom Brown}, \bibinfo{person}{Benjamin Mann},
  \bibinfo{person}{Nick Ryder}, \bibinfo{person}{Melanie Subbiah},
  \bibinfo{person}{Jared~D Kaplan}, \bibinfo{person}{Prafulla Dhariwal},
  \bibinfo{person}{Arvind Neelakantan}, \bibinfo{person}{Pranav Shyam},
  \bibinfo{person}{Girish Sastry}, \bibinfo{person}{Amanda Askell},
  {et~al\mbox{.}}} \bibinfo{year}{2020}\natexlab{}.
\newblock \showarticletitle{Language models are few-shot learners}.
\newblock \bibinfo{journal}{\emph{Advances in neural information processing
  systems}}  \bibinfo{volume}{33} (\bibinfo{year}{2020}),
  \bibinfo{pages}{1877--1901}.
\newblock


\bibitem[Chang et~al\mbox{.}(2017)]%
        {chang2017matterport3d}
\bibfield{author}{\bibinfo{person}{Angel Chang}, \bibinfo{person}{Angela Dai},
  \bibinfo{person}{Thomas Funkhouser}, \bibinfo{person}{Maciej Halber},
  \bibinfo{person}{Matthias Niessner}, \bibinfo{person}{Manolis Savva},
  \bibinfo{person}{Shuran Song}, \bibinfo{person}{Andy Zeng}, {and}
  \bibinfo{person}{Yinda Zhang}.} \bibinfo{year}{2017}\natexlab{}.
\newblock \showarticletitle{Matterport3d: Learning from rgb-d data in indoor
  environments}.
\newblock \bibinfo{journal}{\emph{arXiv preprint arXiv:1709.06158}}
  (\bibinfo{year}{2017}).
\newblock


\bibitem[Chang et~al\mbox{.}(2015b)]%
        {chang2015scenelex}
\bibfield{author}{\bibinfo{person}{Angel Chang}, \bibinfo{person}{Will Monroe},
  \bibinfo{person}{Manolis Savva}, \bibinfo{person}{Christopher Potts}, {and}
  \bibinfo{person}{Christopher~D Manning}.} \bibinfo{year}{2015}\natexlab{b}.
\newblock \showarticletitle{Text to 3d scene generation with rich lexical
  grounding}.
\newblock \bibinfo{journal}{\emph{arXiv preprint arXiv:1505.06289}}
  (\bibinfo{year}{2015}).
\newblock


\bibitem[Chang et~al\mbox{.}(2015a)]%
        {chang2015shapenet}
\bibfield{author}{\bibinfo{person}{Angel~X Chang}, \bibinfo{person}{Thomas
  Funkhouser}, \bibinfo{person}{Leonidas Guibas}, \bibinfo{person}{Pat
  Hanrahan}, \bibinfo{person}{Qixing Huang}, \bibinfo{person}{Zimo Li},
  \bibinfo{person}{Silvio Savarese}, \bibinfo{person}{Manolis Savva},
  \bibinfo{person}{Shuran Song}, \bibinfo{person}{Hao Su}, {et~al\mbox{.}}}
  \bibinfo{year}{2015}\natexlab{a}.
\newblock \showarticletitle{Shapenet: An information-rich 3d model repository}.
\newblock \bibinfo{journal}{\emph{arXiv preprint arXiv:1512.03012}}
  (\bibinfo{year}{2015}).
\newblock


\bibitem[Chen et~al\mbox{.}(2022)]%
        {chen2022incontext9}
\bibfield{author}{\bibinfo{person}{Mingda Chen}, \bibinfo{person}{Jingfei Du},
  \bibinfo{person}{Ramakanth Pasunuru}, \bibinfo{person}{Todor Mihaylov},
  \bibinfo{person}{Srini Iyer}, \bibinfo{person}{Veselin Stoyanov}, {and}
  \bibinfo{person}{Zornitsa Kozareva}.} \bibinfo{year}{2022}\natexlab{}.
\newblock \showarticletitle{Improving In-Context Few-Shot Learning via
  Self-Supervised Training}.
\newblock \bibinfo{journal}{\emph{arXiv preprint arXiv:2205.01703}}
  (\bibinfo{year}{2022}).
\newblock


\bibitem[Community(2018)]%
        {blender}
\bibfield{author}{\bibinfo{person}{Blender~Online Community}.}
  \bibinfo{year}{2018}\natexlab{}.
\newblock \bibinfo{booktitle}{\emph{Blender - a 3D modelling and rendering
  package}}.
\newblock Blender Foundation, Stichting Blender Foundation, Amsterdam.
\newblock
\urldef\tempurl%
\url{http://www.blender.org}
\showURL{%
\tempurl}


\bibitem[Deitke et~al\mbox{.}(2022)]%
        {deitke2022objaverse}
\bibfield{author}{\bibinfo{person}{Matt Deitke}, \bibinfo{person}{Dustin
  Schwenk}, \bibinfo{person}{Jordi Salvador}, \bibinfo{person}{Luca Weihs},
  \bibinfo{person}{Oscar Michel}, \bibinfo{person}{Eli VanderBilt},
  \bibinfo{person}{Ludwig Schmidt}, \bibinfo{person}{Kiana Ehsani},
  \bibinfo{person}{Aniruddha Kembhavi}, {and} \bibinfo{person}{Ali Farhadi}.}
  \bibinfo{year}{2022}\natexlab{}.
\newblock \showarticletitle{Objaverse: A Universe of Annotated 3D Objects}.
\newblock \bibinfo{journal}{\emph{arXiv preprint arXiv:2212.08051}}
  (\bibinfo{year}{2022}).
\newblock


\bibitem[Dong et~al\mbox{.}(2022)]%
        {dong2022incontext11}
\bibfield{author}{\bibinfo{person}{Qingxiu Dong}, \bibinfo{person}{Lei Li},
  \bibinfo{person}{Damai Dai}, \bibinfo{person}{Ce Zheng},
  \bibinfo{person}{Zhiyong Wu}, \bibinfo{person}{Baobao Chang},
  \bibinfo{person}{Xu Sun}, \bibinfo{person}{Jingjing Xu}, {and}
  \bibinfo{person}{Zhifang Sui}.} \bibinfo{year}{2022}\natexlab{}.
\newblock \showarticletitle{A Survey for In-context Learning}.
\newblock \bibinfo{journal}{\emph{arXiv preprint arXiv:2301.00234}}
  (\bibinfo{year}{2022}).
\newblock


\bibitem[Fridman et~al\mbox{.}(2023)]%
        {fridman2023scenescape}
\bibfield{author}{\bibinfo{person}{Rafail Fridman}, \bibinfo{person}{Amit
  Abecasis}, \bibinfo{person}{Yoni Kasten}, {and} \bibinfo{person}{Tali
  Dekel}.} \bibinfo{year}{2023}\natexlab{}.
\newblock \showarticletitle{Scenescape: Text-driven consistent scene
  generation}.
\newblock \bibinfo{journal}{\emph{arXiv preprint arXiv:2302.01133}}
  (\bibinfo{year}{2023}).
\newblock


\bibitem[Fu et~al\mbox{.}(2021a)]%
        {fu20213dfront}
\bibfield{author}{\bibinfo{person}{Huan Fu}, \bibinfo{person}{Bowen Cai},
  \bibinfo{person}{Lin Gao}, \bibinfo{person}{Ling-Xiao Zhang},
  \bibinfo{person}{Jiaming Wang}, \bibinfo{person}{Cao Li},
  \bibinfo{person}{Qixun Zeng}, \bibinfo{person}{Chengyue Sun},
  \bibinfo{person}{Rongfei Jia}, \bibinfo{person}{Binqiang Zhao},
  {et~al\mbox{.}}} \bibinfo{year}{2021}\natexlab{a}.
\newblock \showarticletitle{3d-front: 3d furnished rooms with layouts and
  semantics}. In \bibinfo{booktitle}{\emph{Proceedings of the IEEE/CVF
  International Conference on Computer Vision}}. \bibinfo{pages}{10933--10942}.
\newblock


\bibitem[Fu et~al\mbox{.}(2021b)]%
        {fu20213d}
\bibfield{author}{\bibinfo{person}{Huan Fu}, \bibinfo{person}{Rongfei Jia},
  \bibinfo{person}{Lin Gao}, \bibinfo{person}{Mingming Gong},
  \bibinfo{person}{Binqiang Zhao}, \bibinfo{person}{Steve Maybank}, {and}
  \bibinfo{person}{Dacheng Tao}.} \bibinfo{year}{2021}\natexlab{b}.
\newblock \showarticletitle{3d-future: 3d furniture shape with texture}.
\newblock \bibinfo{journal}{\emph{International Journal of Computer Vision}}
  (\bibinfo{year}{2021}), \bibinfo{pages}{1--25}.
\newblock


\bibitem[Fu et~al\mbox{.}(2022)]%
        {fu2022shapecrafter}
\bibfield{author}{\bibinfo{person}{Rao Fu}, \bibinfo{person}{Xiao Zhan},
  \bibinfo{person}{Yiwen Chen}, \bibinfo{person}{Daniel Ritchie}, {and}
  \bibinfo{person}{Srinath Sridhar}.} \bibinfo{year}{2022}\natexlab{}.
\newblock \showarticletitle{Shapecrafter: A recursive text-conditioned 3d shape
  generation model}.
\newblock \bibinfo{journal}{\emph{arXiv preprint arXiv:2207.09446}}
  (\bibinfo{year}{2022}).
\newblock


\bibitem[Gao et~al\mbox{.}(2022)]%
        {gao2022get3d}
\bibfield{author}{\bibinfo{person}{Jun Gao}, \bibinfo{person}{Tianchang Shen},
  \bibinfo{person}{Zian Wang}, \bibinfo{person}{Wenzheng Chen},
  \bibinfo{person}{Kangxue Yin}, \bibinfo{person}{Daiqing Li},
  \bibinfo{person}{Or Litany}, \bibinfo{person}{Zan Gojcic}, {and}
  \bibinfo{person}{Sanja Fidler}.} \bibinfo{year}{2022}\natexlab{}.
\newblock \showarticletitle{Get3d: A generative model of high quality 3d
  textured shapes learned from images}.
\newblock \bibinfo{journal}{\emph{Advances In Neural Information Processing
  Systems}}  \bibinfo{volume}{35} (\bibinfo{year}{2022}),
  \bibinfo{pages}{31841--31854}.
\newblock


\bibitem[Ho et~al\mbox{.}(2020)]%
        {ho2020denoising}
\bibfield{author}{\bibinfo{person}{Jonathan Ho}, \bibinfo{person}{Ajay Jain},
  {and} \bibinfo{person}{Pieter Abbeel}.} \bibinfo{year}{2020}\natexlab{}.
\newblock \showarticletitle{Denoising diffusion probabilistic models}.
\newblock \bibinfo{journal}{\emph{Advances in Neural Information Processing
  Systems}}  \bibinfo{volume}{33} (\bibinfo{year}{2020}),
  \bibinfo{pages}{6840--6851}.
\newblock


\bibitem[H{\"o}llein et~al\mbox{.}(2023)]%
        {hoellein2023text2room}
\bibfield{author}{\bibinfo{person}{Lukas H{\"o}llein}, \bibinfo{person}{Ang
  Cao}, \bibinfo{person}{Andrew Owens}, \bibinfo{person}{Justin Johnson}, {and}
  \bibinfo{person}{Matthias Nie{\ss}ner}.} \bibinfo{year}{2023}\natexlab{}.
\newblock \bibinfo{title}{Text2Room: Extracting Textured 3D Meshes from 2D
  Text-to-Image Models}.
\newblock
\newblock


\bibitem[Huang et~al\mbox{.}(2022)]%
        {huang2022ladis}
\bibfield{author}{\bibinfo{person}{Ian Huang}, \bibinfo{person}{Panos
  Achlioptas}, \bibinfo{person}{Tianyi Zhang}, \bibinfo{person}{Sergey
  Tulyakov}, \bibinfo{person}{Minhyuk Sung}, {and} \bibinfo{person}{Leonidas
  Guibas}.} \bibinfo{year}{2022}\natexlab{}.
\newblock \showarticletitle{LADIS: Language disentanglement for 3D shape
  editing}.
\newblock \bibinfo{journal}{\emph{arXiv preprint arXiv:2212.05011}}
  (\bibinfo{year}{2022}).
\newblock


\bibitem[Hui et~al\mbox{.}(2022)]%
        {hui2022neural}
\bibfield{author}{\bibinfo{person}{Ka-Hei Hui}, \bibinfo{person}{Ruihui Li},
  \bibinfo{person}{Jingyu Hu}, {and} \bibinfo{person}{Chi-Wing Fu}.}
  \bibinfo{year}{2022}\natexlab{}.
\newblock \showarticletitle{Neural template: Topology-aware reconstruction and
  disentangled generation of 3d meshes}. In
  \bibinfo{booktitle}{\emph{Proceedings of the IEEE/CVF Conference on Computer
  Vision and Pattern Recognition}}. \bibinfo{pages}{18572--18582}.
\newblock


\bibitem[Ilinykh et~al\mbox{.}(2019)]%
        {ilinykh2019tell}
\bibfield{author}{\bibinfo{person}{Nikolai Ilinykh}, \bibinfo{person}{Sina
  Zarrie{\ss}}, {and} \bibinfo{person}{David Schlangen}.}
  \bibinfo{year}{2019}\natexlab{}.
\newblock \showarticletitle{Tell me more: A dataset of visual scene description
  sequences}. In \bibinfo{booktitle}{\emph{Proceedings of the 12th
  international conference on natural language generation}}.
  \bibinfo{pages}{152--157}.
\newblock


\bibitem[Jain et~al\mbox{.}(2022)]%
        {jain2022dreamfields}
\bibfield{author}{\bibinfo{person}{Ajay Jain}, \bibinfo{person}{Ben
  Mildenhall}, \bibinfo{person}{Jonathan~T Barron}, \bibinfo{person}{Pieter
  Abbeel}, {and} \bibinfo{person}{Ben Poole}.} \bibinfo{year}{2022}\natexlab{}.
\newblock \showarticletitle{Zero-shot text-guided object generation with dream
  fields}. In \bibinfo{booktitle}{\emph{Proceedings of the IEEE/CVF Conference
  on Computer Vision and Pattern Recognition}}. \bibinfo{pages}{867--876}.
\newblock


\bibitem[Jun and Nichol(2023)]%
        {jun2023shapE}
\bibfield{author}{\bibinfo{person}{Heewoo Jun} {and} \bibinfo{person}{Alex
  Nichol}.} \bibinfo{year}{2023}\natexlab{}.
\newblock \showarticletitle{Shap-E: Generating Conditional 3D Implicit
  Functions}.
\newblock \bibinfo{journal}{\emph{arXiv preprint arXiv:2305.02463}}
  (\bibinfo{year}{2023}).
\newblock


\bibitem[Lin et~al\mbox{.}(2022)]%
        {lin2022magic3d}
\bibfield{author}{\bibinfo{person}{Chen-Hsuan Lin}, \bibinfo{person}{Jun Gao},
  \bibinfo{person}{Luming Tang}, \bibinfo{person}{Towaki Takikawa},
  \bibinfo{person}{Xiaohui Zeng}, \bibinfo{person}{Xun Huang},
  \bibinfo{person}{Karsten Kreis}, \bibinfo{person}{Sanja Fidler},
  \bibinfo{person}{Ming-Yu Liu}, {and} \bibinfo{person}{Tsung-Yi Lin}.}
  \bibinfo{year}{2022}\natexlab{}.
\newblock \showarticletitle{Magic3D: High-Resolution Text-to-3D Content
  Creation}.
\newblock \bibinfo{journal}{\emph{arXiv preprint arXiv:2211.10440}}
  (\bibinfo{year}{2022}).
\newblock


\bibitem[Ma et~al\mbox{.}(2018)]%
        {ma2018languagescenedatabases}
\bibfield{author}{\bibinfo{person}{Rui Ma}, \bibinfo{person}{Akshay~Gadi
  Patil}, \bibinfo{person}{Matthew Fisher}, \bibinfo{person}{Manyi Li},
  \bibinfo{person}{S{\"o}ren Pirk}, \bibinfo{person}{Binh-Son Hua},
  \bibinfo{person}{Sai-Kit Yeung}, \bibinfo{person}{Xin Tong},
  \bibinfo{person}{Leonidas Guibas}, {and} \bibinfo{person}{Hao Zhang}.}
  \bibinfo{year}{2018}\natexlab{}.
\newblock \showarticletitle{Language-driven synthesis of 3D scenes from scene
  databases}.
\newblock \bibinfo{journal}{\emph{ACM Transactions on Graphics (TOG)}}
  \bibinfo{volume}{37}, \bibinfo{number}{6} (\bibinfo{year}{2018}),
  \bibinfo{pages}{1--16}.
\newblock


\bibitem[Michel et~al\mbox{.}(2022)]%
        {michel2022text2mesh}
\bibfield{author}{\bibinfo{person}{Oscar Michel}, \bibinfo{person}{Roi Bar-On},
  \bibinfo{person}{Richard Liu}, \bibinfo{person}{Sagie Benaim}, {and}
  \bibinfo{person}{Rana Hanocka}.} \bibinfo{year}{2022}\natexlab{}.
\newblock \showarticletitle{Text2mesh: Text-driven neural stylization for
  meshes}. In \bibinfo{booktitle}{\emph{Proceedings of the IEEE/CVF Conference
  on Computer Vision and Pattern Recognition}}. \bibinfo{pages}{13492--13502}.
\newblock


\bibitem[Min et~al\mbox{.}(2021)]%
        {min2021incontext10}
\bibfield{author}{\bibinfo{person}{Sewon Min}, \bibinfo{person}{Mike Lewis},
  \bibinfo{person}{Luke Zettlemoyer}, {and} \bibinfo{person}{Hannaneh
  Hajishirzi}.} \bibinfo{year}{2021}\natexlab{}.
\newblock \showarticletitle{Metaicl: Learning to learn in context}.
\newblock \bibinfo{journal}{\emph{arXiv preprint arXiv:2110.15943}}
  (\bibinfo{year}{2021}).
\newblock


\bibitem[Nichol et~al\mbox{.}(2022)]%
        {nichol2022pointe}
\bibfield{author}{\bibinfo{person}{Alex Nichol}, \bibinfo{person}{Heewoo Jun},
  \bibinfo{person}{Prafulla Dhariwal}, \bibinfo{person}{Pamela Mishkin}, {and}
  \bibinfo{person}{Mark Chen}.} \bibinfo{year}{2022}\natexlab{}.
\newblock \showarticletitle{Point-E: A System for Generating 3D Point Clouds
  from Complex Prompts}.
\newblock \bibinfo{journal}{\emph{arXiv preprint arXiv:2212.08751}}
  (\bibinfo{year}{2022}).
\newblock


\bibitem[Paschalidou et~al\mbox{.}(2021)]%
        {paschalidou2021atiss}
\bibfield{author}{\bibinfo{person}{Despoina Paschalidou},
  \bibinfo{person}{Amlan Kar}, \bibinfo{person}{Maria Shugrina},
  \bibinfo{person}{Karsten Kreis}, \bibinfo{person}{Andreas Geiger}, {and}
  \bibinfo{person}{Sanja Fidler}.} \bibinfo{year}{2021}\natexlab{}.
\newblock \showarticletitle{Atiss: Autoregressive transformers for indoor scene
  synthesis}.
\newblock \bibinfo{journal}{\emph{Advances in Neural Information Processing
  Systems}}  \bibinfo{volume}{34} (\bibinfo{year}{2021}),
  \bibinfo{pages}{12013--12026}.
\newblock


\bibitem[Po and Wetzstein(2023)]%
        {po2023compositional}
\bibfield{author}{\bibinfo{person}{Ryan Po} {and} \bibinfo{person}{Gordon
  Wetzstein}.} \bibinfo{year}{2023}\natexlab{}.
\newblock \showarticletitle{Compositional 3D Scene Generation using Locally
  Conditioned Diffusion}.
\newblock \bibinfo{journal}{\emph{arXiv preprint arXiv:2303.12218}}
  (\bibinfo{year}{2023}).
\newblock


\bibitem[Poole et~al\mbox{.}(2022)]%
        {poole2022dreamfusion}
\bibfield{author}{\bibinfo{person}{Ben Poole}, \bibinfo{person}{Ajay Jain},
  \bibinfo{person}{Jonathan~T Barron}, {and} \bibinfo{person}{Ben Mildenhall}.}
  \bibinfo{year}{2022}\natexlab{}.
\newblock \showarticletitle{Dreamfusion: Text-to-3d using 2d diffusion}.
\newblock \bibinfo{journal}{\emph{arXiv preprint arXiv:2209.14988}}
  (\bibinfo{year}{2022}).
\newblock


\bibitem[Radford et~al\mbox{.}(2021)]%
        {radford2021learning}
\bibfield{author}{\bibinfo{person}{Alec Radford}, \bibinfo{person}{Jong~Wook
  Kim}, \bibinfo{person}{Chris Hallacy}, \bibinfo{person}{Aditya Ramesh},
  \bibinfo{person}{Gabriel Goh}, \bibinfo{person}{Sandhini Agarwal},
  \bibinfo{person}{Girish Sastry}, \bibinfo{person}{Amanda Askell},
  \bibinfo{person}{Pamela Mishkin}, \bibinfo{person}{Jack Clark},
  {et~al\mbox{.}}} \bibinfo{year}{2021}\natexlab{}.
\newblock \showarticletitle{Learning transferable visual models from natural
  language supervision}. In \bibinfo{booktitle}{\emph{International conference
  on machine learning}}. PMLR, \bibinfo{pages}{8748--8763}.
\newblock


\bibitem[Richardson et~al\mbox{.}(2023)]%
        {richardson2023texture}
\bibfield{author}{\bibinfo{person}{Elad Richardson}, \bibinfo{person}{Gal
  Metzer}, \bibinfo{person}{Yuval Alaluf}, \bibinfo{person}{Raja Giryes}, {and}
  \bibinfo{person}{Daniel Cohen-Or}.} \bibinfo{year}{2023}\natexlab{}.
\newblock \showarticletitle{TEXTure: Text-Guided Texturing of 3D Shapes}.
\newblock \bibinfo{journal}{\emph{arXiv preprint arXiv:2302.01721}}
  (\bibinfo{year}{2023}).
\newblock


\bibitem[Ritchie et~al\mbox{.}(2019)]%
        {ritchie2019fast}
\bibfield{author}{\bibinfo{person}{Daniel Ritchie}, \bibinfo{person}{Kai Wang},
  {and} \bibinfo{person}{Yu-an Lin}.} \bibinfo{year}{2019}\natexlab{}.
\newblock \showarticletitle{Fast and flexible indoor scene synthesis via deep
  convolutional generative models}. In \bibinfo{booktitle}{\emph{Proceedings of
  the IEEE/CVF Conference on Computer Vision and Pattern Recognition}}.
  \bibinfo{pages}{6182--6190}.
\newblock


\bibitem[Roberts et~al\mbox{.}(2021)]%
        {roberts2021hypersim}
\bibfield{author}{\bibinfo{person}{Mike Roberts}, \bibinfo{person}{Jason
  Ramapuram}, \bibinfo{person}{Anurag Ranjan}, \bibinfo{person}{Atulit Kumar},
  \bibinfo{person}{Miguel~Angel Bautista}, \bibinfo{person}{Nathan Paczan},
  \bibinfo{person}{Russ Webb}, {and} \bibinfo{person}{Joshua~M. Susskind}.}
  \bibinfo{year}{2021}\natexlab{}.
\newblock \showarticletitle{{Hypersim}: {A} Photorealistic Synthetic Dataset
  for Holistic Indoor Scene Understanding}. In
  \bibinfo{booktitle}{\emph{International Conference on Computer Vision (ICCV)
  2021}}.
\newblock


\bibitem[Rombach et~al\mbox{.}(2022)]%
        {rombach2022high}
\bibfield{author}{\bibinfo{person}{Robin Rombach}, \bibinfo{person}{Andreas
  Blattmann}, \bibinfo{person}{Dominik Lorenz}, \bibinfo{person}{Patrick
  Esser}, {and} \bibinfo{person}{Bj{\"o}rn Ommer}.}
  \bibinfo{year}{2022}\natexlab{}.
\newblock \showarticletitle{High-resolution image synthesis with latent
  diffusion models}. In \bibinfo{booktitle}{\emph{Proceedings of the IEEE/CVF
  Conference on Computer Vision and Pattern Recognition}}.
  \bibinfo{pages}{10684--10695}.
\newblock


\bibitem[Rubin et~al\mbox{.}(2021)]%
        {rubin2021incontext2}
\bibfield{author}{\bibinfo{person}{Ohad Rubin}, \bibinfo{person}{Jonathan
  Herzig}, {and} \bibinfo{person}{Jonathan Berant}.}
  \bibinfo{year}{2021}\natexlab{}.
\newblock \showarticletitle{Learning to retrieve prompts for in-context
  learning}.
\newblock \bibinfo{journal}{\emph{arXiv preprint arXiv:2112.08633}}
  (\bibinfo{year}{2021}).
\newblock


\bibitem[Saharia et~al\mbox{.}(2022)]%
        {saharia2022imagen}
\bibfield{author}{\bibinfo{person}{Chitwan Saharia}, \bibinfo{person}{William
  Chan}, \bibinfo{person}{Saurabh Saxena}, \bibinfo{person}{Lala Li},
  \bibinfo{person}{Jay Whang}, \bibinfo{person}{Emily~L Denton},
  \bibinfo{person}{Kamyar Ghasemipour}, \bibinfo{person}{Raphael
  Gontijo~Lopes}, \bibinfo{person}{Burcu Karagol~Ayan}, \bibinfo{person}{Tim
  Salimans}, {et~al\mbox{.}}} \bibinfo{year}{2022}\natexlab{}.
\newblock \showarticletitle{Photorealistic text-to-image diffusion models with
  deep language understanding}.
\newblock \bibinfo{journal}{\emph{Advances in Neural Information Processing
  Systems}}  \bibinfo{volume}{35} (\bibinfo{year}{2022}),
  \bibinfo{pages}{36479--36494}.
\newblock


\bibitem[Sanghi et~al\mbox{.}(2022)]%
        {sanghi2022clipforge}
\bibfield{author}{\bibinfo{person}{Aditya Sanghi}, \bibinfo{person}{Hang Chu},
  \bibinfo{person}{Joseph~G Lambourne}, \bibinfo{person}{Ye Wang},
  \bibinfo{person}{Chin-Yi Cheng}, \bibinfo{person}{Marco Fumero}, {and}
  \bibinfo{person}{Kamal~Rahimi Malekshan}.} \bibinfo{year}{2022}\natexlab{}.
\newblock \showarticletitle{Clip-forge: Towards zero-shot text-to-shape
  generation}. In \bibinfo{booktitle}{\emph{Proceedings of the IEEE/CVF
  Conference on Computer Vision and Pattern Recognition}}.
  \bibinfo{pages}{18603--18613}.
\newblock


\bibitem[Selvaraju et~al\mbox{.}(2021)]%
        {selvaraju2021buildingnet}
\bibfield{author}{\bibinfo{person}{Pratheba Selvaraju},
  \bibinfo{person}{Mohamed Nabail}, \bibinfo{person}{Marios Loizou},
  \bibinfo{person}{Maria Maslioukova}, \bibinfo{person}{Melinos Averkiou},
  \bibinfo{person}{Andreas Andreou}, \bibinfo{person}{Siddhartha Chaudhuri},
  {and} \bibinfo{person}{Evangelos Kalogerakis}.}
  \bibinfo{year}{2021}\natexlab{}.
\newblock \showarticletitle{BuildingNet: Learning to label 3D buildings}. In
  \bibinfo{booktitle}{\emph{Proceedings of the IEEE/CVF International
  Conference on Computer Vision}}. \bibinfo{pages}{10397--10407}.
\newblock


\bibitem[Shin et~al\mbox{.}(2022)]%
        {shin2022incontext7}
\bibfield{author}{\bibinfo{person}{Seongjin Shin}, \bibinfo{person}{Sang-Woo
  Lee}, \bibinfo{person}{Hwijeen Ahn}, \bibinfo{person}{Sungdong Kim},
  \bibinfo{person}{HyoungSeok Kim}, \bibinfo{person}{Boseop Kim},
  \bibinfo{person}{Kyunghyun Cho}, \bibinfo{person}{Gichang Lee},
  \bibinfo{person}{Woomyoung Park}, \bibinfo{person}{Jung-Woo Ha},
  {et~al\mbox{.}}} \bibinfo{year}{2022}\natexlab{}.
\newblock \showarticletitle{On the effect of pretraining corpora on in-context
  learning by a large-scale language model}.
\newblock \bibinfo{journal}{\emph{arXiv preprint arXiv:2204.13509}}
  (\bibinfo{year}{2022}).
\newblock


\bibitem[Song et~al\mbox{.}(2015)]%
        {song2015sunrgbd}
\bibfield{author}{\bibinfo{person}{Shuran Song}, \bibinfo{person}{Samuel~P
  Lichtenberg}, {and} \bibinfo{person}{Jianxiong Xiao}.}
  \bibinfo{year}{2015}\natexlab{}.
\newblock \showarticletitle{Sun rgb-d: A rgb-d scene understanding benchmark
  suite}. In \bibinfo{booktitle}{\emph{Proceedings of the IEEE conference on
  computer vision and pattern recognition}}. \bibinfo{pages}{567--576}.
\newblock


\bibitem[Song et~al\mbox{.}(2017)]%
        {song2017suncg}
\bibfield{author}{\bibinfo{person}{Shuran Song}, \bibinfo{person}{Fisher Yu},
  \bibinfo{person}{Andy Zeng}, \bibinfo{person}{Angel~X Chang},
  \bibinfo{person}{Manolis Savva}, {and} \bibinfo{person}{Thomas Funkhouser}.}
  \bibinfo{year}{2017}\natexlab{}.
\newblock \showarticletitle{Semantic scene completion from a single depth
  image}. In \bibinfo{booktitle}{\emph{Proceedings of the IEEE conference on
  computer vision and pattern recognition}}. \bibinfo{pages}{1746--1754}.
\newblock


\bibitem[Wang et~al\mbox{.}(2022)]%
        {wang2022incontext4}
\bibfield{author}{\bibinfo{person}{Boshi Wang}, \bibinfo{person}{Xiang Deng},
  {and} \bibinfo{person}{Huan Sun}.} \bibinfo{year}{2022}\natexlab{}.
\newblock \showarticletitle{Iteratively prompt pre-trained language models for
  chain of thought}. In \bibinfo{booktitle}{\emph{Proceedings of the 2022
  Conference on Empirical Methods in Natural Language Processing}}.
  \bibinfo{pages}{2714--2730}.
\newblock


\bibitem[Wang et~al\mbox{.}(2019)]%
        {wang2019planit}
\bibfield{author}{\bibinfo{person}{Kai Wang}, \bibinfo{person}{Yu-An Lin},
  \bibinfo{person}{Ben Weissmann}, \bibinfo{person}{Manolis Savva},
  \bibinfo{person}{Angel~X Chang}, {and} \bibinfo{person}{Daniel Ritchie}.}
  \bibinfo{year}{2019}\natexlab{}.
\newblock \showarticletitle{Planit: Planning and instantiating indoor scenes
  with relation graph and spatial prior networks}.
\newblock \bibinfo{journal}{\emph{ACM Transactions on Graphics (TOG)}}
  \bibinfo{volume}{38}, \bibinfo{number}{4} (\bibinfo{year}{2019}),
  \bibinfo{pages}{1--15}.
\newblock


\bibitem[Wang et~al\mbox{.}(2021)]%
        {wang2021sceneformer}
\bibfield{author}{\bibinfo{person}{Xinpeng Wang}, \bibinfo{person}{Chandan
  Yeshwanth}, {and} \bibinfo{person}{Matthias Nie{\ss}ner}.}
  \bibinfo{year}{2021}\natexlab{}.
\newblock \showarticletitle{Sceneformer: Indoor scene generation with
  transformers}. In \bibinfo{booktitle}{\emph{2021 International Conference on
  3D Vision (3DV)}}. IEEE, \bibinfo{pages}{106--115}.
\newblock


\bibitem[Wei et~al\mbox{.}(2021)]%
        {wei2021incontext1}
\bibfield{author}{\bibinfo{person}{Jason Wei}, \bibinfo{person}{Maarten Bosma},
  \bibinfo{person}{Vincent~Y Zhao}, \bibinfo{person}{Kelvin Guu},
  \bibinfo{person}{Adams~Wei Yu}, \bibinfo{person}{Brian Lester},
  \bibinfo{person}{Nan Du}, \bibinfo{person}{Andrew~M Dai}, {and}
  \bibinfo{person}{Quoc~V Le}.} \bibinfo{year}{2021}\natexlab{}.
\newblock \showarticletitle{Finetuned language models are zero-shot learners}.
\newblock \bibinfo{journal}{\emph{arXiv preprint arXiv:2109.01652}}
  (\bibinfo{year}{2021}).
\newblock


\bibitem[Wei et~al\mbox{.}(2022)]%
        {wei2022incontext6}
\bibfield{author}{\bibinfo{person}{Jason Wei}, \bibinfo{person}{Xuezhi Wang},
  \bibinfo{person}{Dale Schuurmans}, \bibinfo{person}{Maarten Bosma},
  \bibinfo{person}{Ed Chi}, \bibinfo{person}{Quoc Le}, {and}
  \bibinfo{person}{Denny Zhou}.} \bibinfo{year}{2022}\natexlab{}.
\newblock \showarticletitle{Chain of thought prompting elicits reasoning in
  large language models}.
\newblock \bibinfo{journal}{\emph{arXiv preprint arXiv:2201.11903}}
  (\bibinfo{year}{2022}).
\newblock


\bibitem[Xu et~al\mbox{.}(2022)]%
        {xu2022dream3d}
\bibfield{author}{\bibinfo{person}{Jiale Xu}, \bibinfo{person}{Xintao Wang},
  \bibinfo{person}{Weihao Cheng}, \bibinfo{person}{Yan-Pei Cao},
  \bibinfo{person}{Ying Shan}, \bibinfo{person}{Xiaohu Qie}, {and}
  \bibinfo{person}{Shenghua Gao}.} \bibinfo{year}{2022}\natexlab{}.
\newblock \showarticletitle{Dream3D: Zero-Shot Text-to-3D Synthesis Using 3D
  Shape Prior and Text-to-Image Diffusion Models}.
\newblock \bibinfo{journal}{\emph{arXiv preprint arXiv:2212.14704}}
  (\bibinfo{year}{2022}).
\newblock


\bibitem[Yuksekgonul et~al\mbox{.}(2022)]%
        {yuksekgonul2022bagofwords}
\bibfield{author}{\bibinfo{person}{Mert Yuksekgonul}, \bibinfo{person}{Federico
  Bianchi}, \bibinfo{person}{Pratyusha Kalluri}, \bibinfo{person}{Dan
  Jurafsky}, {and} \bibinfo{person}{James Zou}.}
  \bibinfo{year}{2022}\natexlab{}.
\newblock \showarticletitle{When and why vision-language models behave like
  bags-of-words, and what to do about it?}
\newblock \bibinfo{journal}{\emph{arXiv e-prints}} (\bibinfo{year}{2022}),
  \bibinfo{pages}{arXiv--2210}.
\newblock


\bibitem[Zhang et~al\mbox{.}(2023)]%
        {zhang2023text2nerf}
\bibfield{author}{\bibinfo{person}{Jingbo Zhang}, \bibinfo{person}{Xiaoyu Li},
  \bibinfo{person}{Ziyu Wan}, \bibinfo{person}{Can Wang}, {and}
  \bibinfo{person}{Jing Liao}.} \bibinfo{year}{2023}\natexlab{}.
\newblock \showarticletitle{Text2NeRF: Text-Driven 3D Scene Generation with
  Neural Radiance Fields}.
\newblock \bibinfo{journal}{\emph{arXiv preprint arXiv:2305.11588}}
  (\bibinfo{year}{2023}).
\newblock


\bibitem[Zhang et~al\mbox{.}(2022)]%
        {zhang2022incontext3}
\bibfield{author}{\bibinfo{person}{Zhuosheng Zhang}, \bibinfo{person}{Aston
  Zhang}, \bibinfo{person}{Mu Li}, {and} \bibinfo{person}{Alex Smola}.}
  \bibinfo{year}{2022}\natexlab{}.
\newblock \showarticletitle{Automatic chain of thought prompting in large
  language models}.
\newblock \bibinfo{journal}{\emph{arXiv preprint arXiv:2210.03493}}
  (\bibinfo{year}{2022}).
\newblock


\bibitem[Zhao et~al\mbox{.}(2021)]%
        {zhao2021incontext8}
\bibfield{author}{\bibinfo{person}{Zihao Zhao}, \bibinfo{person}{Eric Wallace},
  \bibinfo{person}{Shi Feng}, \bibinfo{person}{Dan Klein}, {and}
  \bibinfo{person}{Sameer Singh}.} \bibinfo{year}{2021}\natexlab{}.
\newblock \showarticletitle{Calibrate before use: Improving few-shot
  performance of language models}. In \bibinfo{booktitle}{\emph{International
  Conference on Machine Learning}}. PMLR, \bibinfo{pages}{12697--12706}.
\newblock


\bibitem[Zhou et~al\mbox{.}(2022)]%
        {zhou2022incontext5}
\bibfield{author}{\bibinfo{person}{Denny Zhou}, \bibinfo{person}{Nathanael
  Sch{\"a}rli}, \bibinfo{person}{Le Hou}, \bibinfo{person}{Jason Wei},
  \bibinfo{person}{Nathan Scales}, \bibinfo{person}{Xuezhi Wang},
  \bibinfo{person}{Dale Schuurmans}, \bibinfo{person}{Olivier Bousquet},
  \bibinfo{person}{Quoc Le}, {and} \bibinfo{person}{Ed Chi}.}
  \bibinfo{year}{2022}\natexlab{}.
\newblock \showarticletitle{Least-to-most prompting enables complex reasoning
  in large language models}.
\newblock \bibinfo{journal}{\emph{arXiv preprint arXiv:2205.10625}}
  (\bibinfo{year}{2022}).
\newblock


\end{thebibliography}

\pagebreak

\pagebreak

\clearpage
\begin{appendices}

\section{Templates for Semantic Upsampling} 

The templates used during the semantic upsampling stage of the system to perform in-context learning with GPT-3 are manually created. We create 4 templates, used different combinations during semantic upsampling phase as the model's focus moves down the scene hierarchy. 

The first template is used to extract ``anchor'' objects of a scene:

\noindent\begin{mdframed}
Here we are building a 3D scene of a french restaurant. At each step, we are not adding more than 8 assets in total into the scene. 

First, we place the most important assets (e.g. furnitures, bigger objects) and use those as our anchors. Here is a list of them:

* Tables : 1 

* Chairs : 4
 
* Bar : 1
 
* Bar stools : 2
\end{mdframed}

The second template is used to progress \emph{down} the asset hierarchy towards the more peripheral (and sometimes more decorative) assets.

\noindent\begin{mdframed}
Next we enhance the scene with more assets, in relation to the anchor objects. 
In relation to the `table`, here is the list of assets we add:

* Tablecloth : 1

* Plates : 4
 
* Silverware : 4 
 
* Wine glasses : 2
\end{mdframed}

At each level of the hierarchy we can generate appearance attributes by conditioning the LLM input with this template:

\noindent\begin{mdframed}
Suppose we want to create a shopping list for the items we need to create  the above scene of a fancy french restaurant. It would look like, being specific about the brand and the visual properties:

 * Table : country style farmhouse table, oakwood and dark brown.
 
 * Chairs : provincial style chairs, upholstered in ivory velvet. 

 * Bar : Traditional style bar counter, white marble, gold accents on the corners.

 * Bar stools : provincial style bar stools, upholstered in ivory velvet, golden accents on corners
\end{mdframed}

As well as their physical condition, by conditioning using this template:

\noindent\begin{mdframed}
Describe the physical condition of these items in a scene of a fancy french restaurant:

* Table :  smooth, polished finish.

* Chairs : slight signs of wear on the sides.

* Bar : slight signs of wear.

* Bar stools : slight wear on the rattan seats.
\end{mdframed}

It is possible that using only a small set of templates based on a single indoor scene may limit the LLM's ability to perform semantic upsampling for outdoor scenes. We see such behavior in Figure \ref{fig:busy_new_york_street_ours}, a failure case of our system. Future works should consider how to better incorporate a more diverse set of templates to improve the system's generalizability to outdoor scenes and other scenes.

\section{Hierarchy of Semantic Shopping Lists}

During the semantic upsampling step of our system, we condition GPT-3 to hallucinate semantic detail in a hierarchical fashion; that is, it first starts out by generating details of ``anchor'' objects (i.e. the key objects within the scene) before recursively generating details of ``peripheral objects''. A natural question is, does this method naturally yield meaningful hierarchies in object groupings within the scene?

\begin{figure*}
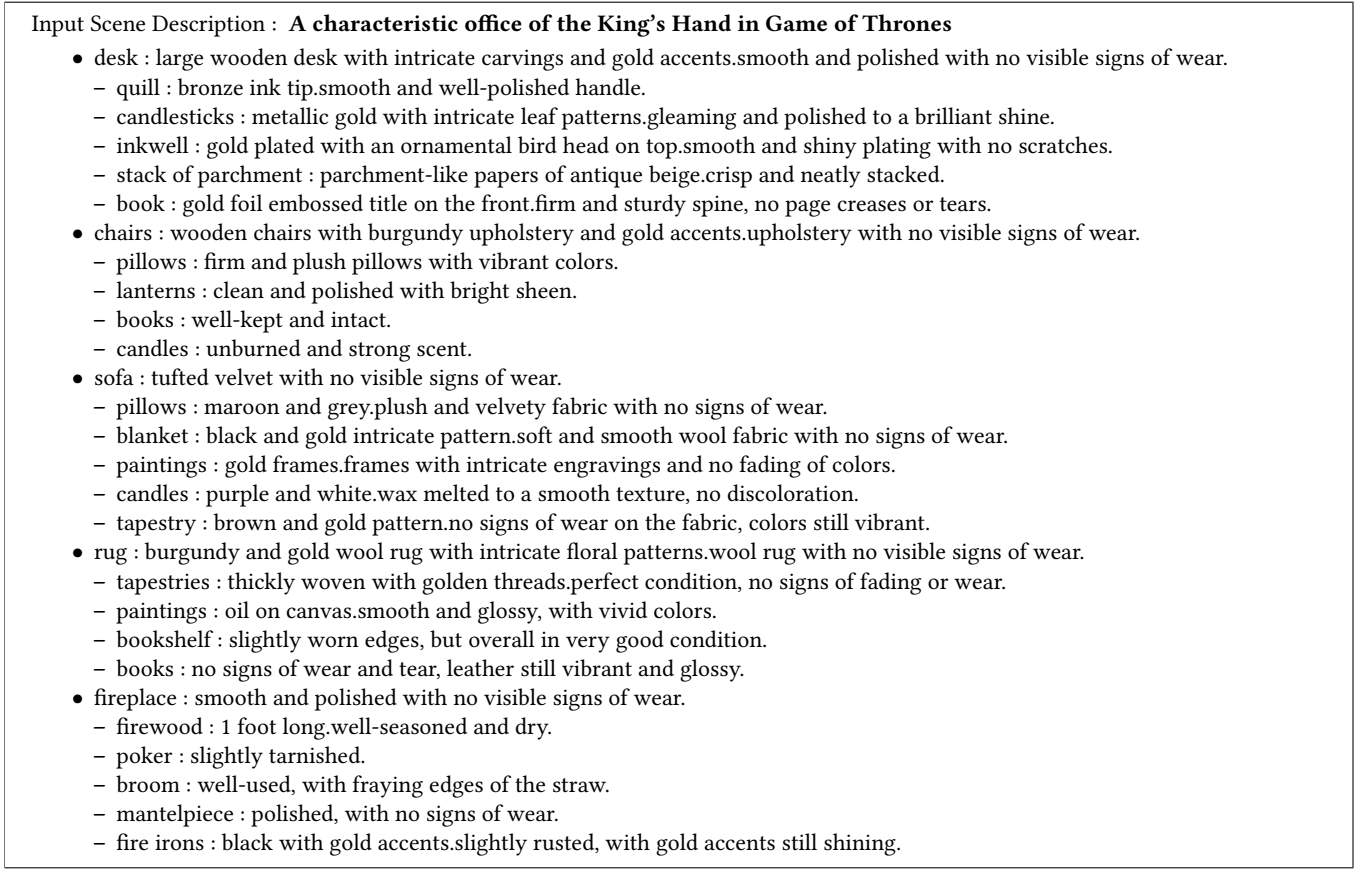

    \centering

\noindent\begin{mdframed}
Input Scene Description : \textbf { A characteristic office of the King's Hand in Game of Thrones}

\begin{itemize}

\item desk : large wooden desk with intricate carvings and gold accents.smooth and polished with no visible signs of wear.
\begin{itemize}
    \item quill : bronze ink tip.smooth and well-polished handle.
    \item candlesticks : metallic gold with intricate leaf patterns.gleaming and polished to a brilliant shine.
    \item inkwell : gold plated with an ornamental bird head on top.smooth and shiny plating with no scratches.
    \item stack of parchment : parchment-like papers of antique beige.crisp and neatly stacked.
    \item book : gold foil embossed title on the front.firm and sturdy spine, no page creases or tears.
\end{itemize}
\item chairs : wooden chairs with burgundy upholstery and gold accents.upholstery with no visible signs of wear.
\begin{itemize}
    \item pillows : firm and plush pillows with vibrant colors.
    \item lanterns : clean and polished with bright sheen.
    \item books : well-kept and intact.
    \item candles : unburned and strong scent.
\end{itemize}
\item sofa : tufted velvet with no visible signs of wear.
\begin{itemize}
    \item pillows : maroon and grey.plush and velvety fabric with no signs of wear.
    \item blanket : black and gold intricate pattern.soft and smooth wool fabric with no signs of wear.
    \item paintings : gold frames.frames with intricate engravings and no fading of colors.
    \item candles : purple and white.wax melted to a smooth texture, no discoloration.
    \item tapestry : brown and gold pattern.no signs of wear on the fabric, colors still vibrant.
\end{itemize}
\item rug : burgundy and gold wool rug with intricate floral patterns.wool rug with no visible signs of wear.
\begin{itemize}
    \item tapestries : thickly woven with golden threads.perfect condition, no signs of fading or wear.
    \item paintings : oil on canvas.smooth and glossy, with vivid colors.
    \item bookshelf : slightly worn edges, but overall in very good condition.
    \item books : no signs of wear and tear, leather still vibrant and glossy.
\end{itemize}
\item fireplace : smooth and polished with no visible signs of wear.
\begin{itemize}
    \item firewood : 1 foot long.well-seasoned and dry.
    \item poker : slightly tarnished.
    \item broom : well-used, with fraying edges of the straw.
    \item mantelpiece : polished, with no signs of wear.
    \item fire irons : black with gold accents.slightly rusted, with gold accents still shining.
\end{itemize}
\end{itemize}

\end{mdframed}
    \caption{The \textit{hierarchical} version of the semantic shopping list given by semantic upsampling, for the scene description of ``a characteristic office of the King's Hand in Game of Thrones.'' We use only a single level of recursion here (i.e. max depth of the semantic shopping list tree is 1). Notice how for certain anchor objects like the table and fireplace, the children grouped underneath them are plausible. However, for anchor objects that are often stand-alone like chairs, the peripheral objects are plausible object categories found \textit{near} the anchor object.
    }
    \label{fig:kingshand_hierarchical}
\end{figure*}

Figure \ref{fig:kingshand_hierarchical} shows the resultant hierarchy of object categories for a single scene. We can see that at times, GPT-3 doesn't output peripheral objects that form natural groups with the anchor objects, and this varies based on the class of the anchor object. For objects that typically have objects placed on, inside or around them, GPT-3 is typically able to capture this regularity (see the peripheral objects around the Table in Figure \ref{fig:kingshand_hierarchical}, for example).

Generating more semantically meaningful groupings of objects using LLM's is a challenging task for future works, and can lay the foundation for methods that predict placements of objects according to coarse specifications in their positional relations.

\section{Input Scene Descriptions, Semantic Shopping Lists \& Scene Renderings}

Below, we display the semantic shopping lists and scene renderings for each of the 20 scenes mentioned in the main paper. 11 were selected randomly for the user study, and have baseline visualizations used for our human evaluation. For each scene, its input scene description and the corresponding scene reference (used in tables in the Experiments section of the main paper) is indicated in the section title. A star ($\star$) indicates that the scene was selected for the user study.

\begin{figure*}\centering\includegraphics[width=0.9 \textwidth]{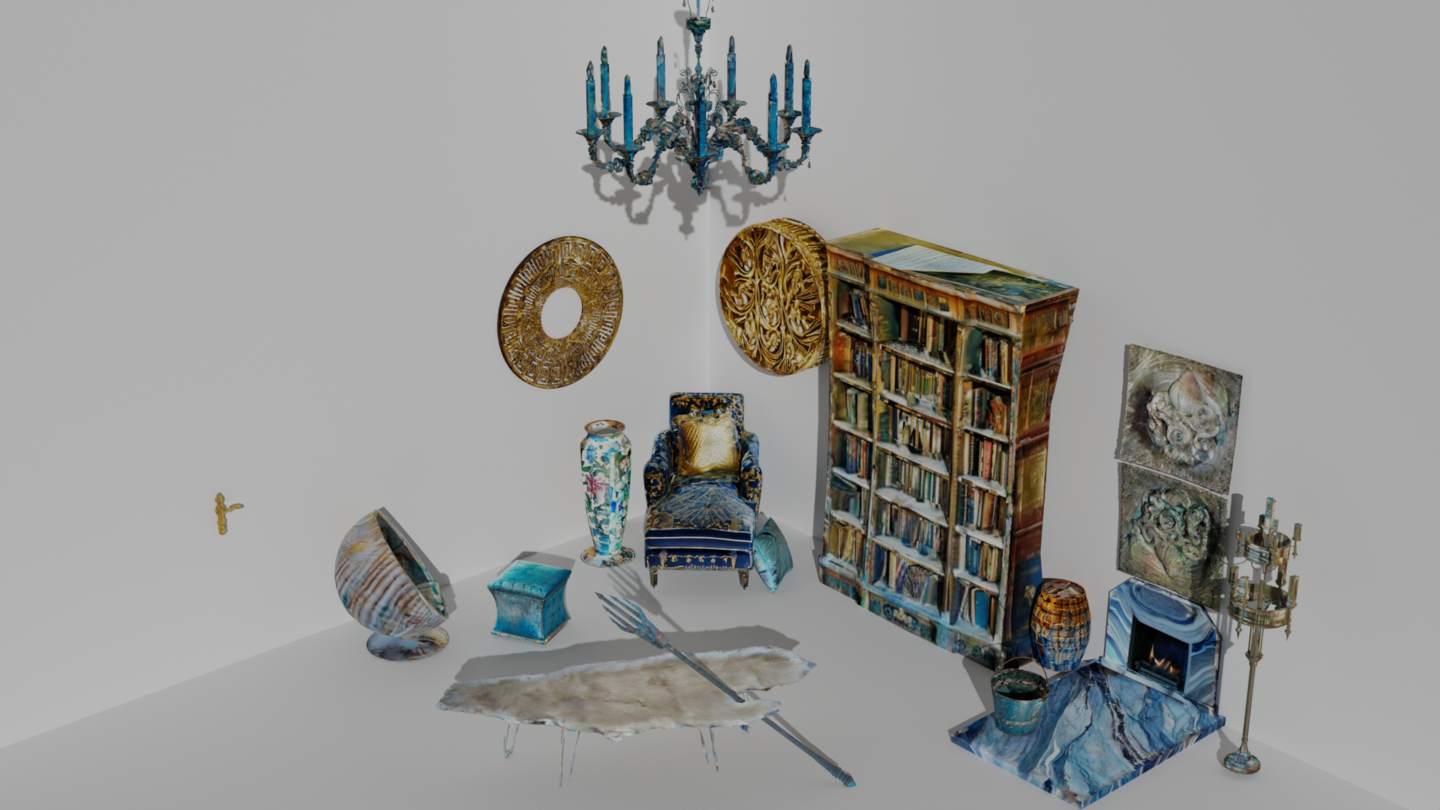}\caption{Our output for ``Poseidon's living room"}\label{fig:poseidon_living_room_ours}\end{figure*}

\begin{figure*}\centering\includegraphics[width=0.9 \textwidth]{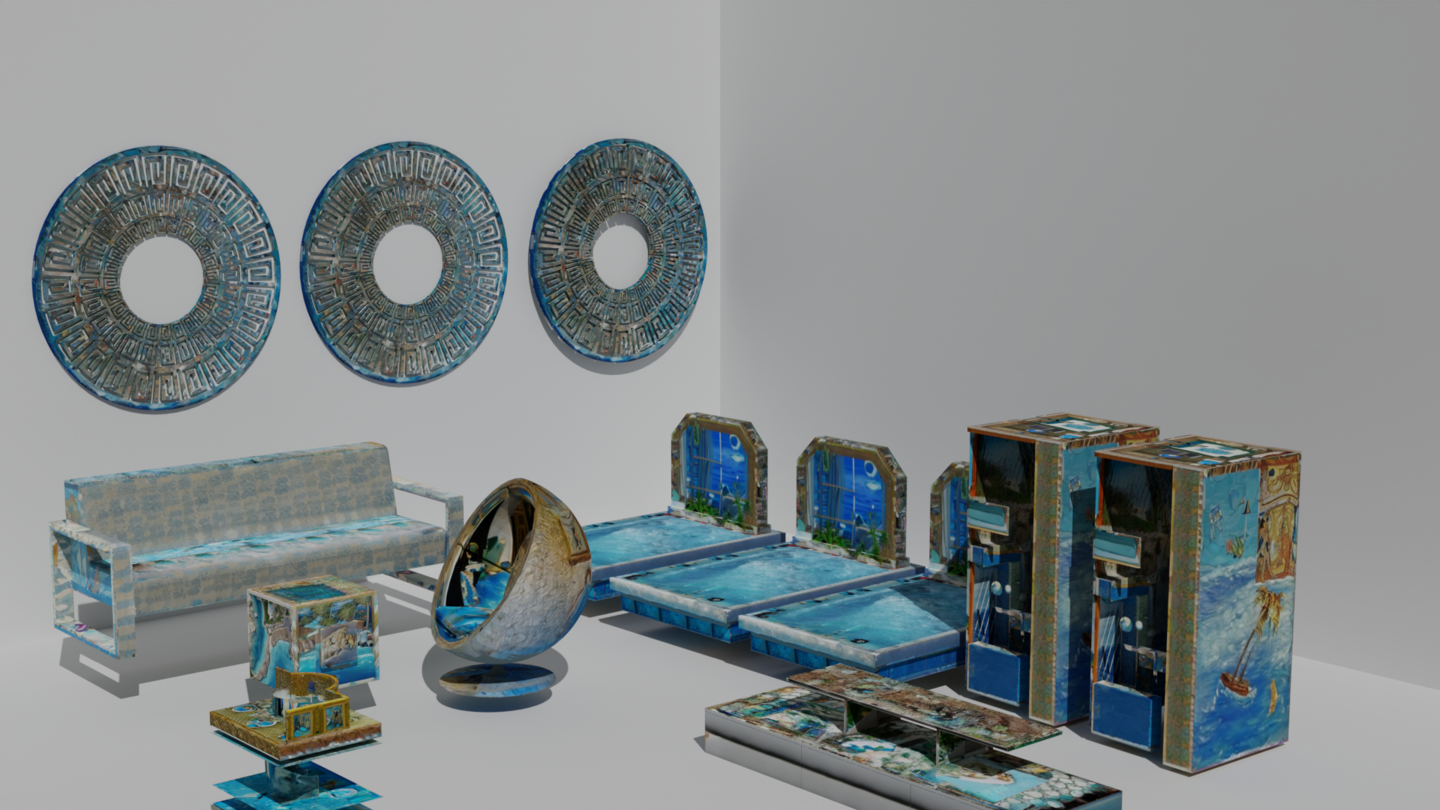}\caption{Baseline output for ``Poseidon's living room"}\label{fig:poseidon_living_room_baseline}\end{figure*}

\subsection{($\star$) Poseidon's living room (poseidon living room)-- Figures \ref{fig:poseidon_living_room_ours} and \ref{fig:poseidon_living_room_baseline}}
\noindent\begin{mdframed}
Input Scene Description : \textbf { Poseidon's living room } 
Semantic shopping list
\begin{enumerate}
\item throne: with intricate carvings of ocean life.glossy, polished finish.
\item fur rug: ivory and white with a hint of blue.soft and fluffy, with subtle wave patterns.
\item pillows: navy blue with gold accents.plump and luxurious.
\item candelabras: gold-plated with intricate designs.shiny and lustrous.
\item shield: gold-plated with intricate designs.gleaming and regal.
\item trident: gold-plated with intricate designs.sturdy and majestic.
\item fireplace: with blue marble accents and a marble mantel.smooth, polished stone.
\item firewood logs: split and ready to burn.clean and dry, ready to burn.
\item coal bucket: clean and unscratched.
\item fire poker: free of rust and in good condition.
\item fireplace screen: clean and undamaged.
\item candelabra: shiny with no visible signs of wear.
\item couches: upholstered in a deep navy blue.plush and soft, with no signs of wear.
\item throw pillows: soft and fluffy to the touch.
\item blankets: crisp and plush.
\item candlesticks: shiny and well-polished.
\item books: new and pristine condition.
\item ottoman: with gold accents and an ocean-blue tufted upholstery.no signs of wear, pristine condition.
\item pillows: plump and pristine, with no signs of wear.
\item carpet: plush and vibrant, with no signs of wear.
\item vase with flowers: vibrant and fresh, with no signs of wear.
\item bowl with fruits: vibrant and colorful, with no signs of wear.
\item books: sturdy and well-preserved, with no signs of wear.
\item mermaid statues: standing atop two large seashells.smooth marble, with intricate details in the carvings.
\item seashells: white and glossy.smooth, glossy finish.
\item fish statues: intricately detailed.no signs of wear or discoloration.
\item coral: intricate patterns and realistic texture.vibrant colors, no signs of fading or discoloration.
\end{enumerate}
\end{mdframed}


\begin{figure*}\centering\includegraphics[width=0.9 \textwidth]{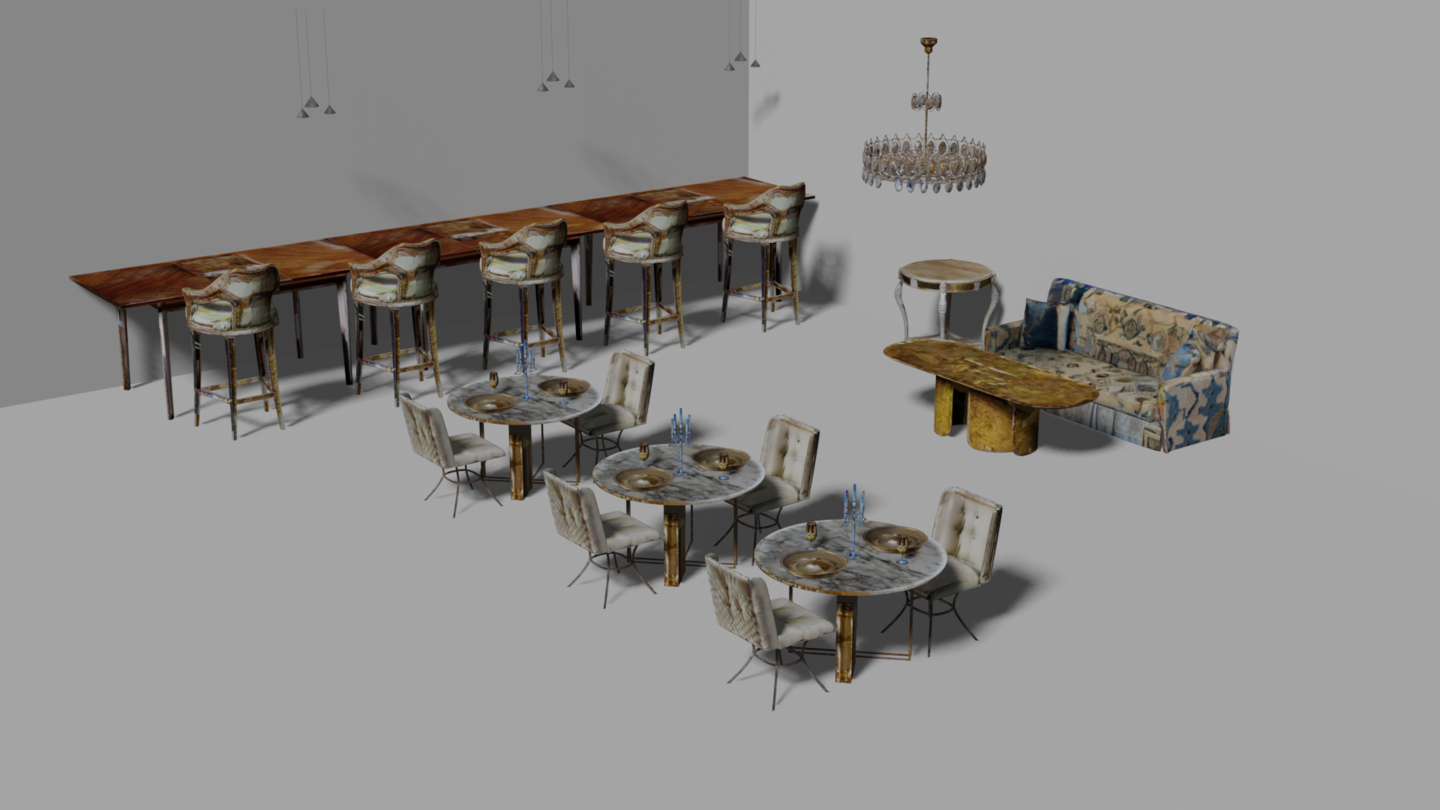}\caption{Our output for ``a romantic french restaurant"}\label{fig:romantic_french_restaurant_ours}\end{figure*}

\begin{figure*}\centering\includegraphics[width=0.9 \textwidth]{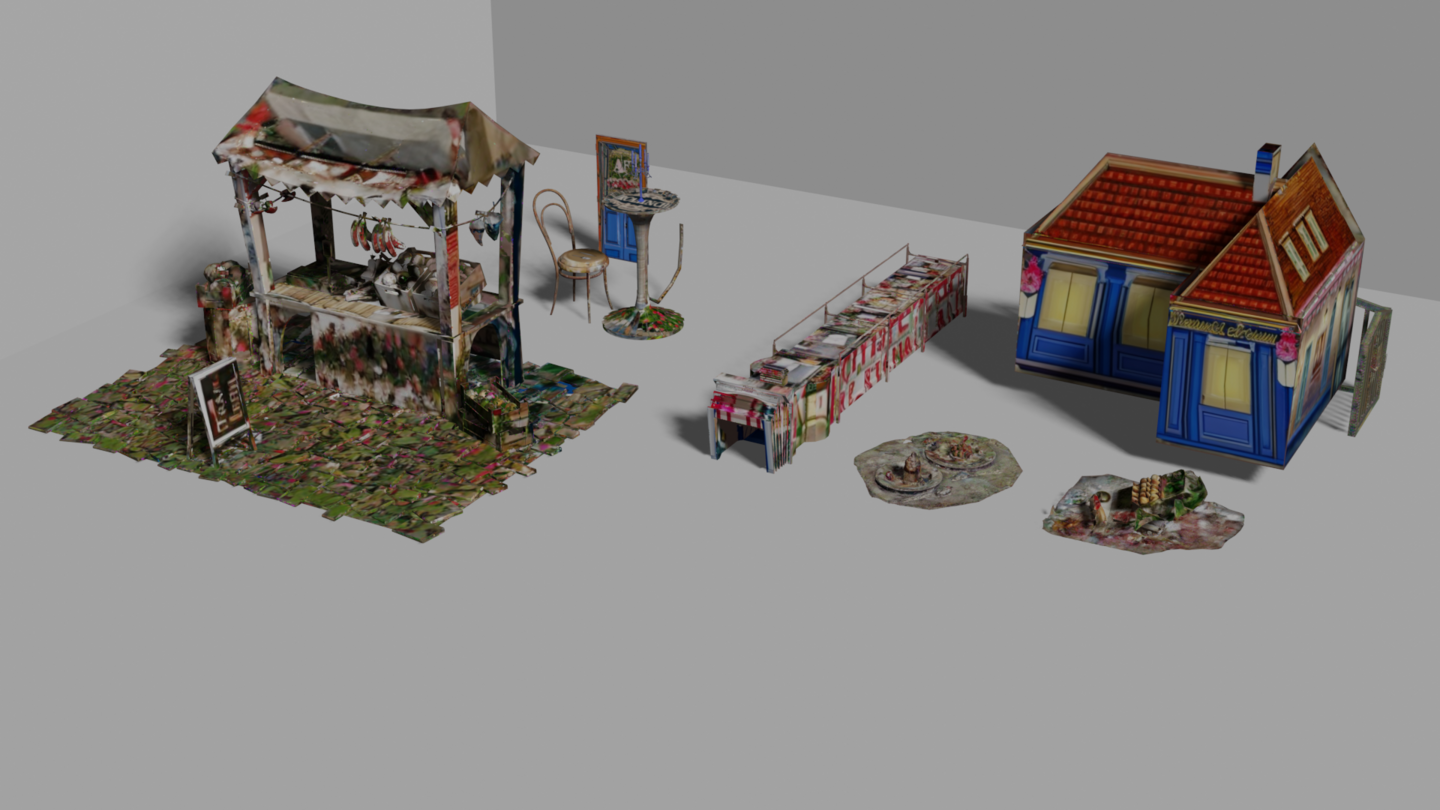}\caption{Baseline output for ``a romantic french restaurant"}\label{fig:romantic_french_restaurant_baseline}\end{figure*}

\subsection{($\star$) A romantic french restaurant (romantic restaurant)-- Figures \ref{fig:romantic_french_restaurant_ours} and \ref{fig:romantic_french_restaurant_baseline}} 
\noindent\begin{mdframed}
Input Scene Description : \textbf { a romantic french restaurant } 
Semantic shopping list
\begin{enumerate}
\item tables: two-tone oakwood finish.smooth, polished finish.
\item tablecloths: crisp and clean.
\item plates: no chips or cracks.
\item silverware: polished and gleaming.
\item wine glasses: no scratches or smudges.
\item candles: unburned and fragrant.
\item flowers: bright and freshly blooming.
\item chairs: upholstered in ivory velvet.slight signs of wear on the sides.
\item chair cushions: ivory color.soft and fluffy.
\item table linen: clean and wrinkle free.
\item candles: scented with lavender.in perfect condition, no dripping wax.
\item flower vase: polished and sparkling.
\item centerpiece: intricately detailed and in perfect condition.
\item bar: white marble, gold accents on the corners.slight signs of wear.
\item bar stools: upholstered in an off-white velvet fabric, with gold accents on the corners.slight signs of wear on the fabric.
\item wine glasses: with golden rims.shiny, with no signs of wear.
\item cocktail shakers: sleek and modern design.clean, no signs of wear.
\item coasters: with a distressed finish and a gold laurel leaf in the center.slight scratches due to regular use.
\item bottle openers: with a sleek and modern design.shiny and unscratched.
\item bar stools: upholstered in ivory velvet, golden accents on corners.slight wear on the rattan seats.
\item tablecloth: white and ivory.crisp and clean, no signs of wear.
\item coasters: white marble coasters with gold accents.smooth, polished finish.
\item ashtrays: black marble ashtrays with gold accents.slight signs of wear.
\item bar napkins: white and ivory.crisp and clean, no signs of wear.
\item cocktail shaker: smooth, polished finish.
\item chandelier: with crystal pendants and antique gold finish.ornate carvings and sparkling crystal pendants.
\item candles: ivory wax.melting wax and spreading soft light.
\item beaded curtains: ivory color.shimmering in the light, hanging gracefully.
\item wall sconces: gold metal with intricate designs.slightly aged, but still gleaming with their intricate designs.
\item lamps: ivory porcelain base with gold accents.softly illuminating the room in a warm and inviting light.
\item sofa: upholstered in ivory velvet, with golden accents on the legs.slight signs of wear on the velvet upholstery.
\item pillows: ivory and burgundy color.soft and fluffy.
\item blanket: ivory and burgundy color.smooth and plush.
\item end table: whitewashed oakwood, gold accents.smooth polished finish.
\item lamp: bronze base, ivory drum shade.no signs of wear.
\item candles: ivory and burgundy color.unscented.
\item fireplace: ornate carvings and polished gold accents.
\item firewood: cut into 16-inch lengths.freshly cut and dried.
\item fireplace tools: black finish.smooth and glossy finish.
\item candles: scented with lavender.unscented, clean surfaces.
\item throw pillows: in a pink and gold pattern.bright and vibrant colors.
\item rug: in ivory and blue.tightly woven, plush texture.
\item wall art: with gilded frame.crisp colors and gilded frame.
\item curtains: off-white color, with ruffles and lace.crisp and clean, with no signs of fading or wear.
\item vase: with intricate designs.shiny and sparkly, with no nicks or scratches.
\item flowers: in pink and ivory shades.soft and fragrant, looking freshly cut.
\item candle holder: gold plated.smooth and polished finish with no signs of rust.
\item candles: beeswax, scented with lavender.no signs of melting, wax evenly distributed.
\end{enumerate}
\end{mdframed}


\begin{figure*}\centering\includegraphics[width=0.9 \textwidth]{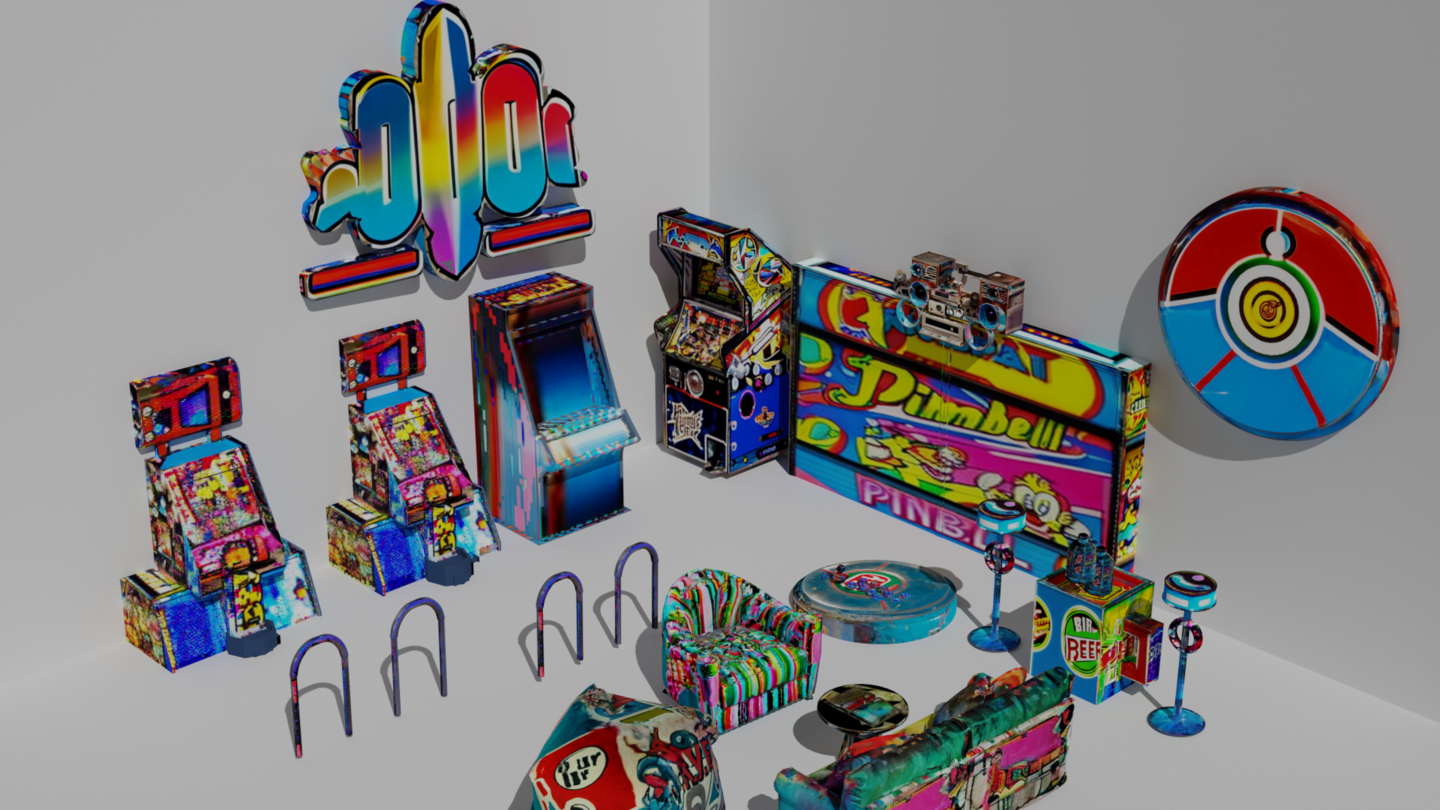}\caption{Our output for ``a retro arcade in the style of the 1980s"}\label{fig:retro_arcade_ours}\end{figure*}

\begin{figure*}\centering\includegraphics[width=0.9 \textwidth]{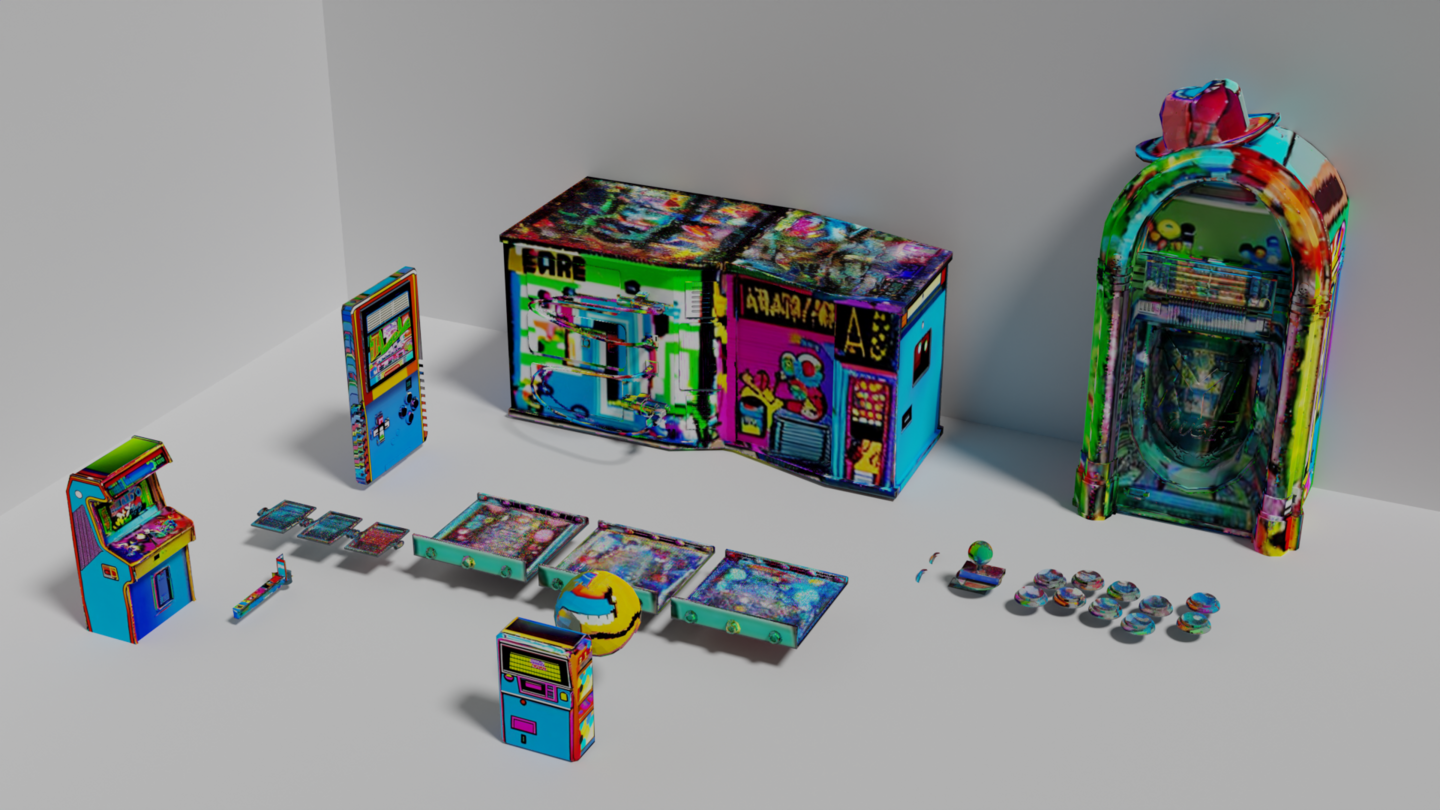}\caption{Baseline output for ``a retro arcade in the style of the 1980s"}\label{fig:retro_arcade_baseline}\end{figure*}

\subsection{($\star$) A retro arcade in the style of the 1980s (retro arcade) -- Figures \ref{fig:retro_arcade_ours} and \ref{fig:retro_arcade_baseline}}
\noindent\begin{mdframed}
Input Scene Description : \textbf { a retro arcade in the style of the 1980s } 
Semantic shopping list
\begin{enumerate}
\item pinball machine: featuring lights and sound effects from the 80s.brightly lit led lights, glossy finish and smooth movement.
\item flippers: black with chrome accents.minor scratches on the chrome accents.
\item pinball bumpers: bright and colorful with cartoon characters.vibrant colors and cartoon characters on the bumpers.
\item pinball slingshots: chrome with black accents.light rust on the chrome accents.
\item arcade machine: space invaders, etc.slight signs of wear on the buttons and joystick.
\item atari posters: bright colors and bold font.crisp colors, no rips or tears.
\item joysticks: bright colors with red and blue buttons.no signs of wear, bright colors.
\item retro speakers: chrome and with a retro design.no signs of wear, polished chrome finish.
\item coin dispenser: metal with a retro design.slight signs of wear, chrome finish still intact.
\item bar: preferably in a bright color.shiny finish with slight signs of wear.
\item neon-lights: brightly illuminated, no signs of wear.
\item drinks: full bottles with clean and crisp labels.
\item beer mugs: free of chips and cracks, vibrant colors.
\item ashtrays: no signs of wear, vivid colors.
\item bar stools: some rust on the metal frames, but vinyl seats still vibrant in color.
\item neon signs: operated with a remote.brightly lit and colorful.
\item ashtrays: slightly faded with signs of wear.
\item retro music system: cassette player, and cd player.dusty and slightly weathered.
\item retro decorations: highly detailed and brightly-colored.
\item retro seating: preferably in bright colors.slight signs of wear on the fabric, but still vibrant in color.
\item throw pillows: slightly faded, with some minor signs of wear.
\item drink holders: slightly tarnished, but still in good condition.
\item cushions: slightly faded and worn, but otherwise in good condition.
\item vintage poster: slightly faded, but still vibrant colours.
\end{enumerate}
\end{mdframed}


\begin{figure*}\centering\includegraphics[width=0.9 \textwidth]{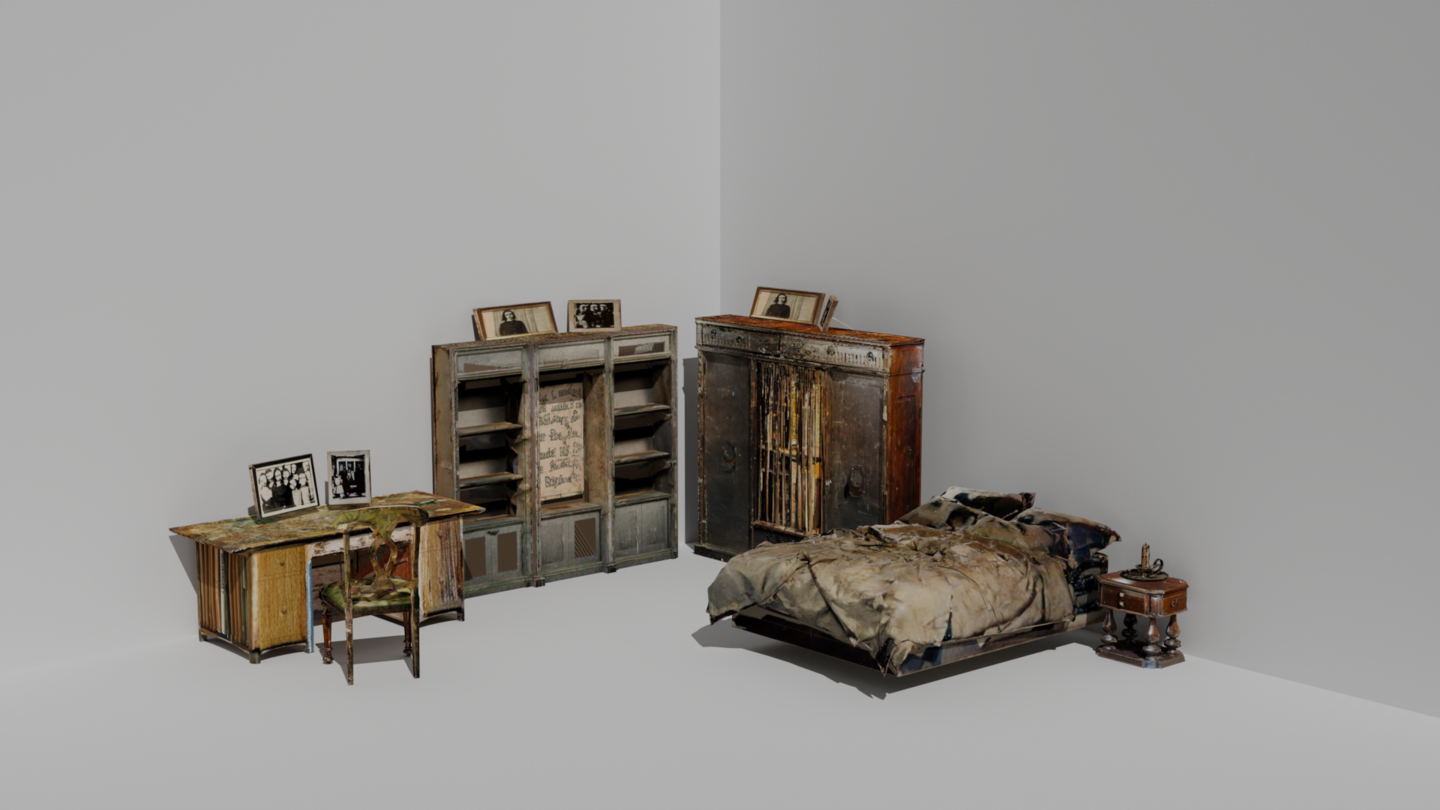}\caption{Our output for ``anne frank's room during world war II"}\label{fig:anne_frank_room_ours}\end{figure*}

\begin{figure*}\centering\includegraphics[width=0.9 \textwidth]{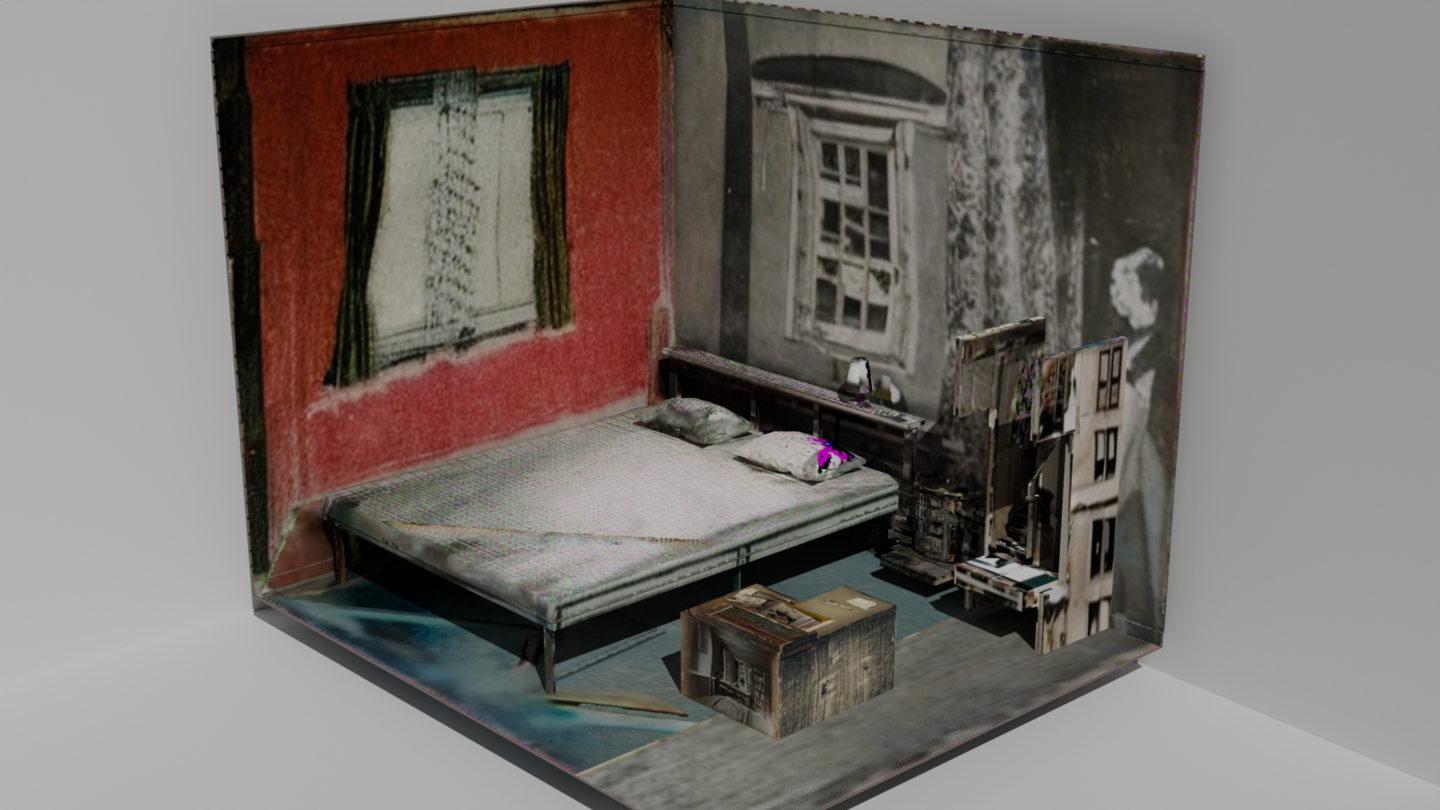}\caption{Baseline output for ``anne frank's room during world war II"}\label{fig:anne_frank_room_baseline}\end{figure*}

\subsection{($\star$) Anne frank's room during world war II (anne frank room)-- Figures \ref{fig:anne_frank_room_ours} and \ref{fig:anne_frank_room_baseline}}
\noindent\begin{mdframed}
Input Scene Description : \textbf { anne frank's room during world war II } 
Semantic shopping list
\begin{enumerate}
\item bed: made of solid wood.slightly weathered and with a few scuffs.
\item blanket: beige color.discolored with age, but still intact.
\item pillow: white cover.slightly lumpy from years of use.
\item bedside table: dark brown stain.well-worn with scratches and chips.
\item lamp: yellow glass shade.tarnished, but still functional.
\item book: embossed cover.faded around the edges, with notes written inside.
\item clothes: blue and white checkered patterns.slightly worn, but still in good condition.
\item desk: wooden construction and a dark wood finish.slightly worn on the edges.
\item pencils: natural wood color.slightly worn and dull, with a few scratches.
\item books: with antique paper pages.faded covers, pages slightly yellowed from aging.
\item paper: aged look.delicate and fragile, with visible signs of wear.
\item postcards: slightly faded, with minor creases and wrinkles.
\item bookshelf: slightly worn on the edges.
\item books: with the original cover design.slightly worn cover with some fading.
\item candle: with a simple black holder.wax melted down to the base.
\item picture frame: with a white matte finish.slight scratches on the corner of the frame.
\item doll: wearing a white dress with a floral pattern.age-related discoloration, with some loose threads.
\item chair: slightly worn on the fabric.
\item cushion: patchwork design with embroidery.slightly worn and faded from time.
\item blankets: one with patchwork, the other plain.slight fraying and discoloration from wear.
\item books: such as anne of green gables or the adventures of tom sawyer.faded pages, with some discoloration due to age.
\item pen: with an ornate design.slight wear from age, but still usable.
\item notebook: with embossed floral design.vintage leather, but still in good condition.
\item clothes: gray or navy with white buttons.slightly faded, but still in wearable condition.
\item cabinet: some scratches, but overall in relatively good condition.
\item books: slightly faded and tattered covers.
\item clothes: simple and worn.
\item photos: slightly faded and yellowed.
\item letters: wrinkled and slightly faded.
\item curtains: slightly faded and with some small tears.
\item window: with a thin glass pane and black iron hingesslightly worn with scratches from use, but still in good condition.
\item photos: crisp black and white photos, without any discoloration.
\item books: in a variety of colors such as black, brown, and red.some signs of wear on the spines and edges, but overall in good condition.
\item bedding: slightly worn from years of use, but still in good condition.
\item clothes: and warm sweaters in muted colors.slightly worn, but still in good condition.
\end{enumerate}
\end{mdframed}


\begin{figure*}\centering\includegraphics[width=0.9 \textwidth]{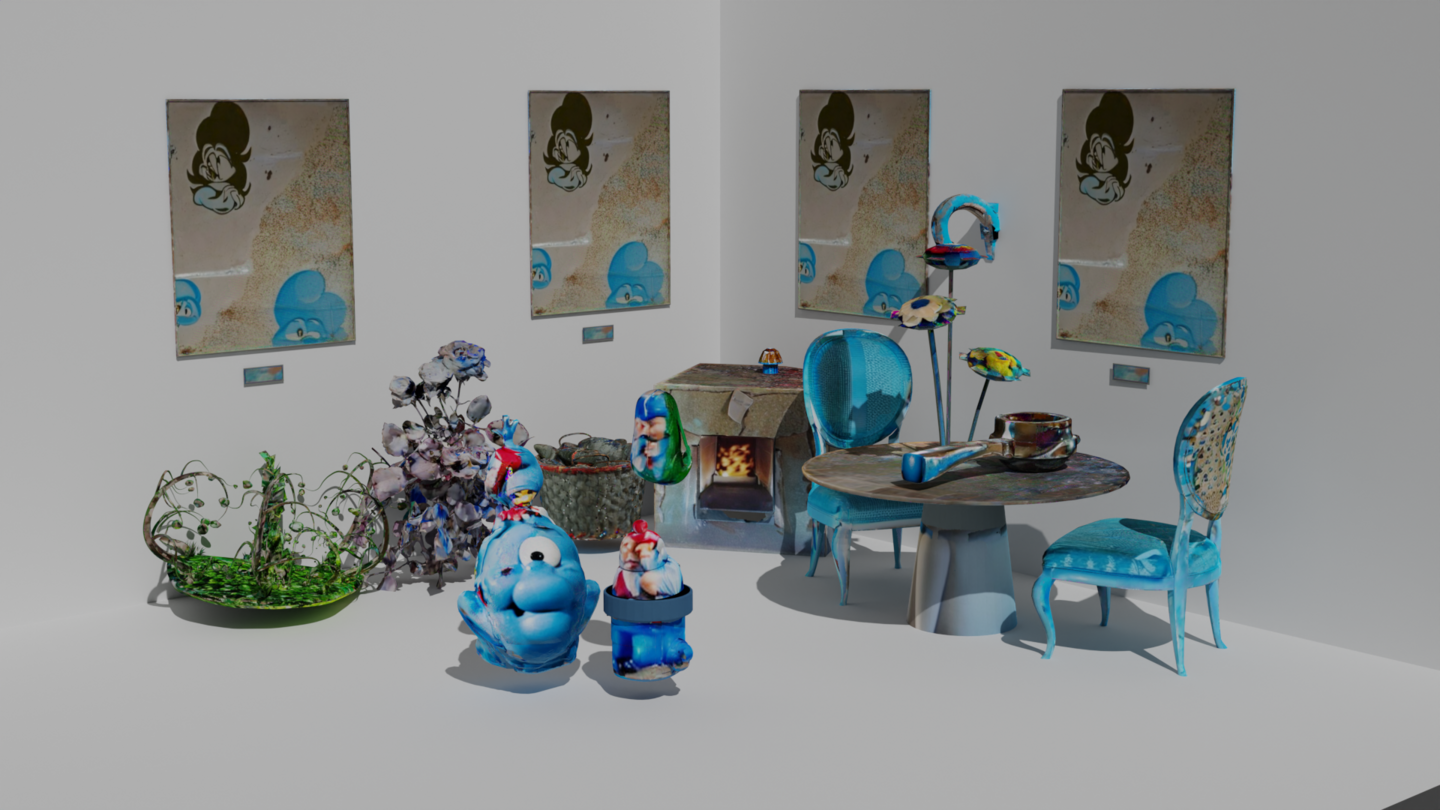}\caption{Our output for ``the interior of the smurf house" "}\label{fig:smurf_house_ours}\end{figure*}

\begin{figure*}\centering\includegraphics[width=0.9 \textwidth]{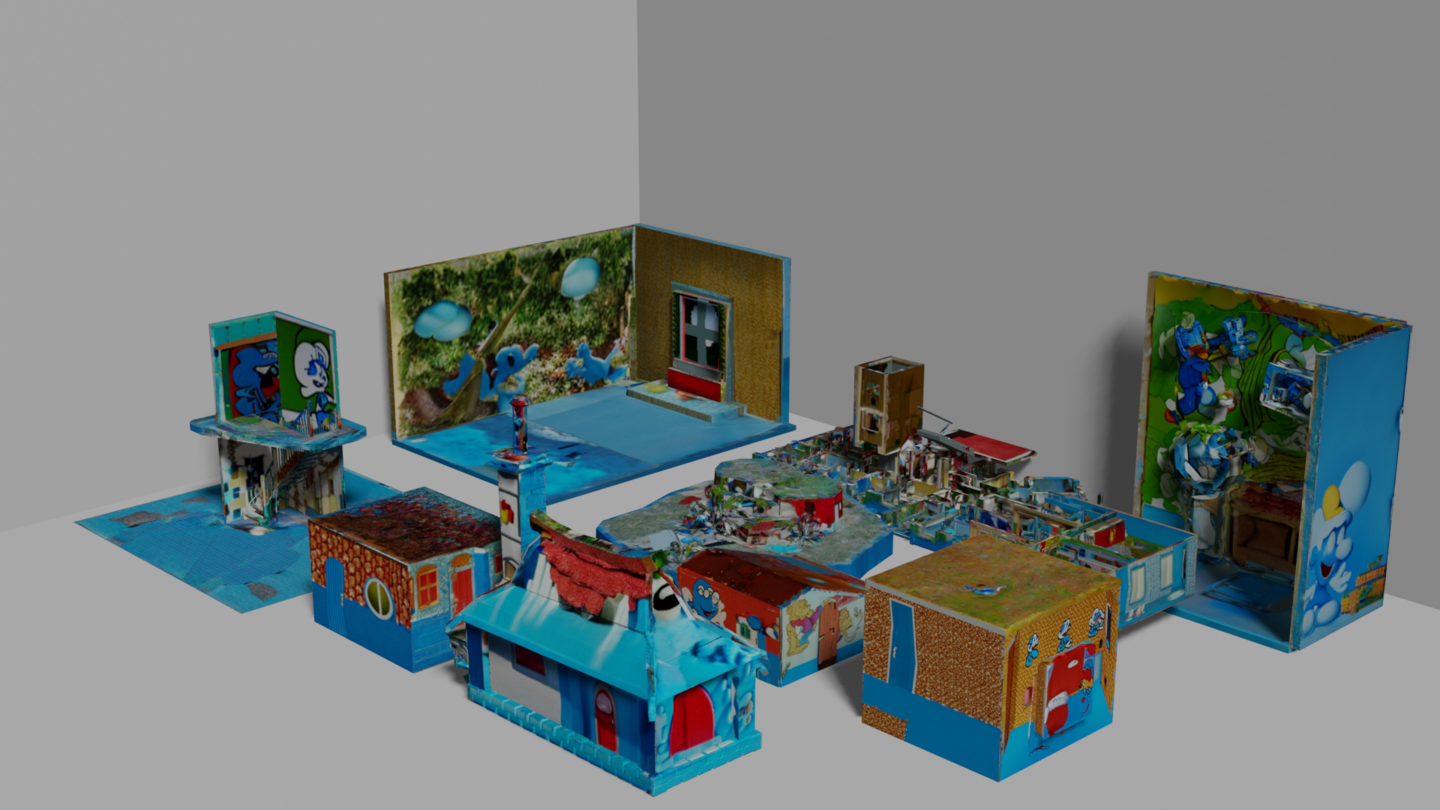}\caption{Baseline output for ``the interior of the smurf house" "}\label{fig:smurf_house_baseline}\end{figure*}

\subsection{($\star$) The interior of the smurf house (smurf house)-- Figures \ref{fig:smurf_house_ours} and \ref{fig:smurf_house_baseline}}
\noindent\begin{mdframed}
Input Scene Description : \textbf { the interior of the smurf house } 
Semantic shopping list
\begin{enumerate}
\item mushroom house: blue and white in color.new and pristine
\item mushrooms: with blue-green dots.vibrant colors, no chips or cracks.
\item flowers: petals slightly wilted, with no fading of colors.
\item window frames: smooth surface, no signs of wear and tear.
\item fireplace: no signs of wear or discoloration.
\item trees: green leaves.vibrant with fresh leaves
\item wildflowers: vibrant and lifelike, with no tears or signs of wear.
\item leaves: soft to the touch, with no rips or tears.
\item acorns: glossy and realistic, with no cracks or chips.
\item mushrooms: white caps, and red spots.smooth and glossy, with no tears or creases.
\item table: clean and smooth
\item tablecloth: crisp and clean.
\item plates: no chips or cracks.
\item silverware: no tarnishing.
\item smurf-shaped cupcakes: with edible smurf decorations.freshly baked and frosted.
\item chairs: sturdy and clean
\item flowers: in various shapes.pristine condition, no signs of wear or tear.
\item birdhouse: with a pointed roof.freshly painted and in good condition.
\item gnome figurines: dressed in traditional smurf attire.vibrant colors, no chips or cracks.
\item paintings: crisp edges, no fading of colors.
\item smurf figures: new and vivid in color.
\item mushrooms: bright and glossy, with no signs of wear.
\item flowers: bright and vibrant colors, no signs of fading.
\item fishing rod: smooth and painted with bright colors, no signs of wear.
\item basket: tightly woven and sturdy, no signs of fraying.
\end{enumerate}
\end{mdframed}


\begin{figure*}\centering\includegraphics[width=0.9 \textwidth]{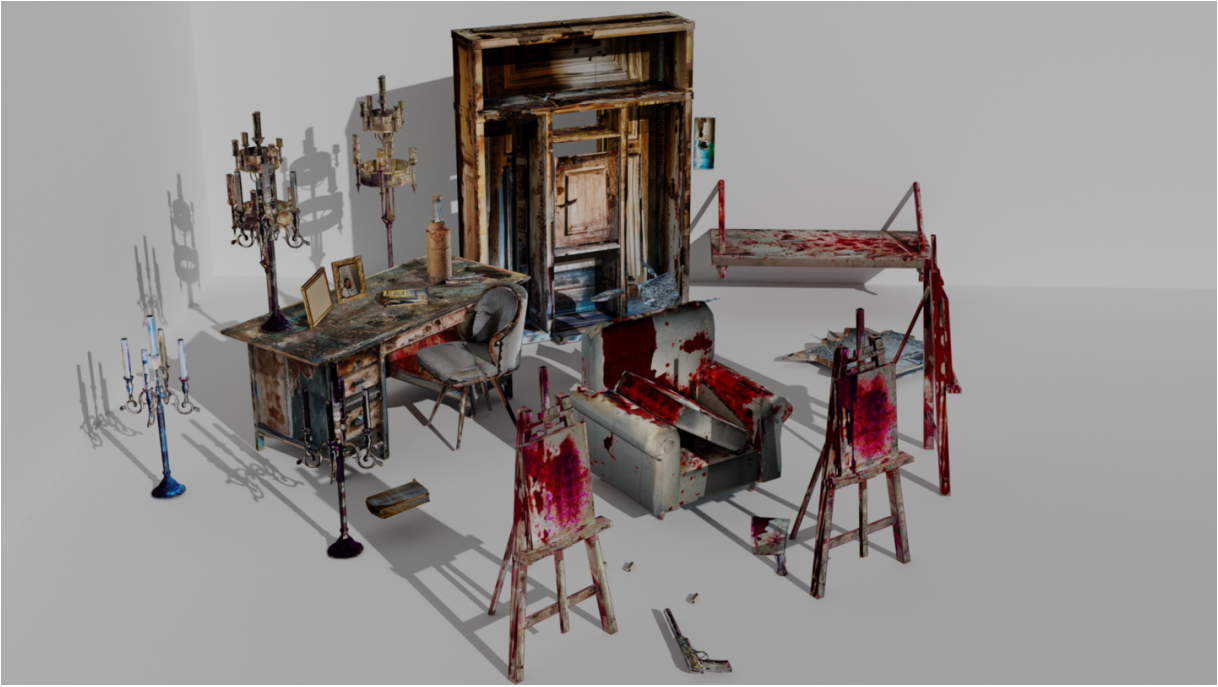}\caption{Our output for ``a murder in an abandoned living room"}\label{fig:murder_abandoned_livingroom_ours}\end{figure*}

\begin{figure*}\centering\includegraphics[width=0.9 \textwidth]{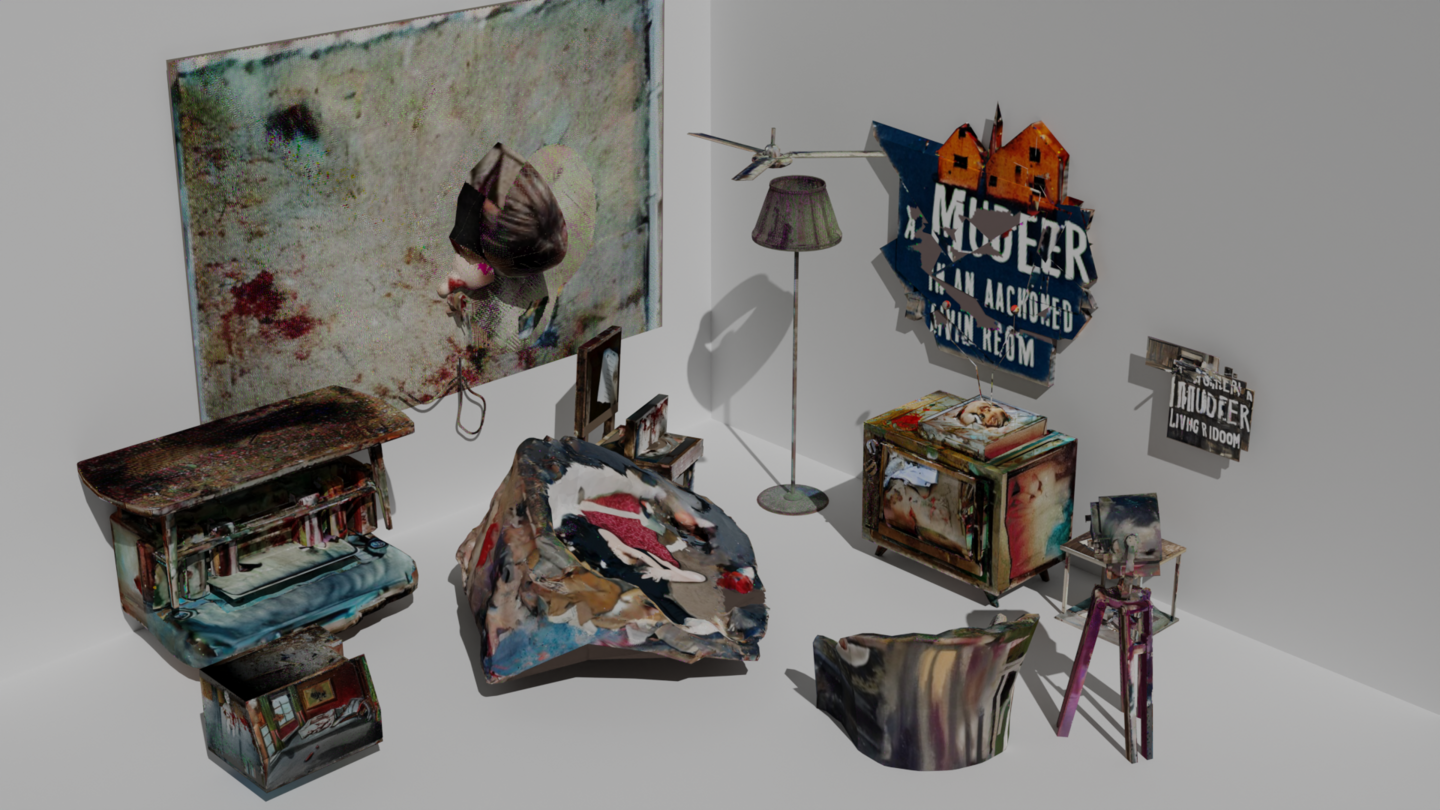}\caption{Baseline output for ``a murder in an abandoned living room"}\label{fig:murder_abandoned_livingroom_baseline}\end{figure*}

\subsection{($\star$) A murder in an abandoned living room (murder room)-- Figures \ref{fig:murder_abandoned_livingroom_ours} and \ref{fig:murder_abandoned_livingroom_baseline}}
\noindent\begin{mdframed}
Input Scene Description : \textbf { a murder in an abandoned living room } 
Semantic shopping list
\begin{enumerate}
\item couch: stained and torn fabric, fraying at the edges.
\item pillows: slightly worn and stained.
\item blankets: light greytorn and stained.
\item blood stains: dried and smeared across surfaces.
\item magazine: preferably one from the 1960saged and yellowed.
\item candle: melted and partially burned.
\item lamp: dusty and cobwebbed.
\item table: made from teak wood.scuffed and scratched surfaces.
\item lamp: dusty and worn.
\item papers: scattered on the floor.
\item pen: traces of dried ink on the nib.
\item gun: scratched and worn.
\item blood spatters: with a realistic texture.slightly damp in places.
\item desk: made of reclaimed wood.dust and dirt accumulated in the corners.
\item lamp: bronze and cream.tarnished, dented, and dusty.
\item pen holder: scuffed and scratched.
\item pens/pencils: dull, worn, and faded.
\item stack of paper: aged parchment paper.yellowed and aged.
\item mug: with a faded sketch of a crow.chipped and cracked.
\item chair: upholstered in a grey linen fabric.signs of wear in the upholstery.
\item pillow: slightly stained and worn out.
\item blanket: faded grey.discolored and frayed.
\item candle: waxen and melted.
\item book: discolored pages with visible creases and wrinkles.
\item newspaper: torn and crumpled.
\item coffee cup: cracked handle.broken handle and stained.
\item bottle: empty, with a cork stopper.empty, with a cork stopper.
\item window: cracked and broken glass.
\item shattered glass: silver and black.shards of broken glass scattered on the floor.
\item curtains: black and grey.torn and tattered.
\item window blinds: black and grey.torn and damaged, some of the slats are missing.
\item bullet hole: black.a large circular hole in the wall with frayed edges.
\item blood stains: red and black.dark red splatters on walls and furniture.
\item door: made of reclaimed wood.large dents and scratches.
\item key: rusty and tarnished.
\item knob: rusty and tarnished.
\item light switch: flickering and weak.
\item window curtains: black velvet curtains.ripped and tattered.
\item bloodstain: red, water-based paint.fresh and bright.
\item lamp: bronze with a white shade.slightly dusty.
\item photo frame: antique gold frame for a 4x6 photo.slightly dusty and cracked.
\item rug: faded colors and worn out threads.
\item blood stains: non-toxic and non-hazardous.smeared and splattered on the walls and floors.
\item broken glass: realistic-looking, made of plastic.scattered across the room in pieces.
\item empty beer bottle: made of plastic.lying upside down on the floor.
\item bloodied knife: lying next to the empty beer bottle.
\item bullet casing: made of plastic.scattered across the floor.
\item bloodied rag: lying in a corner of the room.
\end{enumerate}
\end{mdframed}


\begin{figure*}\centering\includegraphics[width=0.9 \textwidth]{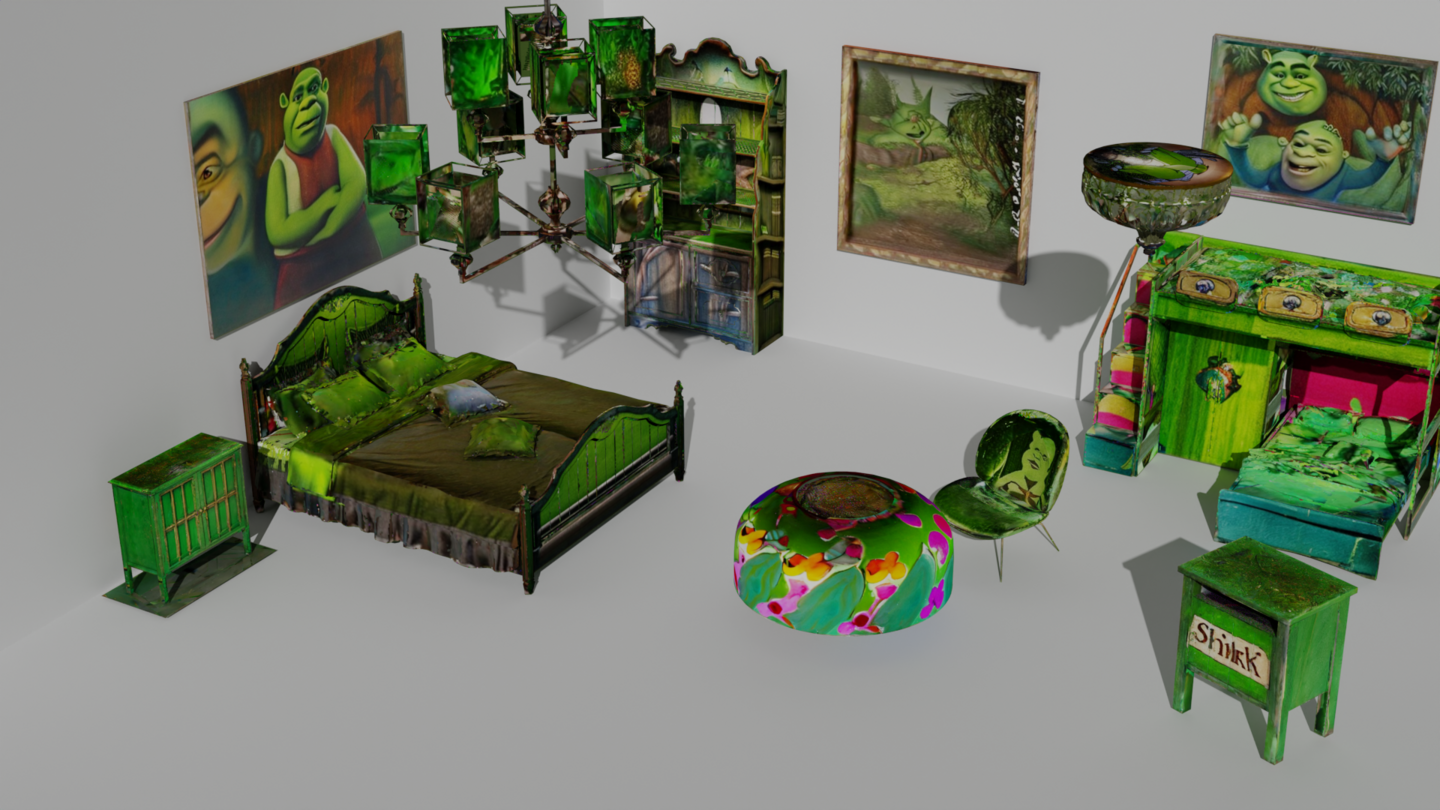}\caption{Our output for ``bedroom in shrek's home in the swamp"}\label{fig:shrek_home_ours}\end{figure*}

\begin{figure*}\centering\includegraphics[width=0.9 \textwidth]{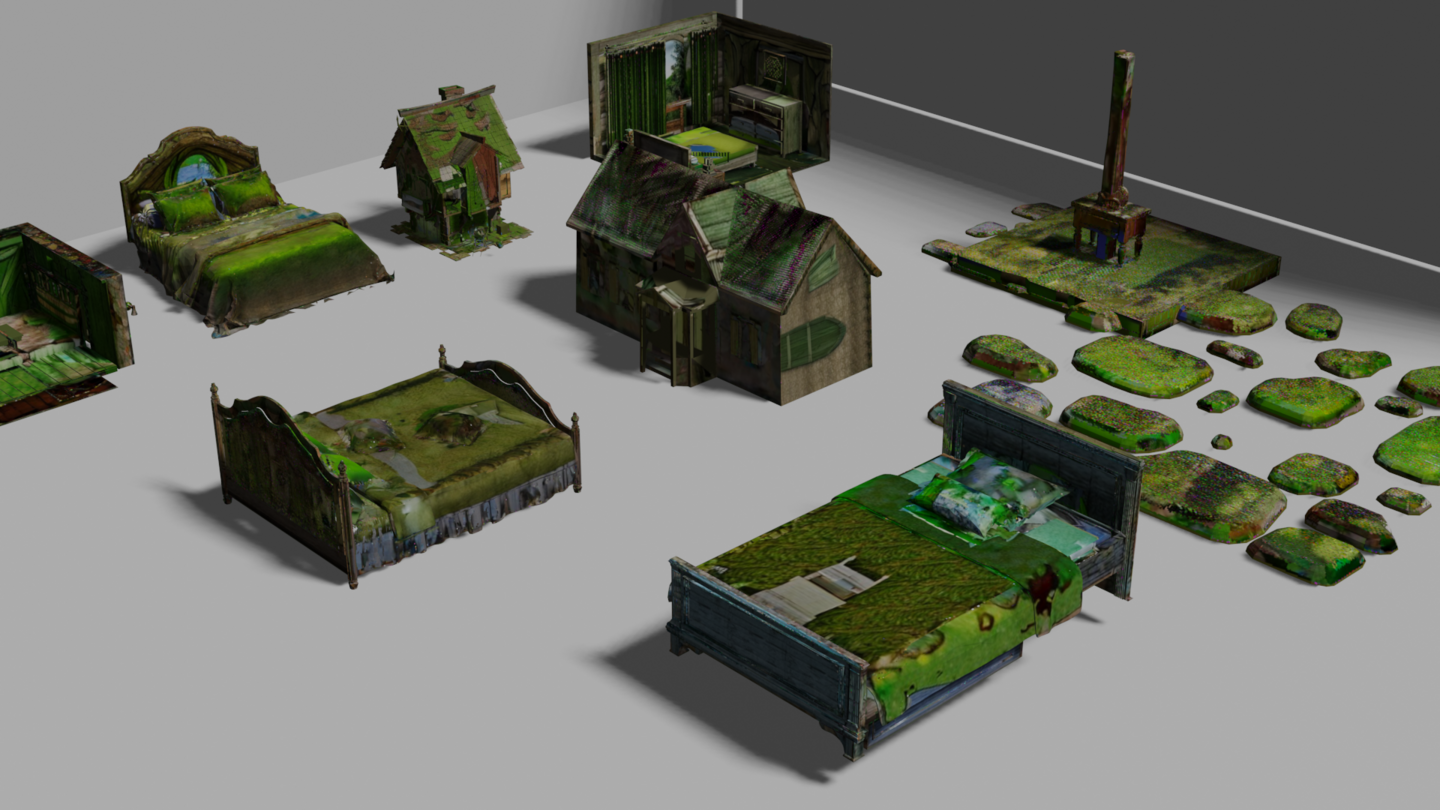}\caption{Baseline output for ``bedroom in shrek's home in the swamp"}\label{fig:shrek_home_baseline}\end{figure*}

\subsection{($\star$) Bedroom in Shrek's home in the swamp (shrek home)-- Figures \ref{fig:shrek_home_ours} and \ref{fig:shrek_home_baseline}}
\noindent\begin{mdframed}
Input Scene Description : \textbf { bedroom in shrek's home in the swamp } 
Semantic shopping list
\begin{enumerate}
\item bed: dark wood with green fabric accents.sturdy and in good condition.
\item pillows: fluffy, with no signs of wear.
\item blanket: slightly worn edges, with a soft texture.
\item stuffed animal: plush, cartoon-style ogre.soft, plush, with no signs of wear.
\item books: slightly tattered corners, with no major scuffs or scratches.
\item shoes: cartoon-style ogre shoes.new, with bright colors and sturdy soles.
\item cabinet: green paint with gold accents.minor signs of wear.
\item cabinet knob: cast iron with a green and gold finish.slightly worn, with a few scratches.
\item dishes: with a green and brown glaze.no chips or cracks, with a glossy finish.
\item bowls: with a green and brown glaze.no chips or cracks, with a glossy finish.
\item mugs: with a green and brown glaze.no chips or cracks, with a glossy finish.
\item desk: distressed wood with iron accents.minor signs of wear.
\item pencil holder: stained with a dark green finish.sturdy, with no visible scratches.
\item notepad: wooden notepad with a "shrek" motif carved on the cover.smooth, with no visible signs of wear and tear.
\item pens: with green and yellow feathers, stored in a velvet pouch.clean and sharp, with no visible stains.
\item books: bound in green leather and embossed with golden lettering.unmarked, with gold lettering intact and no visible wear and tear.
\item chair: green velvet upholstery and gold accents.minor signs of wear.
\item blanket: soft and warm.
\item bookshelf: well-crafted and sturdy.
\item books: featuring characters from the shrek movie.new and in pristine condition.
\item rug: green and gold motif.thick and plush.
\item throw pillows: squishy texture.fluffy with no visible signs of wear.
\item blanket: plaid design woolen blanket in brown and green tones.free from any snags or tears.
\item bookshelf: smooth, polished finish with minimal signs of wear.
\item books: perfectly bound and crisp pages with no signs of fading.
\item lamp: green glass with bronze accents.no chips or cracks.
\item lamp shade: smooth and polished finish with vibrant colors.
\item lightbulb: bright and evenly lit.
\item lampshade finial: smooth and well-polished.
\item lamp base: smoothly painted with vibrant colors.
\item lampshade fringe: soft and lightweight.
\item lampshade harp: shiny and polished with a patina finish.
\item lampshade diffuser: vibrant colors and free of wrinkles.
\item wall art: framed in gold.vibrant colors, free of any damage.
\item stuffed animals: one donkey and one puss in boots.soft and cuddly.
\item painting: depicting various characters from the movie.vivid and colorful.
\item picture frame: sturdy and unblemished.
\item bookshelf: with a whimsical design.polished and smooth-to-the-touch.
\item books: titles to include "shrek's adventures" and "fairy tales of the swamp".crisp pages, with vibrant illustrations.
\end{enumerate}
\end{mdframed}


\begin{figure*}\centering\includegraphics[width=0.9 \textwidth]{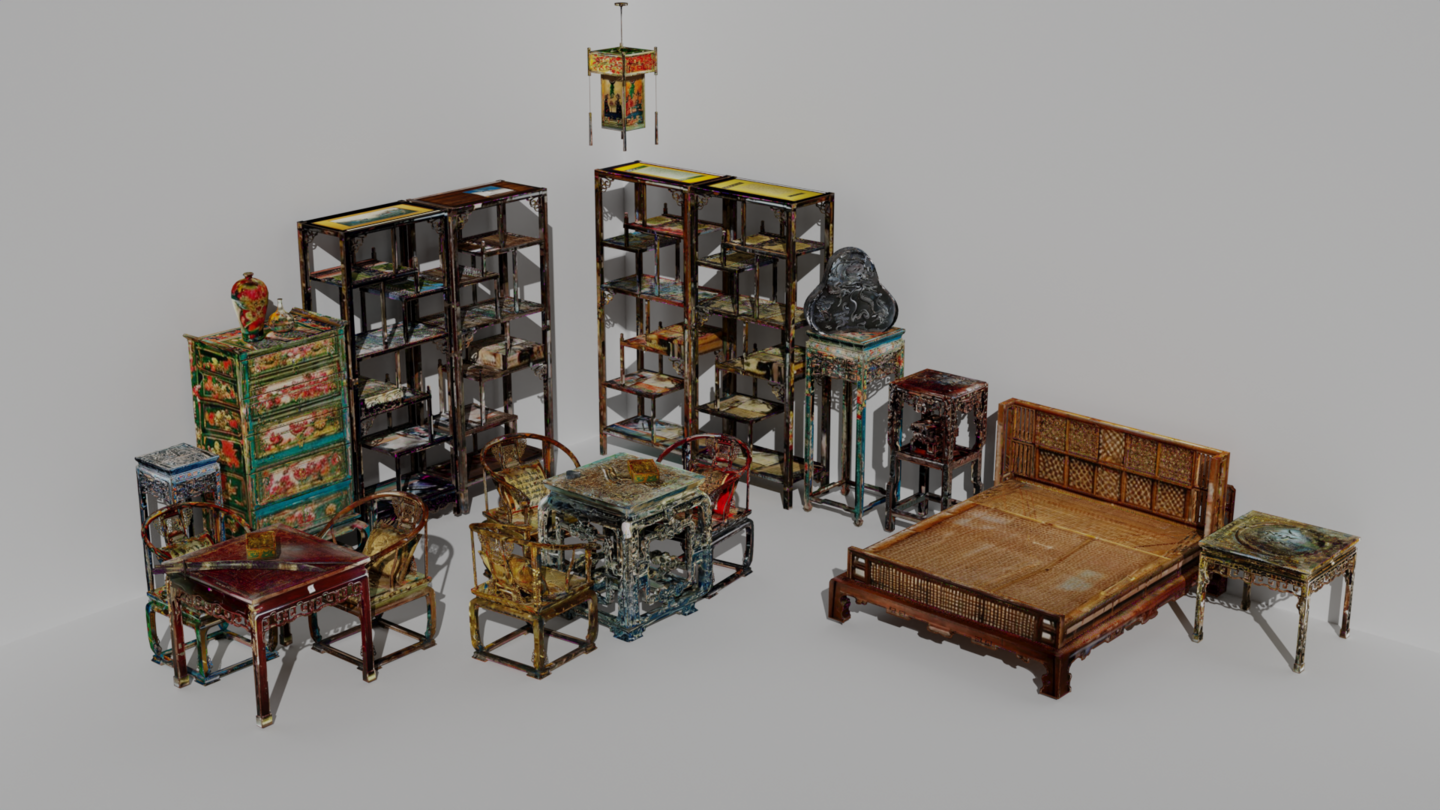}\caption{Our output for ``“confucius's bedroom”"}\label{fig:confucius_bedroom_ours}\end{figure*}

\subsection{($\star$) Confucius's bedroom (confucius bedroom)-- Figures \ref{fig:confucius_bedroom_ours} and \ref{fig:confucius_bedroom_ours}}
\noindent\begin{mdframed}
Input Scene Description : \textbf { confucius's bedroom } 
Semantic shopping list
\begin{enumerate}
\item bed: solid oak wood, dark brown finish.gently worn but still in good condition.
\item pillows: with intricate golden embroidery.soft and plush.
\item blankets: with tassels and embroidered motifs.soft and smooth to touch.
\item bedside cabinet: carved with dragons and made of dark wood.polished and gleaming.
\item scrolls: unrolled and in perfect condition.
\item incense burner: with dragon etchings.bright and shining.
\item dresser: painted with floral motifs, gold accents.some signs of wear but still colorful and vibrant.
\item boxes: lacquered in gold.lightly distressed with some wear around the edges.
\item scrolls: mounted and framed.bright and colorful, lightly aged.
\item books: hardcover in blue and green.slightly faded covers with faint traces of wear.
\item vases: painted with floral motifs.some signs of wear and cracking, but still in good condition.
\item desk: dark cherry wood, golden detailing.well-maintained with a slight patina.
\item pen holder: glossy, with intricate carvings on the edges.
\item ink stone: made of black slate.polished smooth surface.
\item ink brush: made of bamboo and horsehair bristles.bristles in perfect condition.
\item scroll: with calligraphy writing.no signs of wear or tear.
\item papers: made of hemp.crisp and clean.
\item chair: carved and upholstered in silk.well-maintained with minimal signs of wear.
\item blanket: plain blue or red with yellow embroiderysoft and well worn.
\item bookshelf: dark brown, metal accents on edgeswell-polished with metal accents gleaming.
\item books: aged and well-handled, with a few loose pages.
\item scroll: traditional red or black inkcrisp and vibrant, with the ink color vividly preserved.
\item inkstone: black stone with carved dragon designglossy and smooth, with the dragon design still clearly visible.
\item bamboo screen: natural brown.strong and sturdy, with a few scratches.
\item scrolls: with chinese characters.bright colors, free of tears and creases.
\item ink-stone pot: ceramic, with intricate patterns.smooth, polished finish, free of scratches and cracks.
\item brush and ink: bristles are intact, handle is strong and sturdy.
\item calligraphy scrolls: with chinese characters.neatly cut edges, free of tears and wrinkles.
\item paper fans: with painted chinese characters.no tears or creases, colors are still vibrant.
\item table lamps: golden finish, hand-painted shades.slightly weathered but still bright and colorful.
\item paintings: framed in black lacquer.smooth, glossy finish.
\item writing utensils: black lacquer handle.intact and in pristine condition.
\item scrolls: framed in black lacquer.slightly yellowed edges from age but overall in good condition.
\item candles: with intricate gold designs.intact with no chips or cracks.
\end{enumerate}
\end{mdframed}
\begin{figure*}\centering\includegraphics[width=0.9 \textwidth]{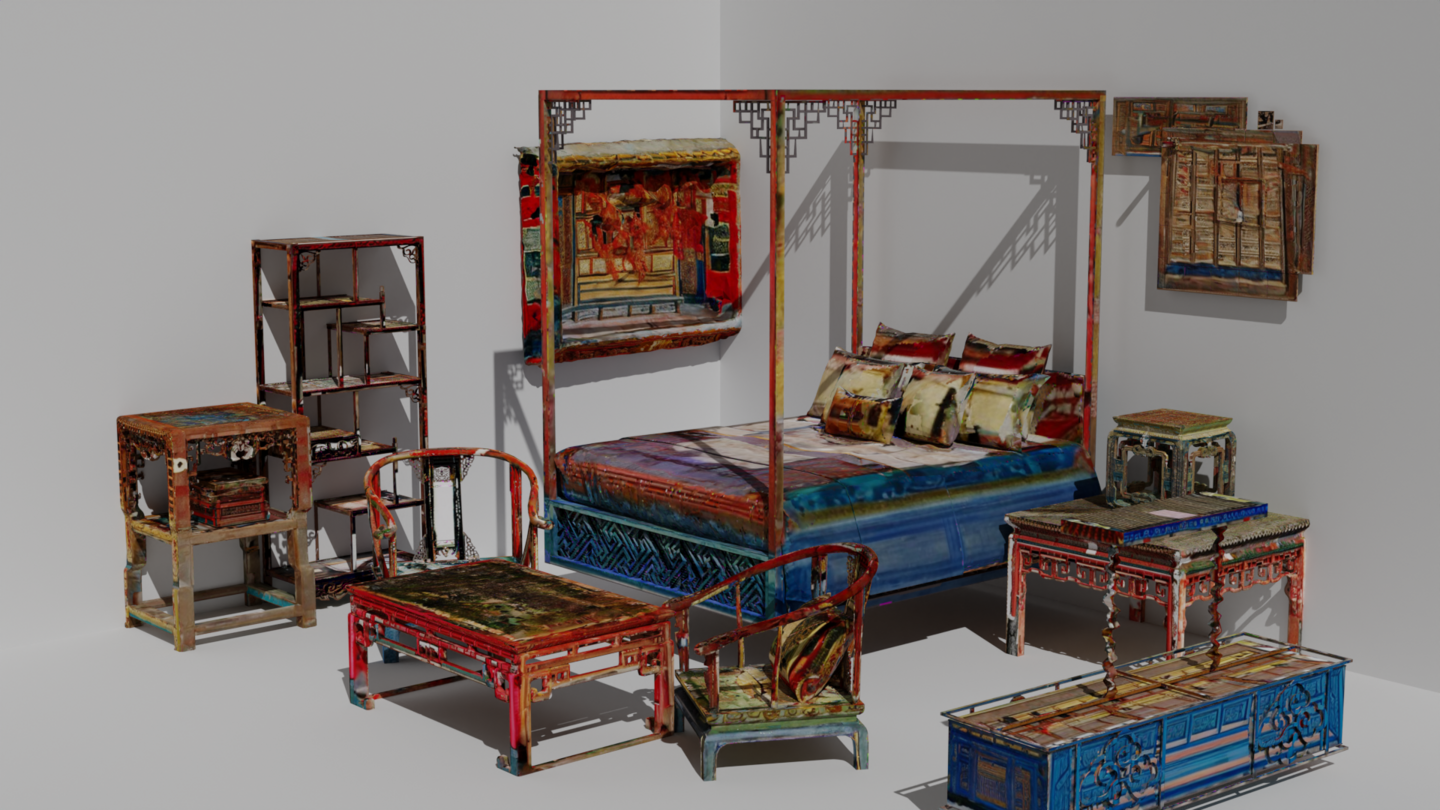}\caption{Baseline output for ``“confucius's bedroom”"}\label{fig:confucius_bedroom_baseline}\end{figure*}

\begin{figure*}\centering\includegraphics[width=0.9 \textwidth]{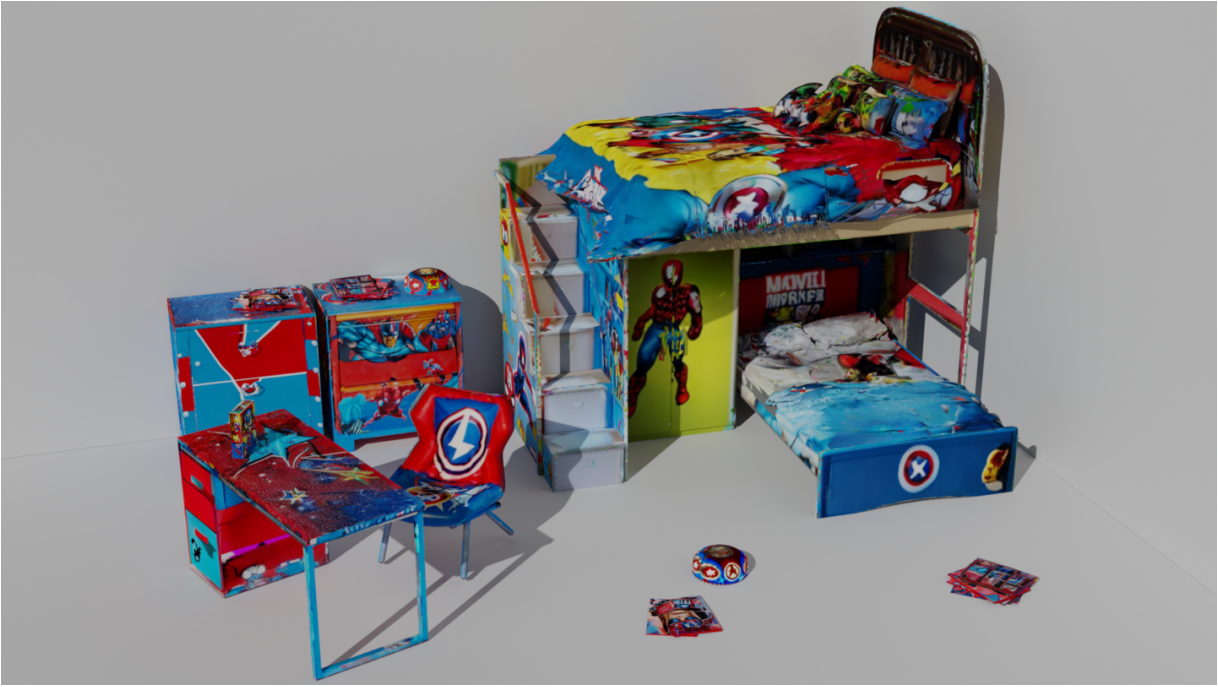}\caption{Our output for ``“a marvel-themed bedroom of a five-year old toddler"}\label{fig:marvel_bedroom_ours}\end{figure*}

\begin{figure*}\centering\includegraphics[width=0.9 \textwidth]{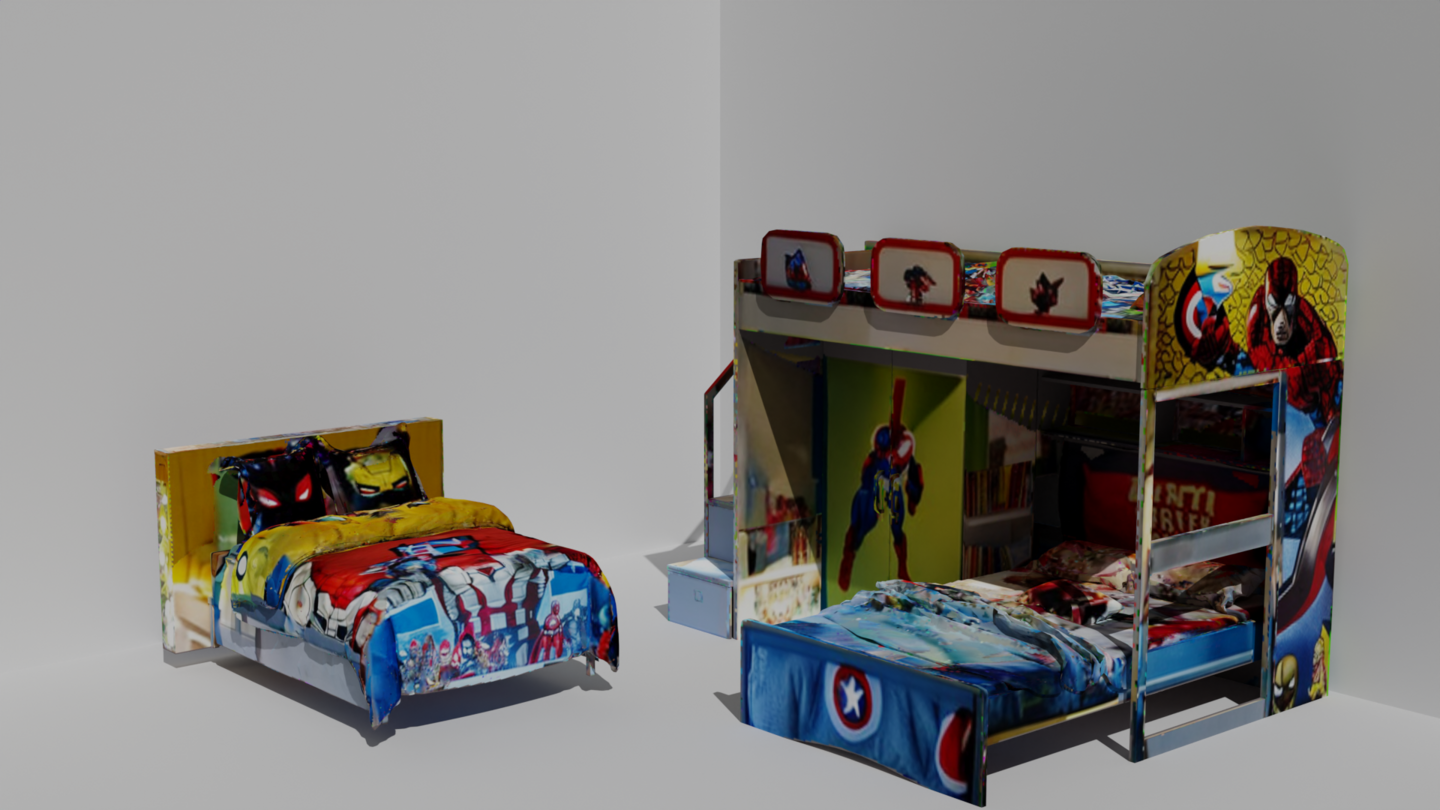}\caption{Baseline output for ``“a marvel-themed bedroom of a five-year old toddler"}\label{fig:marvel_bedroom_baseline}\end{figure*}

\subsection{($\star$) A Marvel-themed bedroom of a five-year old toddler (marvel bedroom)-- Figures \ref{fig:marvel_bedroom_ours} and \ref{fig:marvel_bedroom_baseline}}
\noindent\begin{mdframed}
Input Scene Description : \textbf { a marvel-themed bedroom of a five-year old toddler } 
Semantic shopping list
\begin{enumerate}
\item bed: red and blue colors.no signs of wear and tear, firm mattress.
\item blanket: designed with the avengers characters.bright and vibrant colors with no visible signs of wear.
\item pillows: designed with the avengers characters.bright and vibrant colors with no visible signs of wear.
\item stuffed toys: no visible signs of wear, with soft and fluffy stuffing.
\item marvel posters: with motivational quotes.crisp paper stock, with bright and vibrant colors.
\item dresser: red and blue colors.smooth finish, no signs of wear and tear.
\item basket for clothes: decorated with captain america imagery.brand new.
\item lamp: led lighting.brand new.
\item mirror: brand new.
\item decorative items: 1- iron man action figure, 1- captain america shield wall art.brand new.
\item chair: red and blue colors.no signs of wear and tear, firm supporting cushions.
\item pillow: fluffy and soft to the touch.
\item blanket: printed with comic book characters.soft and lightweight with vibrant colors.
\item marvel figures: 4-6 inches tall.pristine condition with no scratches or marks.
\item marvel posters: crisp and vibrant colors, no signs of wear.
\item marvel stickers: bright and colorful, no peeling or tearing.
\item desk: red and blue colors.smooth finish, no signs of wear and tear.
\item desk lamp: with bright colors and a cartoonish design.bright, colorful, and cartoonish design.
\item pen holder: with bright colors and cartoonish design.bright, colorful, and cartoonish design.
\item pencils: with bright colors and cartoonish design.bright, colorful, and cartoonish design.
\item notebook: with bright colors and cartoonish design.bright, colorful, and cartoonish design.
\item marvel-themed action figures: with bright colors and cartoonish design.brightly colored and cartoonish design, with no signs of wear and tear.
\item toy chest: red and blue colors.smooth finish, no signs of wear and tear.
\item stuffed animals: iron man, captain america and thor.well-loved, lightly used with minimal fading or wear.
\item action figures: iron man, captain america and thor.well-loved, lightly used with minimal fading or wear.
\item toys: marvel avengers lego kit, marvel avengers puzzle and marvel avengers playdough set.well-loved, lightly used, with minor scratches and fading.
\item books: well-loved, lightly used, with minor creases and scuffs.
\item wall art: vibrant colors.vibrant colors, no signs of fading.
\item bedding sheets: incredible hulk, thor, and captain america.brightly colored and with a soft texture.
\item wall stickers: captain america shield wall stickers, iron man wall stickers and thor hammer wall sticker with 3d effects.high quality and vibrant colors.
\item wall decals: detailed designs with 3d effects.
\item pillows: soft and fluffy with 3d effects.
\end{enumerate}
\end{mdframed}


\begin{figure*}\centering\includegraphics[width=0.9 \textwidth]{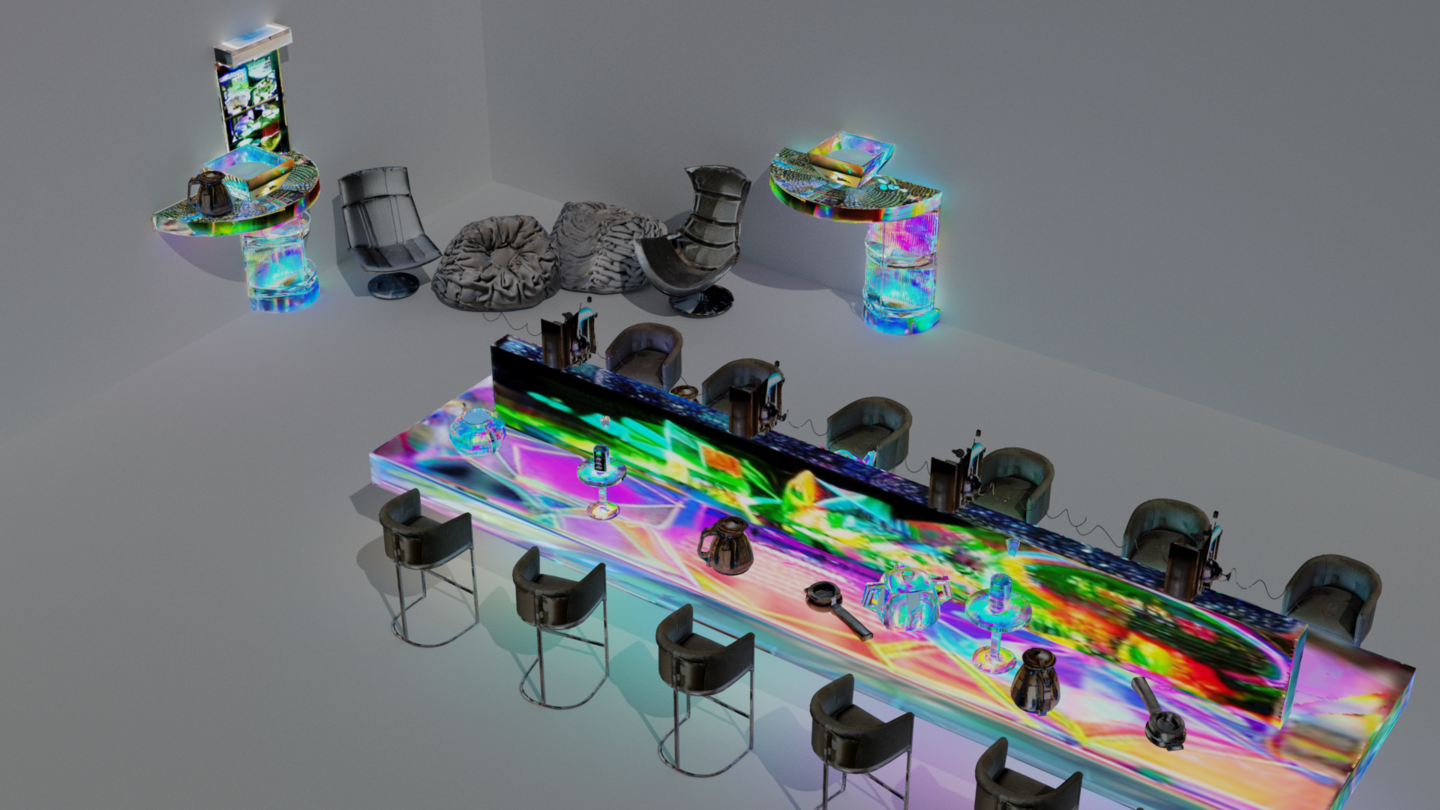}\caption{Our output for ``a futuristic teahouse from the future in silicon valley"}\label{fig:futuristic_teahouse_ours}\end{figure*}

\begin{figure*}\centering\includegraphics[width=0.9 \textwidth]{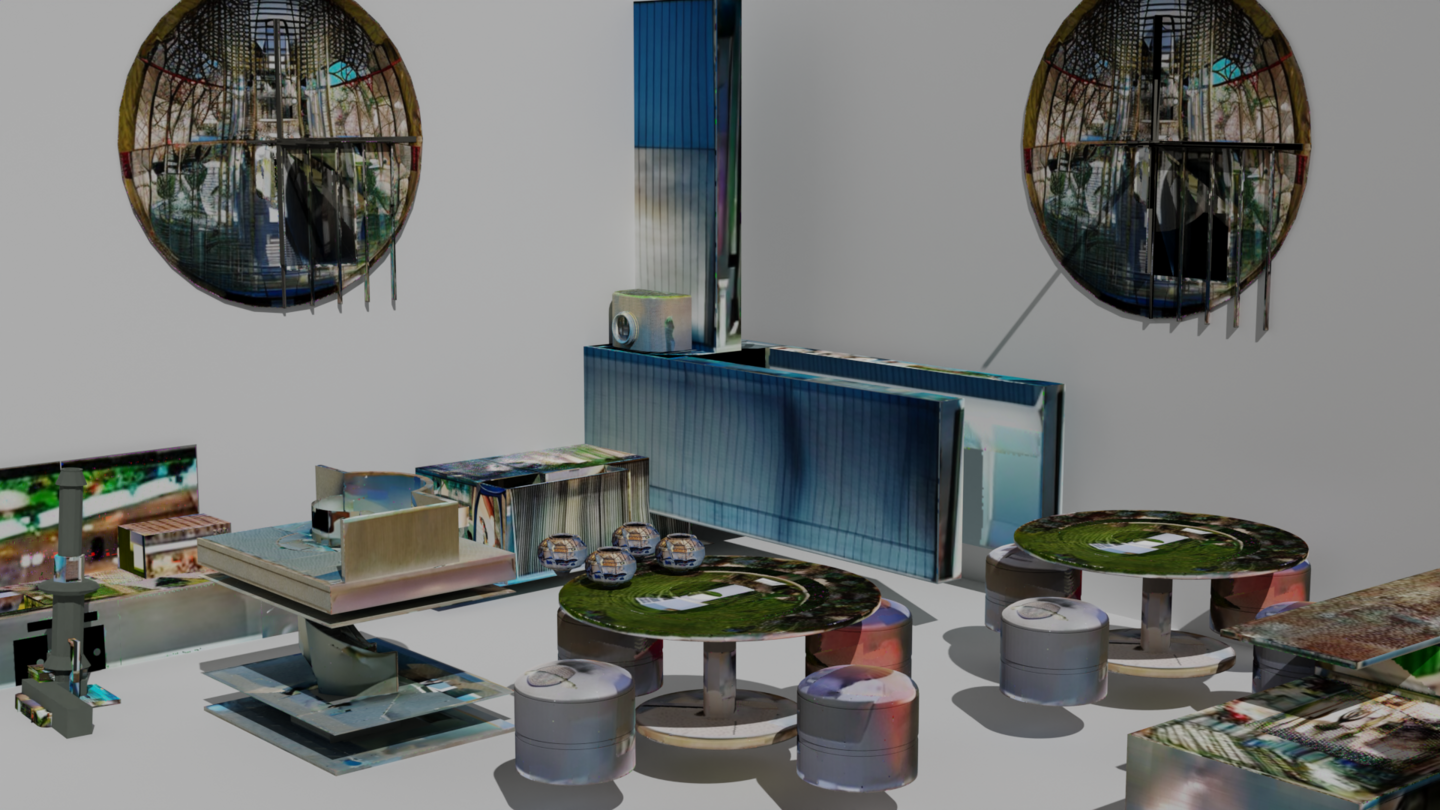}\caption{Baseline output for ``a futuristic teahouse from the future in silicon valley"}\label{fig:futuristic_teahouse_baseline}\end{figure*}

\subsection{($\star$) A futuristic teahouse from the future in silicon valley (futuristic teahouse)-- Figures \ref{fig:futuristic_teahouse_ours} and \ref{fig:futuristic_teahouse_baseline}}
\noindent\begin{mdframed}
Input Scene Description : \textbf { a futuristic teahouse from the future in silicon valley } 
Semantic shopping list
\begin{enumerate}
\item tables: with a plasma top, stainless steel legs.shiny and reflective surface.
\item tablecloths: blue and white geometric pattern.pristine condition without any creases or wrinkles.
\item plates: stainless steel with holographic accents.shiny, sleek finish with no chips or scratches.
\item silverware: made from titanium with sleek design.clean, smooth and no dents or scratches.
\item tea cups: with holographic accents.no chips or cracks, holographic accents in perfect condition.
\item holographic menus: with interactive voice activated options.crisp, clear display with no distortions or defects.
\item chairs: upholstered in light grey fabric.no signs of wear and tear.
\item arm rests: adjustable height.sleek and modern design, polished metal finish.
\item cushions: light grey in color with silver accents.soft and comfortable, with no signs of wear.
\item holographic menus: adjustable to display any type of menu.bright and vibrant colors, displaying any type of menu in 3d.
\item tea dispensers: able to dispense any type of tea.high-tech and modern design, no signs of wear.
\item bar: with curved edges and led lighting.smooth, glossy finish.
\item tea pot: sleek and shiny, with no signs of use.
\item teacups: unworn, with a holographic design shining brightly.
\item saucers: unworn, with a holographic design shining brightly.
\item holographic menu: brightly glowing, with voice activated commands responding quickly.
\item bar stools: upholstered in light grey fabric.no signs of wear and tear.
\item holographic napkins: crisp and clean with vibrant colors.
\item holographic tea cups: pristine and dust free with vibrant colors.
\item holographic saucers: pristine and dust free with vibrant colors.
\item holographic teapot: gleaming and unscratched with vibrant colors.
\item holographic displays: bright and clear images.
\item floating screens: crisp, vibrant colours and high resolution display.
\item holographic images: flickerless, with sharp edges and vivid colours.
\item interactive screens: responsive to touch and voice commands.
\item artificial intelligence (ai) bots: smooth, gliding motions with natural-sounding speech.
\end{enumerate}
\end{mdframed}


\begin{figure*}\centering\includegraphics[width=0.9 \textwidth]{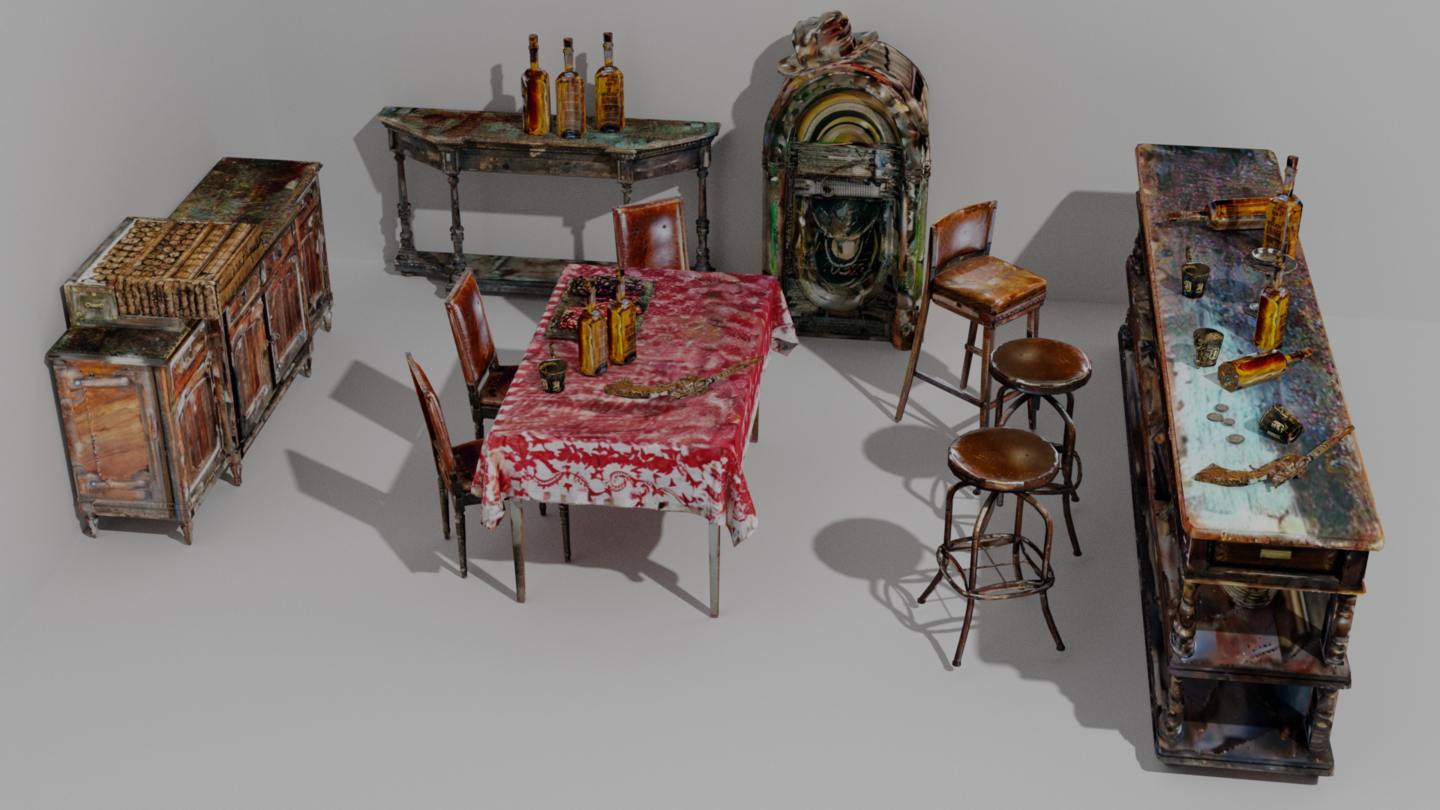}\caption{Our output for ``a saloon from an old western"}\label{fig:western_saloon_ours}\end{figure*}

\begin{figure*}\centering\includegraphics[width=0.9 \textwidth]{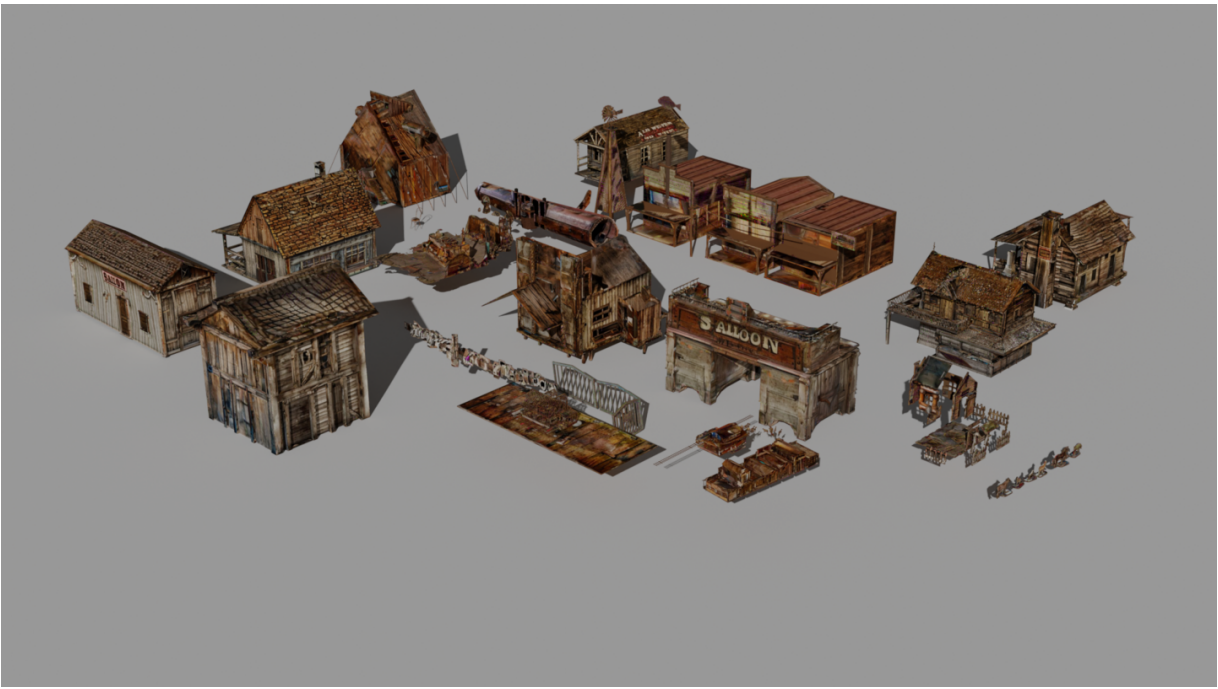}\caption{Baseline output for ``a saloon from an old western"}\label{fig:western_saloon_baseline}\end{figure*}

\subsection{($\star$) A saloon from an old western-- Figures \ref{fig:western_saloon_ours} and \ref{fig:western_saloon_baseline}}
\noindent\begin{mdframed}
Input Scene Description : \textbf { a saloon from an old western } 
Semantic shopping list
\begin{enumerate}
\item bar: dark walnut wood, brass foot rail and accents.slightly worn edges, signs of age on the finish.
\item bar stools: wooden frames, leather seat cushions.worn, distressed wood finish.
\item bottles: clear glass.slightly dusty.
\item glasses: faceted crystal glass.etched, lightly scratched.
\item bar counter: metal foot rail.weathered, aged patina.
\item chairs: upholstered in leather, with metal or wooden frames.soft leather, with a few scuff marks.
\item cowboy hats: with weathered bandanas on the sides.slightly dusty, with signs of wear.
\item cowboy boots: with weathered stitching.scuffed and worn with age.
\item cigars: slightly weathered and dry.
\item whisky bottle: aged, with signs of condensation.
\item tables: wooden legs and a square top.aged wood with a rustic finish, some minor scratches.
\item tablecloth: red and white.slightly frayed on the edges.
\item beer mugs: black with gold accents.rustic finish, showing signs of wear and tear.
\item playing cards: slightly creased edges.
\item poker chips: white with red and black accents.lightly worn, with minor scratches.
\item counter: wood and metal accents.aged wood with a few dents and scratches.
\item whiskey bottle: clear glass and brown stopper.dull glass with some scratches and dull brown stopper.
\item shot glasses: clear glass and gold rims.slightly worn edges and faded gold rims.
\item poker chips: red and black with gold accents.some chips are slightly faded and chipped.
\item ashtray: made of bronze and silver accents.some tarnishing on the bronze and silver accents.
\item jukebox: bright colors, and lights.minor wear and tear, with a few chips in the paint.
\item music discs: slightly worn edges, but still in good condition.
\item coins: slightly tarnished, but still in good condition.
\item posters: slightly faded, but still in good condition.
\item stools: dark brown finish.slightly worn edges, but still in good condition.
\end{enumerate}
\end{mdframed}


\begin{figure*}\centering\includegraphics[width=0.9 \textwidth]{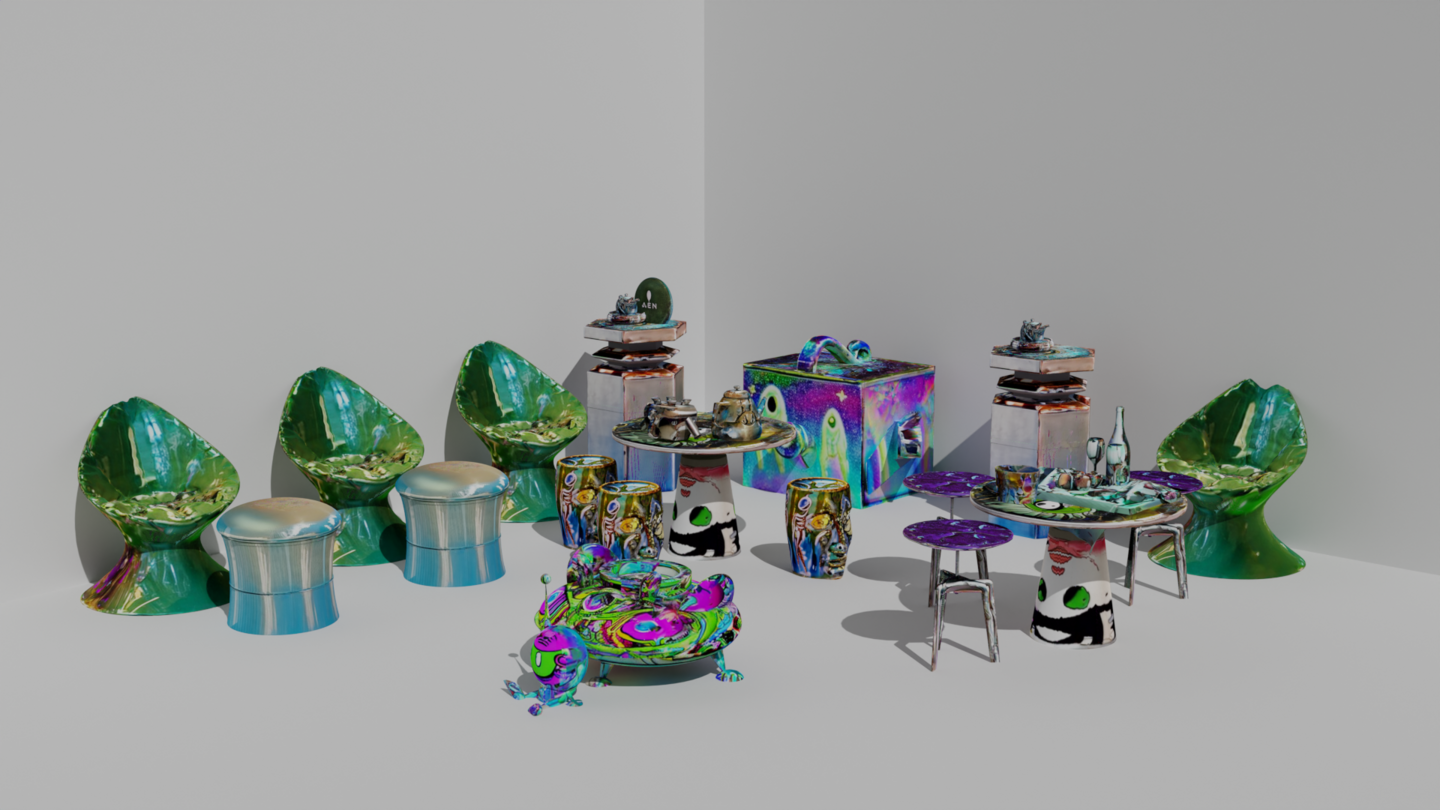}\caption{Our output for ``an alien tea garden on mars"}\label{fig:alien_teagarden_ours}\end{figure*}

\subsection{An alien tea garden on Mars (alien teagarden)-- Figure \ref{fig:alien_teagarden_ours}}
\noindent\begin{mdframed}
Input Scene Description : \textbf { an alien tea garden on mars } 
Semantic shopping list
\begin{enumerate}
\item alien tea table: metallic finish in a deep purple hue.smooth, glossy finish.
\item alien tea pot: copper metal with silver accents.shiny, polished finish.
\item alien tea cups: glazed ceramic with green and blue accents.smooth and glossy.
\item alien tea spoons: silver metal with gold accents.no signs of wear.
\item alien tea plates: glazed ceramic with blue and purple accents.no signs of wear.
\item alien tea chairs: bright green hue with a glossy finish.no signs of wear.
\item alien tea tablecloths: unblemished and pristine condition.
\item alien tea cups: smooth and glossy finish.
\item alien tea saucers: featuring holographic images.shiny and reflective surface.
\item alien tea kettles: with a sleek design and a futuristic handle.gleaming and metallic look.
\item alien tea bar: dark blue hue with a matte finish.no signs of wear.
\item alien tea kettle: with an alien-shaped handle.brand-new, sparkling with glitter.
\item alien tea pot: with a star-shaped lid and alien designs on it.brand-new, shimmering with iridescent hues.
\item alien tea cups: featuring alien designs on them.brand-new, shining with crystal clarity.
\item alien tea saucers: featuring alien designs on them.brand-new, bright with vibrant colors.
\item alien tea bar stools: light blue hue with a glossy finish.no signs of wear.
\item alien tea cups: with gold accents and a cartoon alien motif.smooth, with no chips or scratches.
\item alien tea saucers: with gold accents and a cartoon alien motif.smooth, with no chips or scratches.
\item alien tea kettle: with a spout and handle in the shape of an alien head.bright and shining, with no rust or dents.
\item alien tea lights: with alien faces on the front.unused and unwaxed, with bright green glow.
\end{enumerate}
\end{mdframed}

\begin{figure*}\centering\includegraphics[width=0.9 \textwidth]{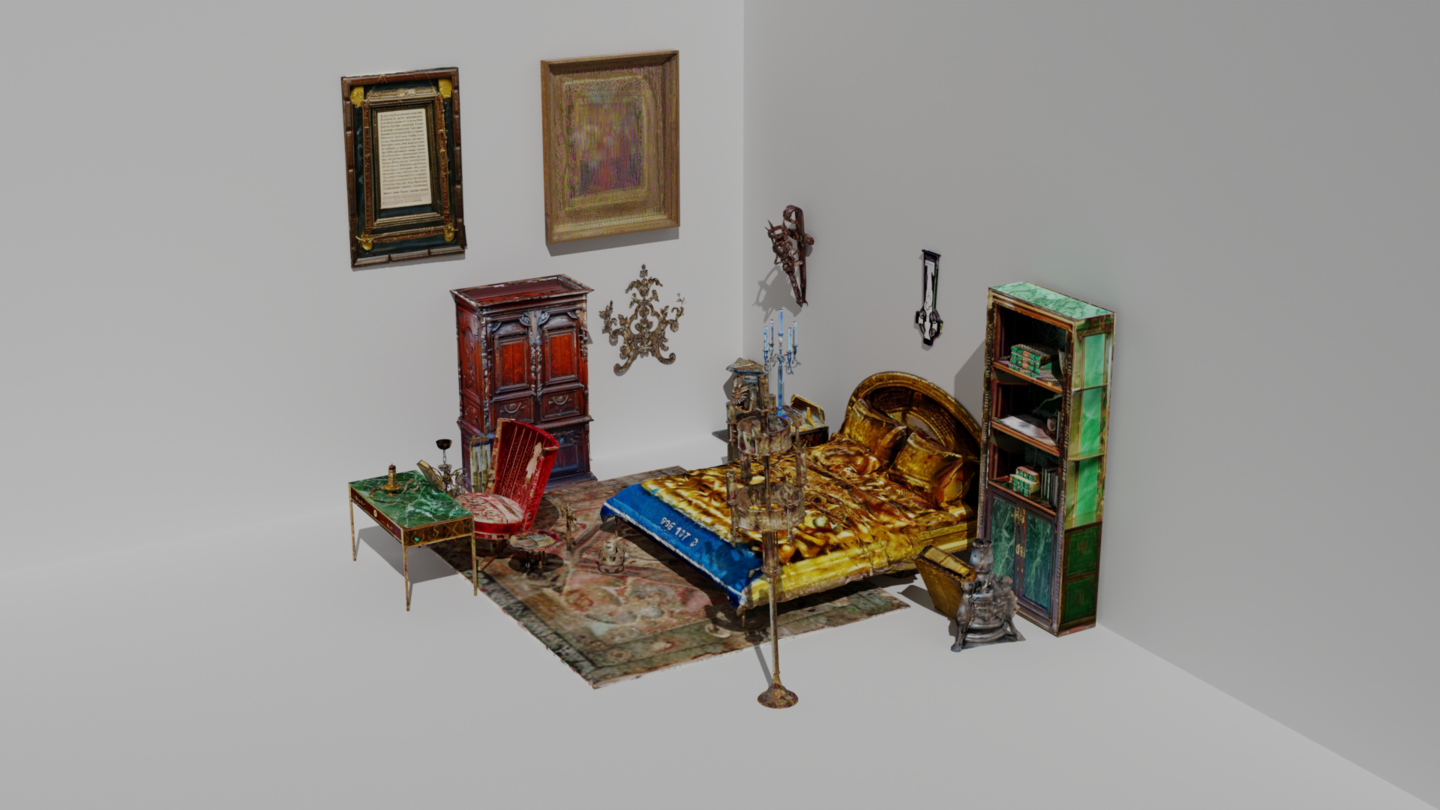}\caption{Our output for ``antichrist in the bedroom of the pope"}\label{fig:antichrist_vatican_ours}\end{figure*}

\subsection{Antichrist in the bedroom of the pope (antichrist vatican)-- Figure \ref{fig:antichrist_vatican_ours}}
\noindent\begin{mdframed}
Input Scene Description : \textbf { antichrist in the bedroom of the pope } 
Semantic shopping list
\begin{enumerate}
\item bed: crafted with dark, solid wood and royal blue velvet drapes.solid, with slight signs of age.
\item pillows: black with golden embroidery.perfectly plump and inviting.
\item blanket: black with golden trim.soft and smooth to the touch.
\item nightstand: dark wood with gold accents.polished and gleaming.
\item lamp: black iron with gold details.intricate and delicate.
\item bible: black leather with gold lettering.slightly worn on the edges.
\item crucifix: black iron with gold details.detailed and gleaming.
\item desk: crafted with mahogany wood, gold accents and a green marble top.smooth, polished finish with subtle gold accents.
\item inkwell: made of iron, with a burnished finish.smooth surface, no signs of corrosion.
\item quill: gold plated, with a feathery texture.sharpened to a point, no visible fraying.
\item book: with intricate designs on the cover.frayed edges, some pages yellowing with age.
\item papers: handmade and decorated with gold leaf.intricate designs still visible, slight discoloration on the edges.
\item lamp: made of brass, with a matte finish.strong and sturdy with no visible dents.
\item chair: upholstered in red velvet with slight signs of wear.
\item armrests: made of dark wood, with intricate carvings.smooth, with no scratches or dents.
\item pillows: made of velvet with gold accents.soft and plush to the touch.
\item bible: with silver accents and a velvet cover.perfectly preserved with no signs of wear.
\item crucifix: unblemished and shining.
\item candles: with ornate holders.perfectly shaped and in pristine condition.
\item bookshelf: crafted with mahogany wood, gold accents and a green marble top.smooth, polished finish with subtle gold accents.
\item books: pages slightly yellowed with age.
\item cross: glint of gold in the details and worn edges.
\item bible: well-worn with traces of use.
\item candelabra: intricate details with slight signs of age.
\item candlesticks: ornate design.polished and free of tarnish.
\item candle: scented red roses.unburned, with the scent of red roses.
\item incense burner: dragon motif.smooth, with intricate designs.
\item incense sticks: unburned, with the scent of sandalwood.
\item bell: with inlaid designs.unrusted, with intricate inlay designs.
\item rug: intricate patterns and colors with slight fading.
\item cross: intricate detailing, polished finish.
\item crown of thorns: aged thorns and a tarnished crown.
\item bible: discoloration on the edges and cover.
\item chalice: intricate detailing, polished finish.
\item candle: wax burn marks and a tarnished candlestick.
\item incense burner: intricate detailing, polished finish.
\item crucifix: intricate detailing, polished finish.
\item rosary: intricate detailing, polished finish.
\item painting: vibrant colors and intricate details, slightly aged.
\item frame: polished finish, but with some signs of age.
\item mirror: glossy finish, but with some signs of age.
\item bible: bound with gold leather.binding intact, with some signs of age.
\item crucifix: made of black iron.black iron, with some signs of oxidation.
\item candles: with gold accents.smooth finish with signs of usage.
\item incense: in a gold incense holder.unscented, but with signs of age.
\item chandelier: gold accents.crystal clear, with gold accents, slightly faded.
\item candles: slightly melted and slightly charred.
\item chain: slightly rusted and with a few dents.
\item bells: slightly tarnished and with a few dents.
\item crosses: slightly weathered with a few scratches.
\end{enumerate}
\end{mdframed}

\begin{figure*}\centering\includegraphics[width=0.9 \textwidth]{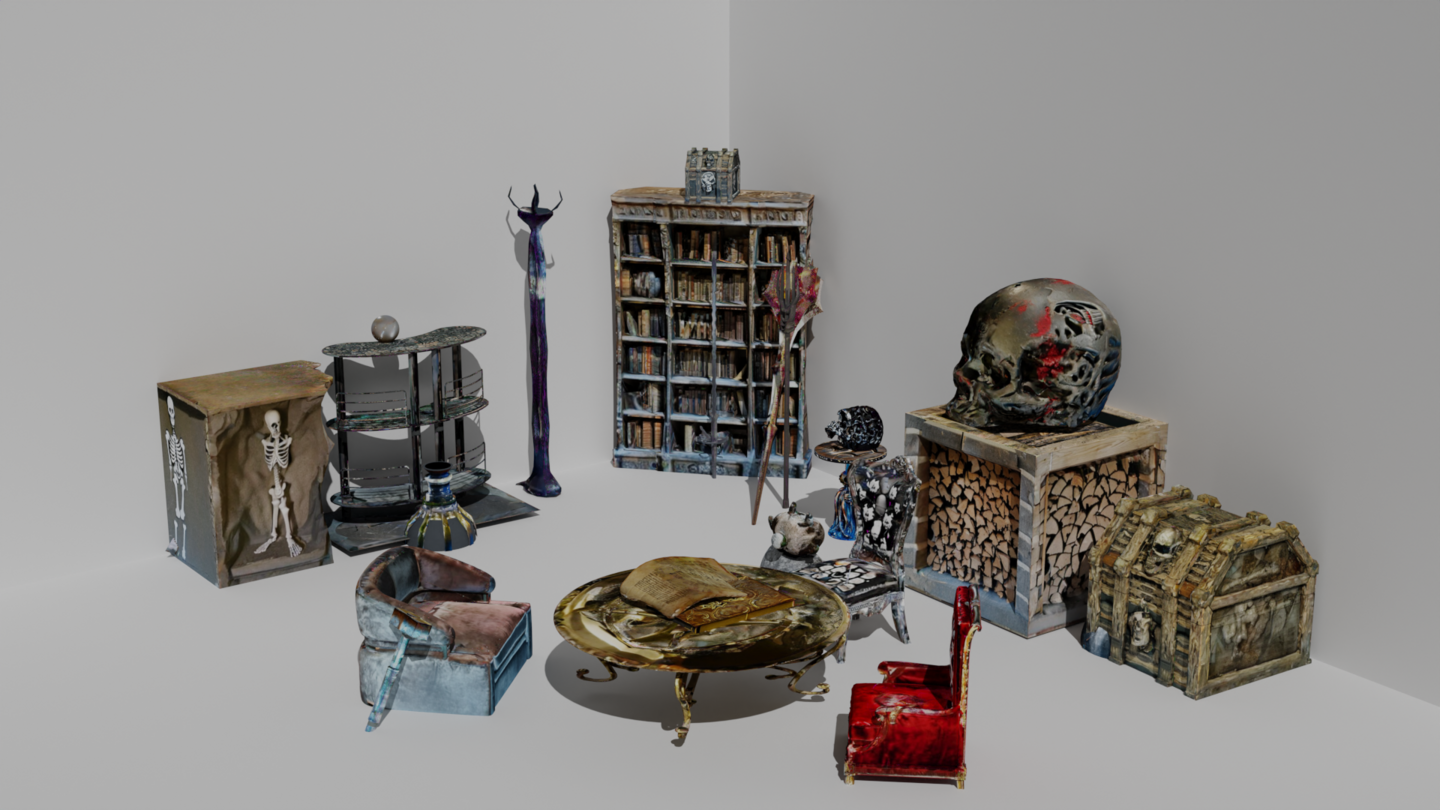}\caption{Our output for ``Hade's man cave"}\label{fig:hades_cave_ours}\end{figure*}

\subsection{Hades's man cave (hades cave)-- Figure \ref{fig:hades_cave_ours}}
\noindent\begin{mdframed}
Input Scene Description : \textbf { hades's man cave } 
Semantic shopping list
\begin{enumerate}
\item throne: with a raised backrest, and skulls decorated on the sides.smooth, polished metal.
\item red velvet throne cover: pristine condition.
\item skeleton: slightly aged but in pristine condition.
\item crown: slightly tarnished but still gleaming.
\item chain: slightly worn with a few scratches.
\item sword: slightly rusted but still sharp.
\item scepter: a few scratches and a bit of tarnish.
\item scroll: yellowed parchment scroll with ancient writings.slightly faded but still legible.
\item bat: black bat with glossy wings.slightly worn but still glossy.
\item pit of despair: flickering flames and smoky mist.
\item chains: blackened and rusted.heavy duty metal chains, heavily rusted with a few missing links.
\item skeletons: hand painted in a metallic black with red highlights.skeleton props, intact with a few parts being chipped away due to age.
\item torches: antiqued metal torches, slightly worn with a few patches of rust.
\item kraken statue: aged bronze with a glossy finish.
\item tentacles: made of soft rubber and flexible enough to be manipulated into different shapes.flexible and smooth.
\item sea shells: curved sea shells, with natural colours and raised patterns.shiny and slightly transparent.
\item rocks: decorative rocks, with rough texture and dark grey colour.rough and textured.
\item pearls: white pearls, with a glossy finish.smooth and glossy.
\item bookshelf: with intricate carvings.aged and weathered, with intricate details still visible.
\item books: bound in dark leather with golden embossed designs.perfectly preserved and complete.
\item scrolls: embossed with ancient runes.slightly weathered with age.
\item potions: made of dark glass with ornate cork stoppers.intact, with no signs of tampering.
\item fire pit: with a black iron surround.rustic and sturdy, with a protective black iron surround.
\item firewood: pre-split for convenience.dry and ready to burn.
\item fireplace tools: wrought iron with black finish.glossy black finish with no visible rust or wear.
\item fireplace grate: black finish.no visible rust or wear.
\item fire poker: no visible rust or wear.
\item fireplace bellows: no visible wear or tear on the leather.
\item cauldron: with a black iron handle.smooth, heavy cast iron with a black iron handle.
\item potions: containing a mysterious black liquid.perfectly preserved and sealed.
\item oils: fragrant and fresh.
\item bones: carefully arranged in the chest and slightly aged.
\item chair: with a velvet cushion.black iron frame, with a velvet cushion slightly worn.
\item skulls: painted black.smooth and glossy finish.
\item goatskin rug: dyed black and red.soft and supple, with no signs of wear.
\item crossbones: painted black.no signs of wear.
\item torch: sculpted with skull and bones, brass finish.smooth and glossy, with no signs of wear.
\item anvil: with a skull etched into it.heavy and solid, with a skull etched into it.
\item hammer: with an 8-inch handle and a 4-inch head.slightly worn handle and a clean head.
\item tongs: slight signs of rust on the flat jaw and a sturdy handle.
\item iron rods: clean and free from dents.
\item iron ingots: no signs of rust.
\item forge: with a charcoal-fired box and adjustable air intake and exhaust.well-maintained, with no signs of damage.
\end{enumerate}
\end{mdframed}

\begin{figure*}\centering\includegraphics[width=0.9 \textwidth]{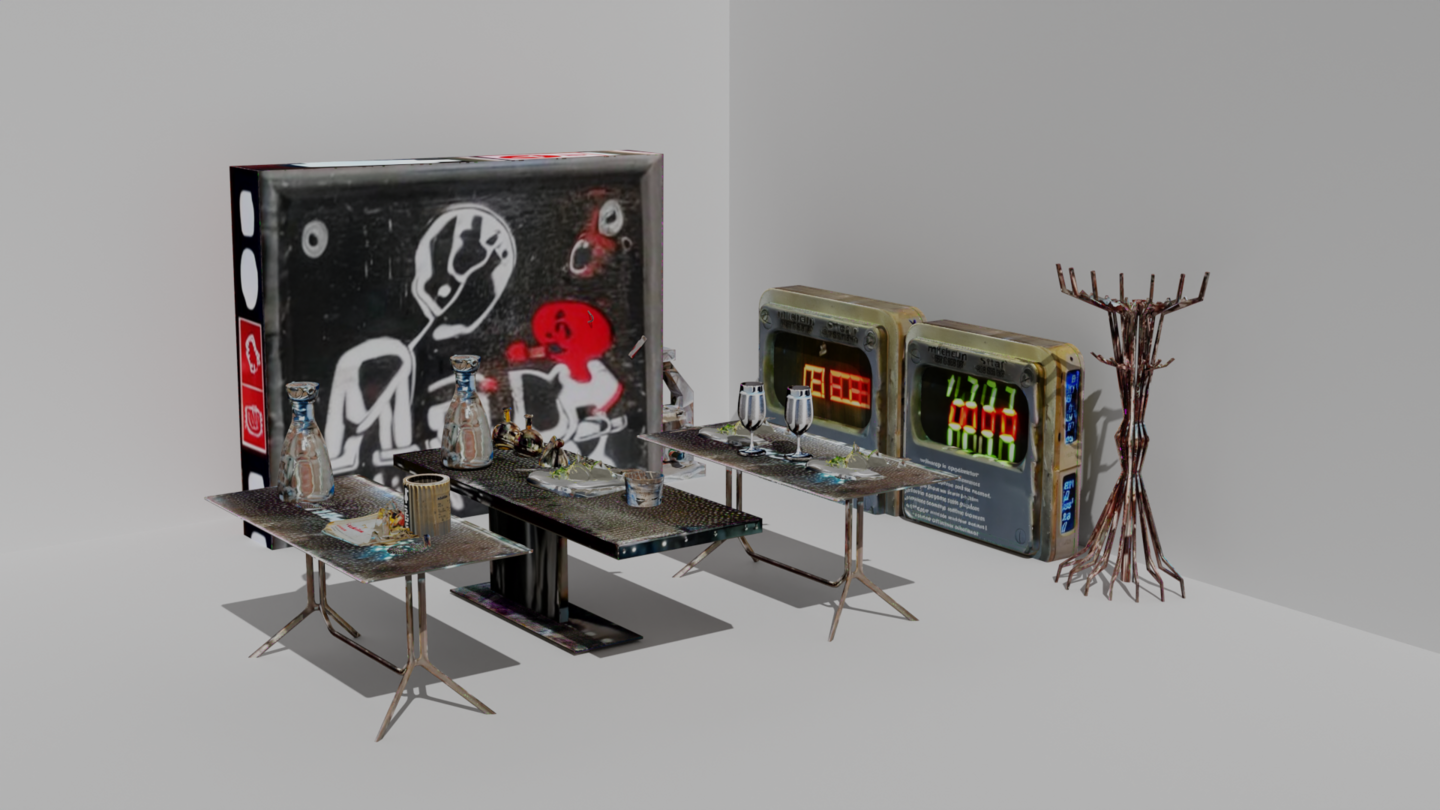}\caption{Our output for ``a michelin star restaurant opened by a mad scientist"}\label{fig:mad_scientist_restaurant_ours}\end{figure*}

\subsection{A michelin star restaurant opened by a mad scientist (mad scientist restaurant)-- Figure \ref{fig:mad_scientist_restaurant_ours}}
\noindent\begin{mdframed}
Input Scene Description : \textbf { a michelin star restaurant opened by a mad scientist } 
Semantic shopping list
\begin{enumerate}
\item tables: black steel with rivets.clean and glossy finish.
\item tablecloths: crisp and wrinkle free.
\item plates: smooth and glossy.
\item silverware: shiny and unscratched.
\item wine glasses: crisp and free of smudges.
\item petri dishes: clean and free of dust.
\item chairs: white plastic with black accents.unused and free of scratches.
\item table linen: crisp, clean and wrinkle free.
\item plates: spotless and shiny.
\item silverware: shining and polished.
\item test tubes: bright and clear.
\item test tube holders: with a copper finish.clean and without corrosion.
\item counter: stainless steel with led lighting.unused and free of scratches.
\item kitchenware: pans, baking trays, mixers, etc. with a steampunk design.sparkling, stainless steel with engraved designs.
\item bottles with potion: glass apothecary bottles with cork stoppers.sealed with a tight cork stopper.
\item cookbooks: slightly worn, yellowed pages with colorful illustrations.
\item scientist's lab coat: slightly worn, with ragged edges and loose threads.
\item test tubes: clean and free of cracks.
\item test tube rack: 8-test tube capacity.clean and clear, with no signs of corrosion or rust.
\item flasks: 50ml capacity, glass and rubber stoppers.glass containers with clear markings, with no chips or cracks.
\item chemical containers: graduated cylinders.no leaks, no signs of discoloration or damage.
\item dropper: 1ml capacity, with flexible bulb.no cracks or leaks, with a good seal.
\item beakers: clean and free of cracks.
\item chemical flasks: with graduated markingsclean and untouched.
\item beaker racks: powder coatedclean and in new condition.
\item safety goggles: adjustable strapsin perfect condition, free of scratches.
\item lab coats: three-button front closurefreshly laundered and wrinkle-free.
\item pipettes: graduated markings, with rubber tipssterile, free of dirt or dust.
\item bunsen burner: unused and in perfect working order.
\item flask: borosilicate glass.gleaming and sparkling, with no scratches.
\item feed tube: no kinks or bends.
\item gas valve: clean, no rust on the connections.
\item pressure gauge: no scratches, well calibrated.
\item ignition switch: free of dust and grime, working properly.
\item burner pipe: no rust, no dents, free of soot.
\item electrodes: unused and free of rust.
\item wires: insulated, 20 awgno visible signs of damage.
\item voltmeter: lcd displayno visible signs of damage.
\item ammeter: lcd displayno visible signs of damage.
\item switch: red and blackno visible signs of damage.
\item magnifying glass: plastic handleno visible scratches and smudges.
\item safety goggles: adjustable headbandno visible signs of wear and tear.
\end{enumerate}
\end{mdframed}

\begin{figure*}\centering\includegraphics[width=0.9 \textwidth]{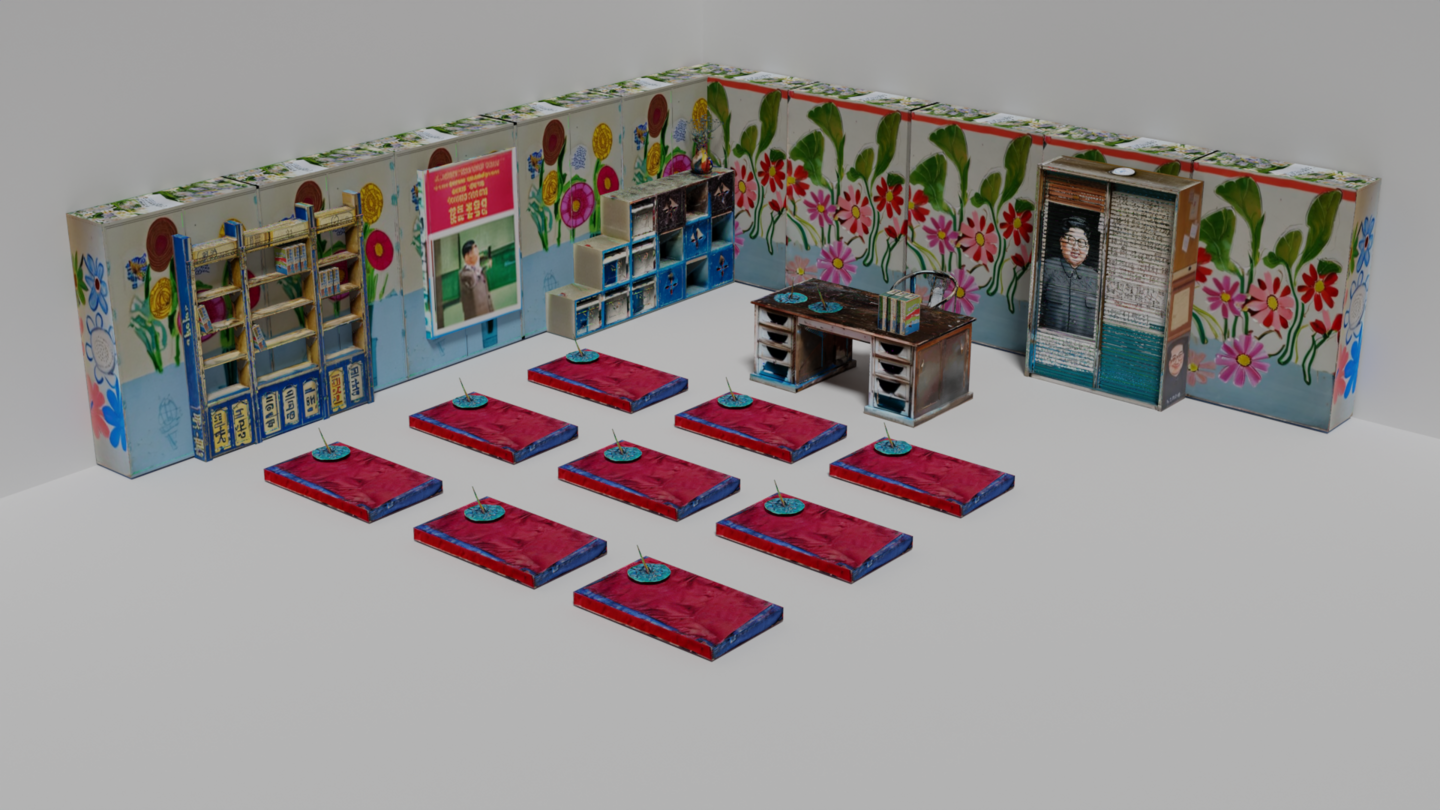}\caption{Our output for ``a north korean classroom"}\label{fig:north_korean_classroom_ours}\end{figure*}

\subsection{A North Korean classroom (north  korean classroom) -- Figure \ref{fig:north_korean_classroom_ours}} 
\noindent\begin{mdframed}
Input Scene Description : \textbf { a north korean classroom } 
Semantic shopping list
\begin{enumerate}
\item desks: made of wood with black paint finish.smooth, dust-free surfaces.
\item pencils: sharpened and ready for use.
\item books: new and in great condition.
\item erasers: new, unused.
\item notebooks: brand new and unused.
\item chairs: made of wood with black paint finish.slight signs of wear on the edges.
\item student uniform: navy blue with red accents.neatly pressed and ironed.
\item pencils: black lead, white erasers on the top.newly sharpened.
\item books: navy blue with gold accents.minimal signs of wear.
\item notebooks: navy blue with red accents.neatly stacked and organized.
\item teacher's desk and chair: some signs of wear on the cushioning of the chair.
\item vase with flowers: with paper flowers.clean and well taken care of.
\item globe: with north korea highlighted in gold.smooth and glossy finish.
\item school textbooks: philosophy and political science textbooks.brand new and untouched.
\item blackboard: with a portrait of kim il sung in the center.spotless and polished.
\item chalk: in white and red.freshly sharpened.
\item bookshelf: metal frame with wooden shelves.no signs of rust or wear.
\item books: north korea's official state doctrine, printed on thick off-white paper with blue binding.crisp pages and spines, in new condition.
\item artifacts: made of bronze and gold-plated.shining bronze and gold plating, highly polished.
\item posters: printed on thick cardstock with vibrant colors.crisp and vibrant colors, free of creases.
\item flag: made of polyester and printed with bold colors.unfaded colors, free of wrinkles.
\end{enumerate}
\end{mdframed}

\begin{figure*}\centering\includegraphics[width=0.9 \textwidth]{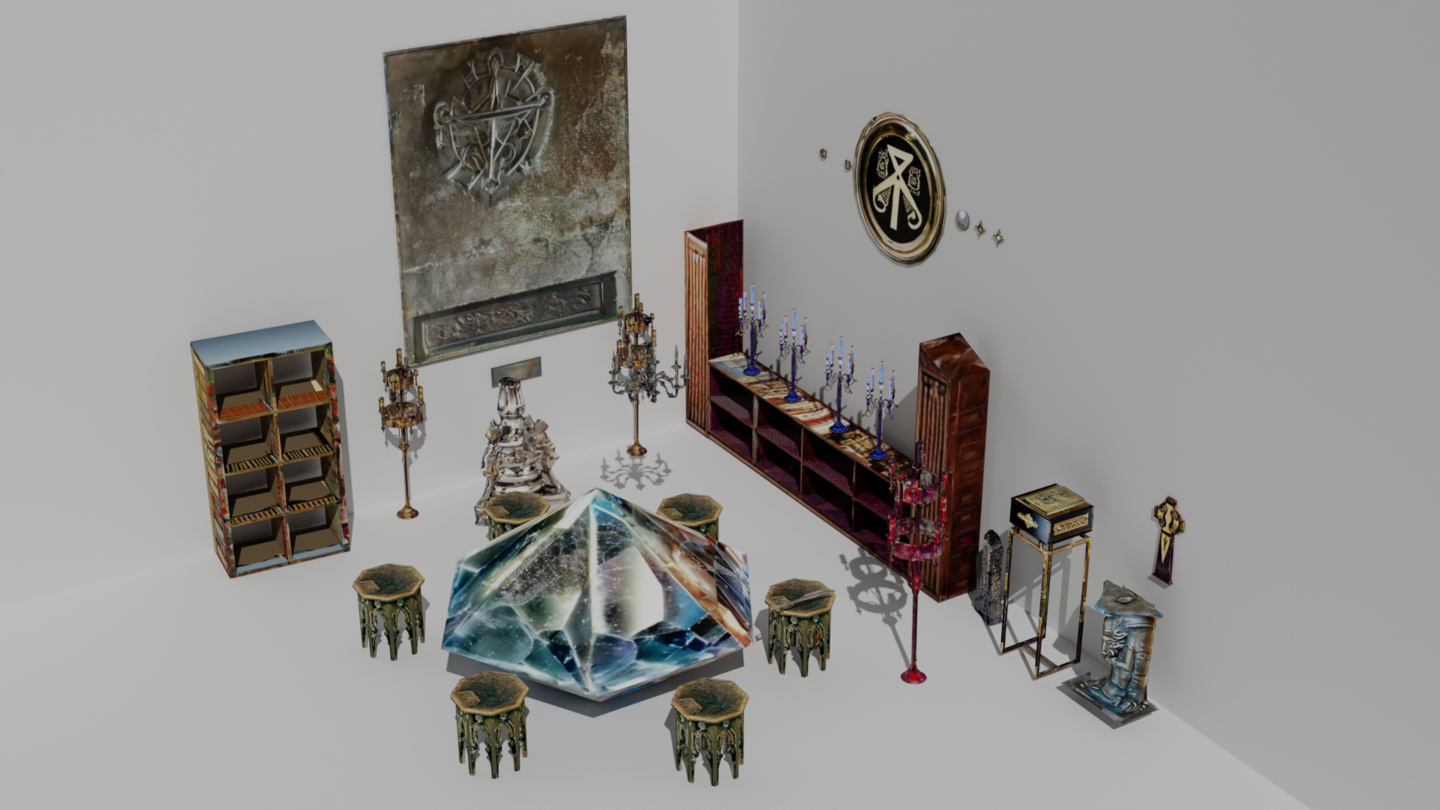}\caption{Our output for ``headquarters of an occult cult"}\label{fig:occult_cult_ours}\end{figure*}

\subsection{Headquarters of an occult cult (occult cult) -- Figure \ref{fig:occult_cult_ours}} 
\noindent\begin{mdframed}
Input Scene Description : \textbf { headquarters of an occult cult } 
Semantic shopping list
\begin{enumerate}
\item altar: polished finish with no visible signs of wear.
\item ritual items: shiny and polished surface.
\item candelabra: black matte finish with five wax candles lit.
\item incense burner: polished silver surface with a hint of smoky scent.
\item ritual objects: smooth, dark surface with intricate symbols etched on them.
\item candles: flawless appearance, no signs of burning.
\item incense: 1 lb.fresh and aromatic.
\item incense burner: polished, clean and free of tarnish.
\item bell: 8 inches in diameter, ornate design on the handle.no scratches or dents.
\item dagger: sharp blade, no nicks.
\item robe: made of velvet fabric.undamaged and without stains.
\item chairs: no signs of wear or tear.
\item candelabras: black iron with red accents.slight tarnishing on the metal.
\item ritual tools: bronze and silver with intricate designs.slightly worn down edges.
\item mystic symbols: carved in onyx stone, painted with gold accents.crisp, sharp lines.
\item bookshelves: no visible signs of wear.
\item books: slightly worn and faded cover.
\item scrolls: slightly aged and fragile.
\item crucifix: intricate engravings, solid and well-crafted.
\item chalice: gleaming silver finish, ornate engravings.
\end{enumerate}
\end{mdframed}

\begin{figure*}\centering\includegraphics[width=0.9 \textwidth]{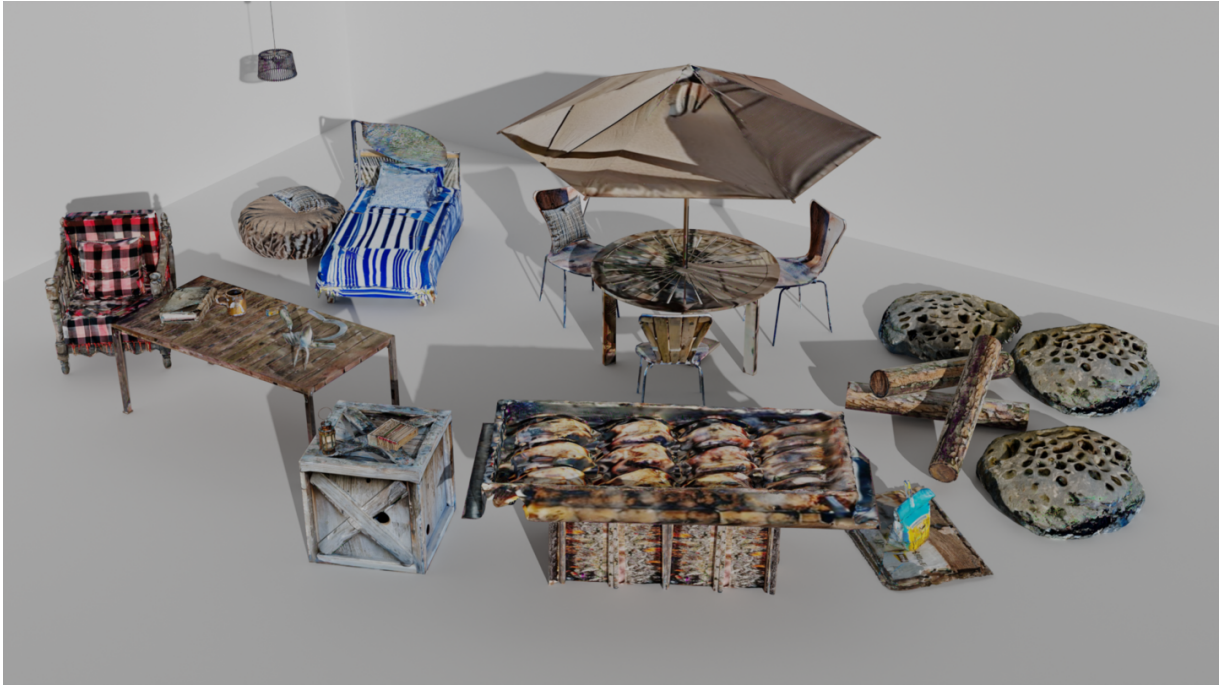}\caption{Our output for ``a rustic backyard in the countryside"}\label{fig:rustic_backyard_ours}\end{figure*}

\subsection{A rustic backyard in the countryside  (rustic backyard) -- Figure \ref{fig:rustic_backyard_ours}} 
\noindent\begin{mdframed}
Input Scene Description : \textbf { a rustic backyard in the countryside } 
Semantic shopping list
\begin{enumerate}
\item patio furniture: dark brown with white cushionsslightly faded from sun exposure.
\item umbrella: beige fabric, wooden pole and ribs.no visible wear or tear.
\item pillows: weather-resistant material with a rustic pattern.no visible wear or tear.
\item chairs: wooden material and natural finish.slight signs of fading on the wood.
\item cooler: canvas material and rustic pattern.no visible wear or tear.
\item sunscreen: spf 50 protection.unopened and not expired.
\item picnic table: dark brownslightly weathered from outdoor elements.
\item picnic blanket: cotton and red and black plaid.slight wear and tear but still intact.
\item plates: rustic style with floral pattern and earthy colors.minor scratches and chips but still intact.
\item silverware: no damage or wear and tear.
\item firepit: well maintained and clean.
\item firewood: 16-18 inches in length.rough texture, split into manageable pieces.
\item logs: 8-10 inches in diameter.bark peeled but still rough to the touch.
\item rocks: at least 12 inches in diameter.smooth, slightly weathered surfaces.
\item firestarter: eco-friendly material.lightweight, dry material.
\item matches: waterproof container.unused, undamaged boxes.
\item grill: stainless steel with black accentsslightly weathered from outdoor elements.
\item charcoal: 10lbs.packed in a sturdy box.
\item lighter fluid: 64oz.sealed tightly in a plastic container.
\item grill utensils: stainless steel with wooden handles.rustic appearance with slight signs of wear.
\item grilling food: freshly packaged.
\item hammock: blue and white stripesslightly faded from sun exposure.
\item blanket: cream and sky blue, with fringed hem.soft and fluffy, slightly faded from the sun.
\item pillow: plump and full of cozy filling.
\item books: with worn covers.slightly bent from use, with creased covers.
\item mug: in a natural ochre color.slightly worn from use, with a few scratches on the surface.
\item lantern: made of copper and glass.rusty and aged from being in the elements, with a few chips in the glass.
\item trees: about 5-6 feet tallthick, full foliage.
\item birdhouses: white-washed finish.smooth, worn surface with age-related discoloration.
\item birdfeeders: white-washed finish.smooth, worn surface with age-related discoloration.
\item wildflowers: chosen for their color and texture.vibrant colors and lush texture.
\item potted plants: chosen for their hardiness and beauty.natural patina and signs of weathering.
\end{enumerate}
\end{mdframed}

\begin{figure*}\centering\includegraphics[width=0.9 \textwidth]{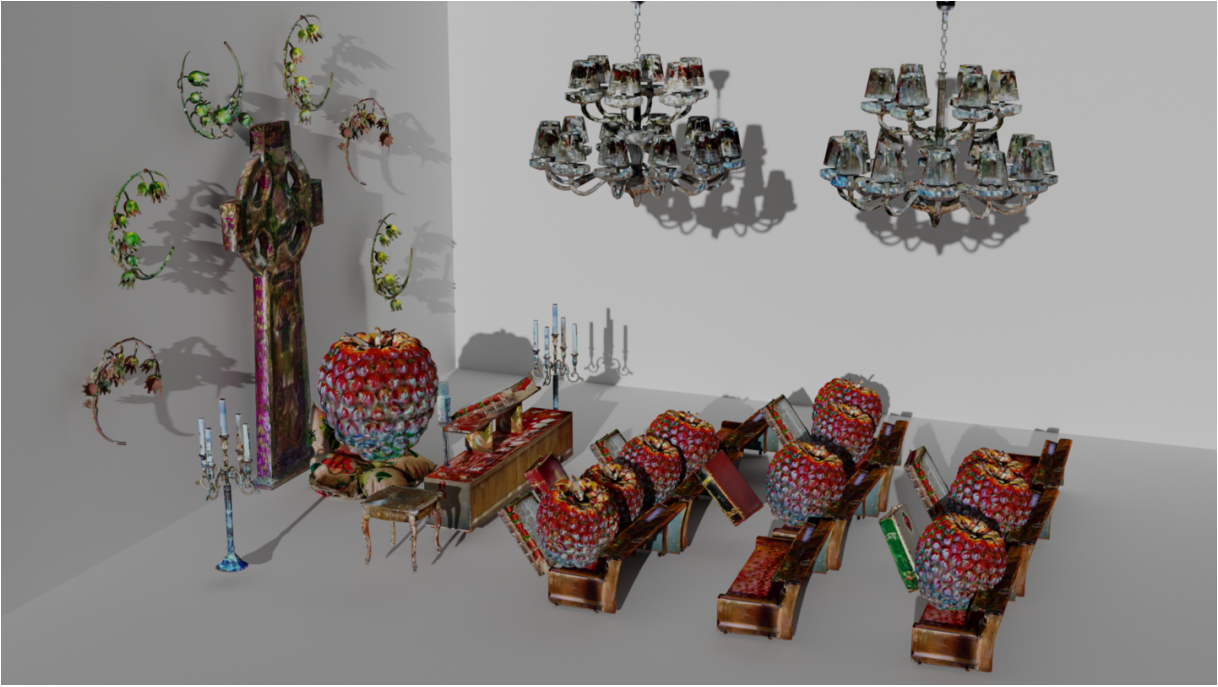}\caption{Our output for ``a church for strawberries"}\label{fig:strawberry_church_ours}\end{figure*}

\subsection{A church for strawberries (strawberry church) -- Figure \ref{fig:strawberry_church_ours}}
\noindent\begin{mdframed}
Input Scene Description : \textbf { a church for strawberries } 
Semantic shopping list
\begin{enumerate}
\item altar: smooth, polished finish.
\item candles: ivory white in color.bright and evenly burning flames.
\item cross: slightly tarnished, but still gleaming with an antique feel.
\item bible: pages appear crisp and clean.
\item flowerpots: glazed with freshly painted designs.
\item incense burner: shiny and polished with no signs of scratches.
\item pews: slight signs of wear on the arms.
\item cushions: embroidered with delicate golden strawberry patternscrisp and clean, without any signs of wear or fading.
\item hymnals: with a strawberry-themed cover designbrand new with no signs of wear or damage.
\item candles: scented with a sweet strawberry fragranceunlit, with no signs of use.
\item pulpit: intricate carvings and polished marble top in perfect condition.
\item bible: burgundy leather binding, gold lettering.brand new, no signs of wear.
\item microphone: wireless microphone with an adjustable stand.no signs of wear and fully functional.
\item flowers: fresh, vibrant petals.
\item choir books: brand new, no signs of wear.
\item candles: bright and vibrant, with no signs of smoke.
\item chandelier: all crystal drops intact, with no discolouration.
\item candles: unburned and well-preserved.
\item banners: crisp and clean edges.
\item candelabras: bright and lustrous finish.
\item incense burner: smooth and polished finish.
\item statue: intricate detailing in perfect condition.
\item candles: 6 inches tall.no wax drips, burning softly.
\item flowers: long stem.fresh, with long stems and large blooms.
\item incense burner: with a strawberry design.shiny, with a detailed strawberry design.
\item prayer beads: smooth to the touch, with a sturdy string.
\item bible: with a red cover and a white strawberry embossed on the front.new, with crisp pages and a glossy cover.
\item stained glass windows: bright colours with no fading or discolouration.
\item candles: lightly scented, with red flower petals around them.new, unscented, with red petals scattered around them.
\item banners: with the words 'love is sweet' written in gold lettering.crisp, with no signs of wear.
\item flowers: with green foliage and red ribbons.fresh, with vibrant colors and no signs of wilting.
\end{enumerate}
\end{mdframed}

\begin{figure*}\centering\includegraphics[width=0.9 \textwidth]{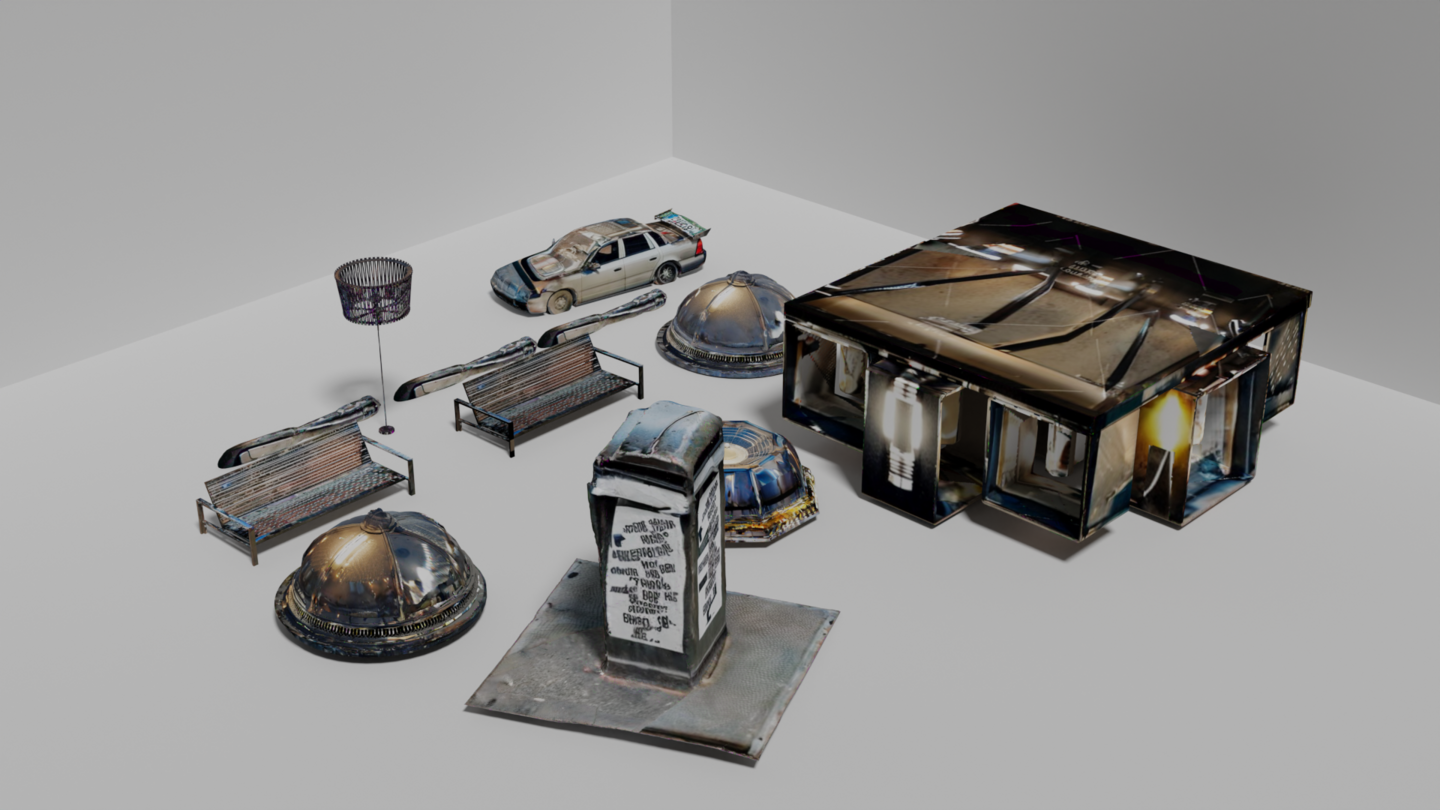}\caption{Our output for ``a busy street in downtown new york". We consider this a failure case, and see the improvement of our system for busy urban outdoor scenes as exciting future work. }\label{fig:busy_new_york_street_ours}\end{figure*}

\subsection{A busy street in downtown New York (busy New York street)-- Figure \ref{fig:busy_new_york_street_ours}} 
\noindent\begin{mdframed}
Input Scene Description : \textbf { a busy street in downtown new york } 
Semantic shopping list
\begin{enumerate}
\item street lamps: white with gold accents.bright and shining.
\item street lamp post: black iron, with a height of 3 feet.unblemished and free from rust.
\item street lamp light: with 500 lumens and a white light.unblemished and free from dirt and dust.
\item street lamp wires: rated to withstand a voltage of 600 v.free from fraying and breaks in the insulation.
\item street lamp bulbs: with a warm white light and a power consumption of 7 watts.bright and free from flickering.
\item car: black with gold accents.in good condition, freshly washed.
\item car windows: no visible scratches or cracks.
\item car doors: no dents or discoloration.
\item car tires: no visible punctures.
\item license plate: no obvious signs of wear or tear.
\item benches: steel, lined with cushioning for comfort.clean and dust-free.
\item trash can: minimal signs of wear from pedestrian traffic.
\item fire hydrant: freshly painted yellow finish, with no rusting or damage.
\item street lights: bright and in good condition, without any dimming.
\item buildings: no visible signs of damage or deterioration.
\item street signs: reflective and weatherproof material.clear and legible.
\item street posts: obvious signs of rust, but otherwise in good condition.
\item street signs: clear lettering with no signs of wear.
\item trash bins: minor dents and scratches, but otherwise in good condition.
\item pigeons: hand-painted pigeons.brightly coloured, no signs of wear or fading.
\end{enumerate}
\end{mdframed}

We consider this example a failure case, and was reflected by the CLIP-D/S score for this scene in our main paper. This shows that this models still has many ways to go for outdoor scenes, a direction for future work.



\end{appendices}
\end{document}